\documentclass{article}

\usepackage{mwe}
\usepackage{xcolor}
\usepackage{colortbl} % \rowcolor for highlighting table rows
\usepackage{array}
\usepackage{booktabs}
\usepackage{multirow}
\usepackage{longtable}
\PassOptionsToPackage{sort}{natbib} % auto-sort \cite key lists low to high (no range compression)
\usepackage[preprint]{corl_2026} % Uncomment for pre-prints (e.g., arxiv); This is like ``final'', but will remove the CORL footnote.

\usepackage{booktabs}
\usepackage{tabularx}
\usepackage{wrapfig}
\usepackage{array} % Required for custom column types
\usepackage{arydshln} % dotted/dashed partial rules (\cdashline) in tables
\usepackage{capt-of} % \captionof for side-by-side float captions
\usepackage{amsmath, amssymb, bm}
\usepackage{enumitem}
\usepackage{titletoc} % appendix-only table of contents via \startcontents/\printcontents
\usepackage[all]{hypcap} % figure/table links jump to top of float, not the caption
\usepackage{pifont}
\newcommand{\cmark}{\textcolor{green!60!black}{\ding{52}}}
\newcommand{\xmark}{\textcolor{red!80!black}{\ding{56}}}

\usepackage{etoolbox}
\AtEndEnvironment{wrapfigure}{\vspace{4pt}}         % bottom-only gap (top stays flush via intextsep=0)
\AtEndEnvironment{wraptable}{\vspace{4pt}}

\makeatletter
\newcommand\blfootnote[1]{%
  \begingroup
  \renewcommand\thefootnote{}%
  \refstepcounter{Hfootnote}%
  \hyper@makecurrent{Hfootnote}%
  \global\let\Hy@footnote@currentHref\@currentHref
  \footnotetext{#1}%
  \endgroup
}
\makeatother

\newcommand{\method}{\textsc{HUG}\xspace}
\newcommand{\dataset}{\textsc{1M-HUGs}\xspace}
\newcommand{\benchmark}{\textsc{HUG-Bench}\xspace}
\newcommand{\website}{\url{https://grasping.io}}
\definecolor{linkblue}{rgb}{0,0.25,0.9}
\definecolor{rowhl}{rgb}{0.90,0.94,1.0} % table row highlight

\usepackage[dvipsnames]{xcolor}

\newcommand{\eg}{\emph{e.g.}\xspace}
\newcommand{\ie}{\emph{i.e.}\xspace}

\title{Human Universal Grasping}

\author{
  Kevin Yuanbo Wu\textsuperscript{\textmd{1,$\dagger$}}\quad
  Tianxing Zhou\textsuperscript{\textmd{1,2}}\quad
  Isaac Tu\textsuperscript{\textmd{1}}\quad
  Billy Yan\textsuperscript{\textmd{1}} \quad
  Irmak Guzey\textsuperscript{\textmd{1}} \\
  {\bfseries David Fouhey}\textsuperscript{1}\quad
  {\bfseries Dandan Shan}\textsuperscript{1,3,$\ddagger$}\quad
  {\bfseries Lerrel Pinto}\textsuperscript{1,$\ddagger$} \\[4pt]
  \textsuperscript{\textmd{1}}New York University \quad
  \textsuperscript{\textmd{2}}Tsinghua University \quad
  \textsuperscript{\textmd{3}}University of Michigan \\[10pt]
  {\hypersetup{urlcolor=linkblue}\large\website{}}
}

\begin{document}
\maketitle
\blfootnote{$\dagger$\,Correspondence to: \texttt{k.wu@nyu.edu}. \quad $\ddagger$\,Equal advising.}

%===============================================================================

% \vspace{-0.2in}
\begin{abstract}
Humans can grasp objects effortlessly, whereas multi-fingered robots are far from this level of generality. We argue that the most natural source of robot grasping data is from humans, who pick up thousands of objects every day.
We present \method, a flow-matching model that generates diverse human grasps for any user-specified object in a single RGB-D image captured from a stereo camera. Using smart glasses, we first collect \dataset, an egocentric dataset of human grasps spanning 1M frames (27.8 hrs) and 6,707 object instances across 41 buildings. Next, to model the distribution of natural human grasps, our novel flow-matching model fuses RGB and depth observations to output a grasp parameterized by wrist translation, wrist rotation, and MANO hand pose. Predicted grasps can be retargeted to various robot hands, enabling zero-shot grasping in everyday scenes. To standardize evaluation, we build a new simulated benchmark, \benchmark, of 90 unseen objects from five geometric categories and various sizes, with metric-scale 3D meshes. We evaluate \method in the real world on the 30-object \texttt{test} set of \benchmark across multiple stereo cameras, robot embodiments, and household environments. \method outperforms the state-of-the-art grasping baselines by $+23\%$ and $+34\%$ on our challenging object set. Code, data, benchmark, checkpoints, and an interactive demo are released on our website.
\end{abstract}

\keywords{Learning from Humans, Dexterous Grasping}

%===============================================================================

\vspace*{\fill}
\begin{center}
    \includegraphics[width=\linewidth]{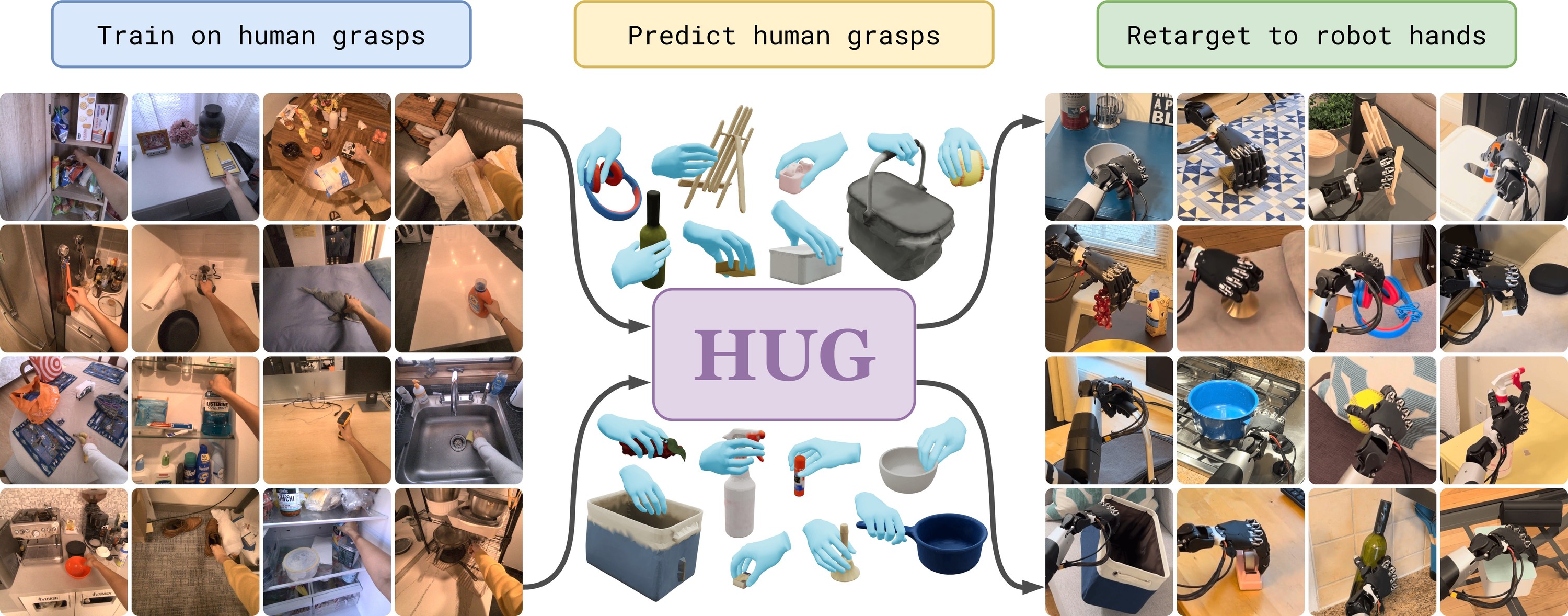}
    \captionof{figure}{\textbf{\method learns dexterous grasping without any robot data.} Trained solely on egocentric human grasp data, \method generates diverse human grasps for real-world objects in a single RGBD image captured from a stereo camera, which can be retargeted to robot hands for zero-shot, in-the-wild dexterous grasping.}
    \label{fig:teaser}
\end{center}

\vspace*{\fill}
\clearpage

\section{Introduction}
\label{sec:introduction}

Grasping arbitrary objects unlocks downstream manipulation, from sorting groceries to operating tools. Humans do this effortlessly, yet current models for dexterous robot hands remain far from such generality. A major bottleneck is data: robots need the diverse, real-world grasping experience humans accumulate daily.

Prior work attacks this from two angles. \emph{Synthetic grasps} are produced in simulation by optimizing analytic objectives like force-closure~\cite{wang2023dexgraspnetlargescaleroboticdexterous, fang2025anydexgraspgeneraldexterousgrasping}, sampling learned generators or RL policies~\cite{ye2025dex1blearning1bdemonstrations, xu2023unidexgraspuniversalroboticdexterous}, or reconstructing hands from web images~\cite{chen2025web2grasp, gupta2026grasp}; they suffer the sim-to-real gap and usually require retraining for each new hand. \emph{Teleoperation}~\cite{iyer2024open, arunachalam2023holo, ding2024bunnyvisionpro, qin2023anyteleop} yields real grasps on the target embodiment but is tedious and cannot cover the open world.

We instead train on in-the-wild human grasps, modeling how people naturally grasp objects rather than every physically valid grasp, yielding reliably executable grasps. Two recent advances make this practical. First, lightweight egocentric sensors like Aria Gen 2~\cite{aria-gen-2} stream calibrated RGB-D and hand tracking, making grasp collection as easy as wearing glasses. Second, anthropomorphic robot hands~\cite{ruka, abilityhand, shaw2023leap, wujihand} and learned retargeting~\cite{qin2023anyteleop, mandi2025dexmachina, li2025maniptrans,wuji2026retargeting} have narrowed the human-robot morphology gap, facilitating direct robot deployment. Together they unlock a previously infeasible pipeline: \emph{collect human grasps at scale, learn from them, and retarget for deployment.}

We demonstrate this with \method (\underline{H}uman \underline{U}niversal \underline{G}rasping), which generates diverse human grasps for objects in a single RGB-D image captured from a stereo camera that can be retargeted to robot hands. \method has three steps: (1) collect \dataset, an egocentric dataset of 1M image-grasp pairs from 6,707 recordings captured with Aria Gen 2 across 41 buildings; (2) train a flow-matching model mapping objects in an RGB-D to a MANO~\cite{romero2017mano} grasp; (3) retarget the predicted grasp to robot hands with no per-embodiment training.

We introduce \benchmark, 90 challenging unseen objects spanning five geometric categories and three size bins. Unlike simulation-only benchmarks, \benchmark starts from real objects, reconstructs each into a metric-scale mesh from egocentric recordings, and evaluates in both simulation and the real world. On the 30 object \texttt{test} set in the real world, \method reaches $66.7\%$ tabletop success, beating baselines by $+23\%$ and $+34\%$, and $62.0\%$ in-the-wild, generalizing zero-shot across stereo cameras, robot hands, and unseen homes.

\noindent In summary, our contributions are the following, all of which we open-source:
\begin{enumerate}[leftmargin=*, noitemsep, topsep=0pt, partopsep=0pt]
    \item To our knowledge, \method is the first grasping framework trained purely on human data and deployable across multiple robot embodiments.
    \item \textbf{Dataset.} \dataset, 1M egocentric image-grasp pairs of natural human grasps across 6,707 recordings and 41 buildings, with MANO-fit hand poses and metric depth (\S~\ref{sec:dataset}).
    \item \textbf{Method.} \method, a point-conditioned flow-matching model that predicts MANO grasps from RGB-D and retargets to multiple embodiments without per-hand training (\S~\ref{sec:method}).
    \item \textbf{Benchmark.} \benchmark, 90 unseen objects with metric-scale 3D meshes for paired simulation and real-world evaluation (\S~\ref{sec:experiments}).
\end{enumerate}

\section{Related Work}
\label{sec:related_work}

\textbf{Robotic object grasp prediction.}
Grasp prediction is long-standing in robotics. Early work targeted two-fingered grippers via self-supervised collection~\cite{pinto2015supersizing} or large-scale datasets~\cite{graspnet-1b, fang2023anygrasprobustefficientgrasp, mousavian20196, sundermeyer2021contact, chavandafle2022simultaneousobjectreconstructiongrasp}, achieving strong real-world performance~\cite{okrobot, cui2026contactanchoredpoliciescontactconditioning}. However, multi-fingered hands remain harder. As dexterous teleoperation data is hard to collect, prior work mostly trains in simulation, via reinforcement learning~\cite{lum2024dextrahg, singh2025dextrahrgbvisuomotorpoliciesgrasp, christen2022dgrasp, wan2023unidexgrasp++} or generative grasp synthesis~\cite{zhong2025dexgraspanythinguniversalrobotic,lu2024ugg, ye2025dex1blearning1bdemonstrations}, both of which struggle with sim-to-real gaps and require retraining per robot. Like \method, these methods~\cite{wan2023unidexgrasp++,zhong2025dexgraspanythinguniversalrobotic,lu2024ugg} use 3D representations, but rely on a complete object point cloud, which hinders generalization during real-world deployment. \method instead predicts from single-view camera-frame RGB-D and, trained on human grasping data, is scalable, generalizable, and easily retargeted to robot hands.

\textbf{Robot learning from non-robot datasets.}
Given the difficulty of collecting robot-specific data, recent work learns robot behaviors from non-teleoperated datasets~\cite{cui2026contactanchoredpoliciescontactconditioning, etukuru2025robot, chi2024universal, xu2025dexumi, aina, hudor} via advances in motion tracking~\cite{cotracker, doersch2023tapirtrackingpointperframe} and hand-object reconstruction~\cite{ye2023diffusion, pavlakos2024reconstructing} and interaction synthesis~\cite{ye2023affordance}. Early efforts used in-domain human data with rich 3D annotations~\cite{demo-diffusion, point-policy, hudor, wang2023mimicplay}, but tying collection to deployment limited scalability. In-the-wild human datasets~\cite{ego4d, track2act, shaw2023videodex, zeromimic, srirama2024hrp, chen2025web2grasp} scale broadly but lack the precise 3D signals for reliable policies, requiring downstream engineering~\cite{zeromimic, srirama2024hrp} or co-training with robot data~\cite{tao2025dexwild, gao2026dreamdojo}. Smart glasses~\cite{aria-gen-1, aria-gen-2} bridge this gap by capturing in-the-wild data with stereo depth and accurate 3D hand poses. Most of these efforts focus on two-finger grippers~\cite{egozero}, and some additionally rely on robot data for training~\cite{egomimic}. Unlike these, \method learns multi-fingered dexterous grasping from in-the-wild human data alone, with no robot data, and retargets to multiple robot embodiments at deployment.

\section{\dataset  Dataset}
\label{sec:dataset}

\begin{figure}[t]
    \centering
    \includegraphics[width=\linewidth]{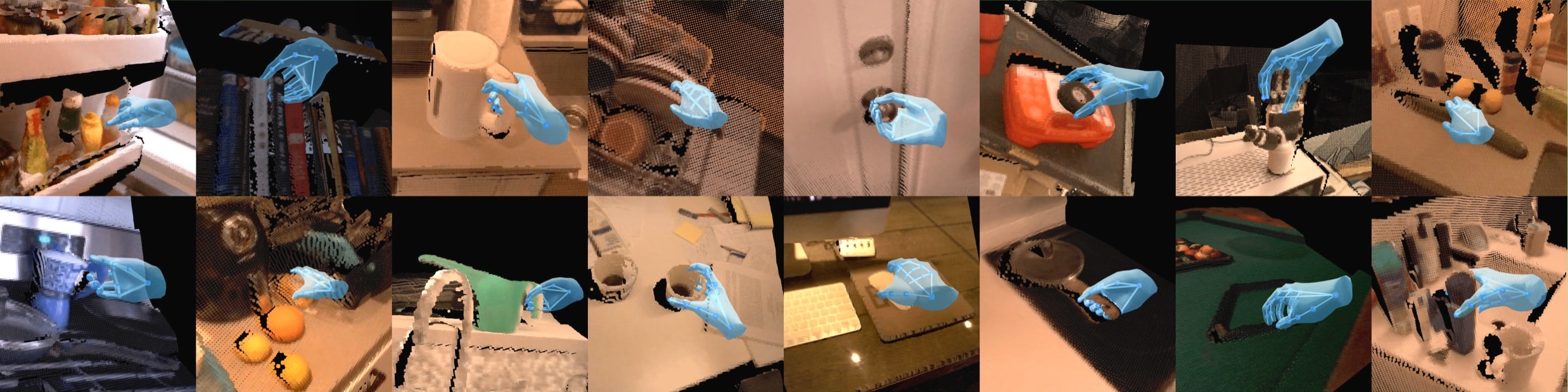}
    \caption{\textbf{\dataset dataset.} Our training data comprises 1M egocentric frames of human grasps, spanning 6,707 object instances. Each entry provides synchronized RGB and grayscale views, metric depth, an object mask, and a MANO hand pose with wrist transformation in the camera frame.}
    \label{fig:dataset}
\end{figure}

\dataset captures a diverse set of natural human grasps with egocentric smart glasses in everyday environments (Figure~\ref{fig:dataset}), differing from existing datasets with grasps from simulation or lab settings (Table~\ref{tab:dataset_comparison}). This section presents the collection protocol, filtering pipeline, and resulting dataset.

\textbf{Video recording protocol.}
We record with Aria Gen 2 glasses~\cite{aria-gen-2}, which provide synchronized RGB and stereo grayscale views together with 6-DoF camera poses and 3D hand landmarks. In one recording, the wearer stands in front of a target object and moves their head for 15-30 seconds, capturing the static scene from diverse viewpoints without hands visible, before reaching in with their right hand and grasping the object. The grasp pose is then propagated back into preceding no-hand frames with Aria Gen 2's camera poses, so a single physical grasp yields hundreds of (object-only image, grasp) training pairs from diverse viewpoints at no additional annotation cost.

\begin{wraptable}{r}{0.5\linewidth}
  \centering
  \caption{\textbf{Comparison of grasping datasets.} \dataset captures real in-the-wild human grasps and automatically yields many (image, grasp) pairs from dynamic views by back-propagating grasp across the no-hand frames.}
  \label{tab:dataset_comparison}
  \scriptsize
  \setlength{\tabcolsep}{2pt}
  \renewcommand{\arraystretch}{1.15}
  \vspace{4pt}

  \begin{tabular*}{\linewidth}{@{\extracolsep{\fill}}lllll}
    \toprule
    \textbf{Dataset}                                                 & \textbf{Real data}         & \textbf{H. Grasp}       & \textbf{\# Obj.}    & \textbf{\# I-G pair} \\
    \midrule
    DexGraspNet~\cite{wang2023dexgraspnetlargescaleroboticdexterous} & \xmark                 & \xmark                    & 5.4K          & 1.3M          \\
    Dex1B~\cite{ye2025dex1blearning1bdemonstrations}                 & \xmark                 & \xmark                    & 4.4K          & 1B            \\
    Web2Grasp~\cite{chen2025web2grasp}                               & \cmark~(web)           & \xmark                    & 1K            & 2.1K          \\
    AnyDexGrasp~\cite{fang2025anydexgraspgeneraldexterousgrasping}   & \cmark~(lab)           & \cmark                    & 144           & 10K           \\
    DexYCB~\cite{chao2021dexycb}                                     & \cmark~(lab)           & \cmark                    & 20            & 1K            \\
    \midrule
    \textbf{\dataset~(Ours)}                                                    & \textbf{\cmark~(wild)} & \textbf{\cmark}           & \textbf{$\sim$1.5K} & \textbf{1.0M} \\
    \bottomrule
  \end{tabular*}
\end{wraptable}

\textbf{Curation.}
Before filtering, each recording is localized to the grasped object and verified. A vision-language model identifies the grasped object, SAM3~\cite{carion2026sam3segmentconcepts} propagates its mask across all frames, and stability and proximity heuristics select the grasp frame. Every recording is then human-reviewed in a web interface before it enters the dataset. We detail the full pipeline in Appendix~\ref{sec:auto_processing} and~\ref{sec:annotation_app}.

\textbf{Frame filtering.}
We evaluate frames against five criteria: (i) the object mask is non-empty; (ii) stereo depth with S2M2~\cite{min2025stextsuperscript2mtextsuperscript2scalablestereomatching} marks $\geq\!60\%$ of the depth map as confident; (iii) the 2D projection of the grasp's hand landmarks intersects the object mask; (iv) at least five grasp landmarks lie within the image bounds; and (v) there is no hand in the current frame.
Each surviving frame is cropped and resized to $224\!\times\!224$. We provide more details in Appendix~\ref{sec:dataset_prep}.

\textbf{Dataset statistics and labels.} The dataset contains 6{,}707 recordings across 41 buildings with an estimated $\sim$1.5K unique objects. Within each building we grasp whatever we can find, spanning hundreds of distinct environments (kitchens, bedrooms, etc.). After filtering, 1M RGB and 1M grayscale frames (from the left stereo camera) remain, for 2M training entries; each grayscale frame shares its timestep with an RGB frame and lets the model generalize to monochrome cameras. Each entry contains a $224\!\times\!224$ image, camera intrinsics, depth map, object mask, and the grasp pose in the camera frame. Since Aria provides only 21 hand landmarks, we fit a full articulated MANO~\cite{romero2017mano} hand to them (details in Appendix~\ref{sec:mano_optimization}). MANO decouples hand geometry into a shape parameter $\boldsymbol{\beta}$ (overall hand size and proportions) and a pose parameter $\boldsymbol{\theta}$ (joint articulation). MANO is convenient for three reasons. First, with $\boldsymbol{\beta}$ fixed to a canonical hand size, the same $\boldsymbol{\theta}$ denotes the same grasp across collectors, removing per-person hand size variation. Second, the resulting articulated mesh can be loaded into simulation to execute predicted grasps (\S~\ref{sec:exp_sim}). Third, fitting under an anatomical-validity prior~\cite{yang2021cpf} regularizes grasps into physically valid poses. The full pipeline, from raw Aria recordings through stereo depth, MANO fitting, and dataset construction, is released as \texttt{aria2mano}.

\section{Method}
\label{sec:method}

\begin{figure}[t]
    \centering
    \includegraphics[width=\linewidth]{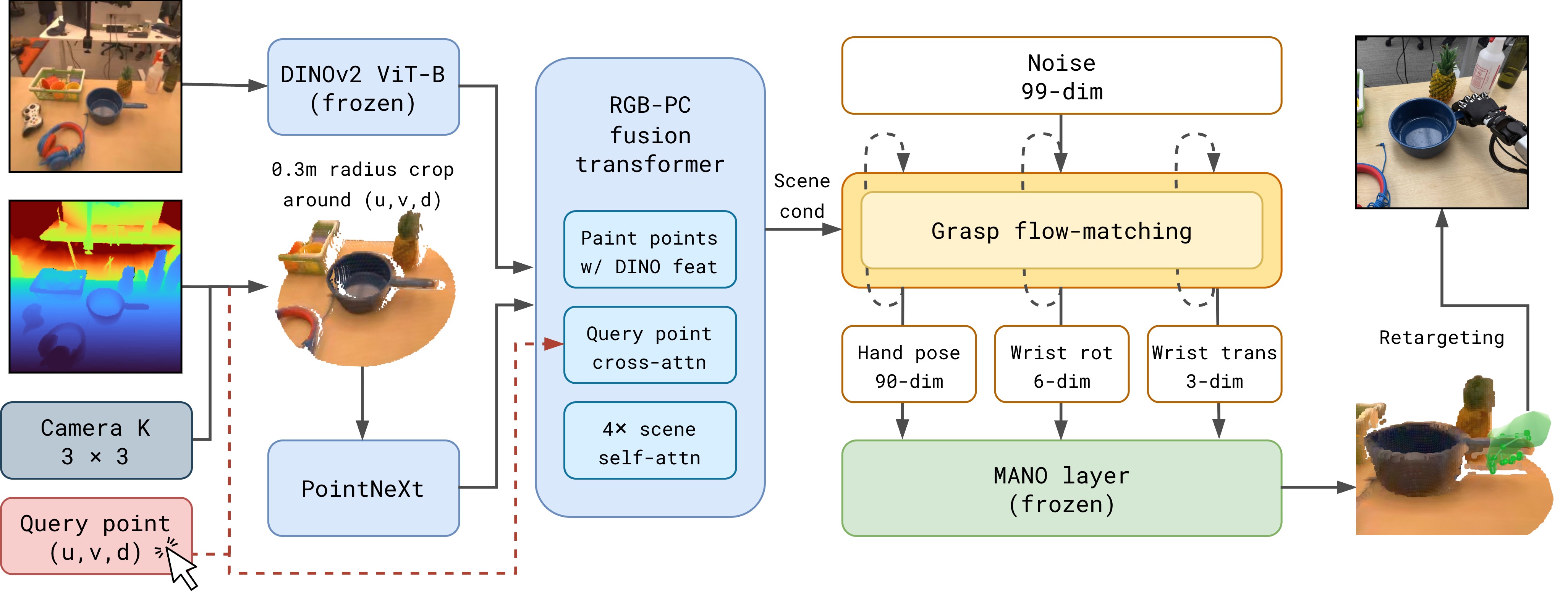}
    \caption{\textbf{\method architecture.} Conditioned on an RGB-D image and a query point on the target object, \method predicts MANO hand grasps via a flow-matching transformer over fused RGB and point cloud features. Predicted human grasps are then retargeted to robot hands.}
    \label{fig:method}
\end{figure}

Figure~\ref{fig:method} shows \method, a flow-matching model that, trained on real-world human grasps, generates diverse natural grasps for any user-specified object in a single RGB-D image from a stereo camera.

\subsection{Model Architecture}
\label{sec:method_architecture}

\begin{figure}[t]
    \centering
    \includegraphics[width=\linewidth]{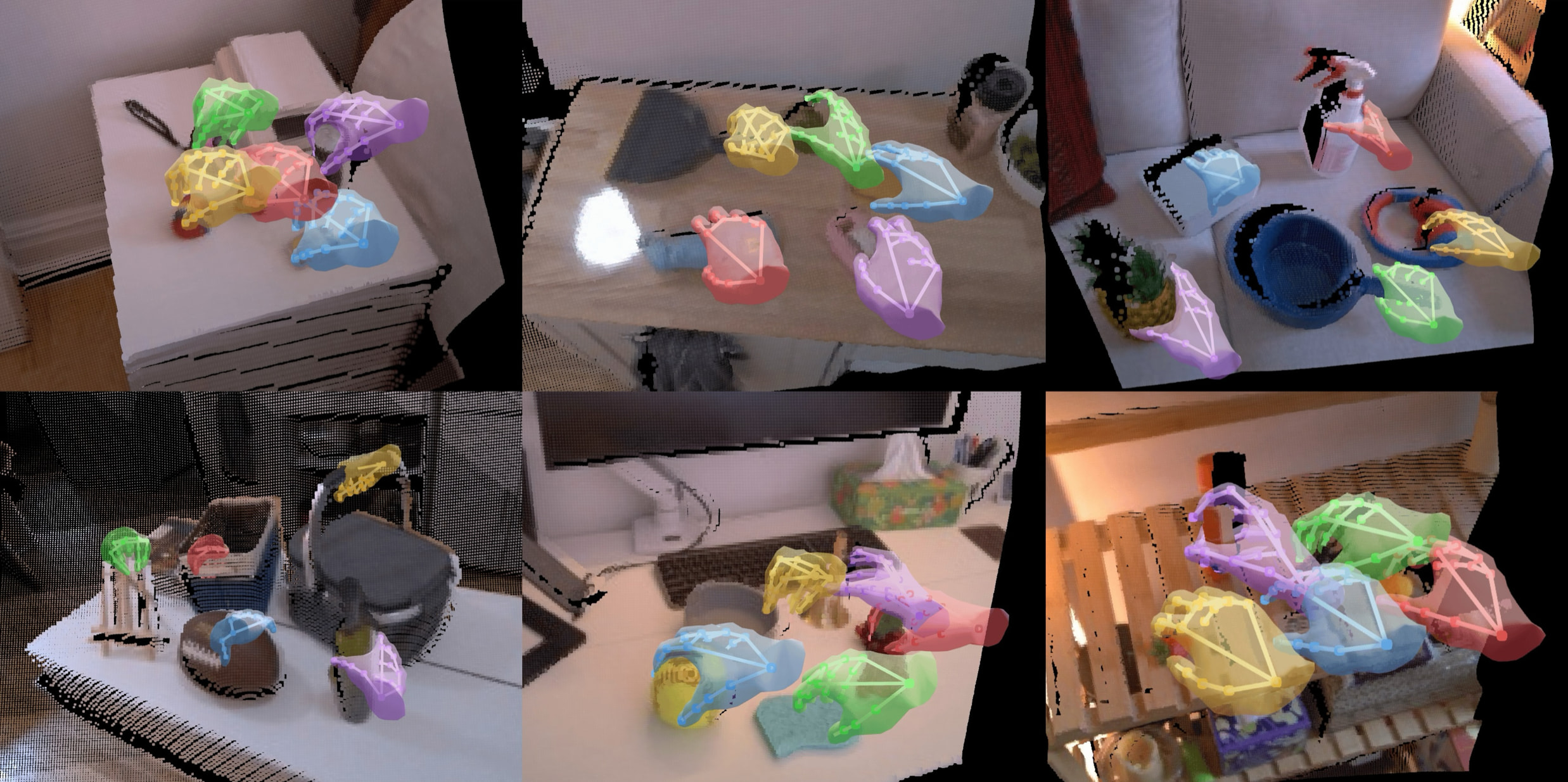}
    \caption{\textbf{Predicted grasps on \benchmark.} \method's predicted grasps for 30 unseen objects across six scenes of the \benchmark \texttt{test} split. \method generalizes across a variety of object shapes and sizes, environments, and camera viewpoints. Top row: \texttt{small\_2}, \texttt{medium\_1}, \texttt{large\_1}. Bottom row: \texttt{large\_2}, \texttt{medium\_2}, \texttt{small\_1}. See Appendix Table~\ref{tab:object_stats} for the objects in each scene.}
    \label{fig:bench_grasps}
\end{figure}

Given an RGB-D observation and a 2D pixel click $(u,v)$ on the target object, \method predicts a $99$-dim grasp state $\mathbf{x}=[\,\mathbf{t},\,\mathbf{R}_{\text{6d}},\,\boldsymbol{\theta}_{\text{6d}}\,]\in\mathbb{R}^{99}$, where $\mathbf{t}\in\mathbb{R}^{3}$ is the wrist translation in the camera frame in the OpenCV convention, $\mathbf{R}_{\text{6d}}\in\mathbb{R}^{6}$ is the global wrist rotation in the continuous 6D rotation representation of~\citet{zhou2019continuity}, and $\boldsymbol{\theta}_{\text{6d}}\in\mathbb{R}^{15\times 6}$ collects the 6D rotations of the 15 MANO finger joints. The depth image is back-projected to a metric point cloud (PC), and the click is lifted to a 3D query point $\mathbf{p}_q\in\mathbb{R}^{3}$ using its depth value and the camera intrinsics $\mathbf{K}$. The MANO shape $\boldsymbol{\beta}\in\mathbb{R}^{10}$ is held fixed at a single canonical value, so the network predicts only articulation and placement; we recompute all grasps in \dataset and \benchmark to this shape for consistency (Appendix~\ref{sec:fixed_shape}).

\textbf{Encoders.} The RGB image is encoded with a frozen DINOv2-Base ViT with register tokens~\cite{oquab2023dinov2,darcet2024registers}, producing $N\!=\!256$ patch tokens. The metric point cloud is cropped to a $0.3$\,m radius ball around the 3D query point $\mathbf{p}_q$. From this crop, $N_p\!=\!4096$ points are sampled and passed to a trainable PointNeXt~\cite{qian2022pointnext} U-Net that outputs $N\!=\!256$ per-region tokens together with their metric XYZ centroids $\{\mathbf{c}_i\}_{i=1}^{N}$. Cropping keeps the tokens dense around the target object; we set the radius to $0.3$\,m, roughly the largest object graspable with one hand (Appendix~\ref{sec:crop_radius}).

\textbf{RGB-PC fusion transformer.} We fuse the two encoder streams with point painting: each point cloud centroid is projected to the RGB image via camera intrinsics $\mathbf{K}$, its DINOv2 patch feature is bilinearly sampled, concatenated with the corresponding PC token, and projected by a two-layer MLP into a $D_f\!=\!1024$-dim fused token $\mathbf{f}_i$. (Without point painting, the two streams are concatenated into $2N$ tokens; we ablate this in \S~\ref{sec:exp_sim}.) Both the query point $\mathbf{p}_q$ and the PC centroids $\{\mathbf{c}_i\}$ are encoded with a shared random Fourier feature encoder $\gamma(\cdot)$~\cite{tancik2020fourier} to retain metric information. The fused tokens $\{\mathbf{f}_i\}_{i=1}^{N}$ cross-attend to the query token $\mathbf{q}\!=\!\mathrm{MLP}(\gamma(\mathbf{p}_q))$ and are refined by a $4$-layer pre-norm transformer, producing the scene-conditioning tokens $\mathbf{s}\!\in\!\mathbb{R}^{N\times D_f}$ used by the flow transformer. Since $\mathbf{K}$ enters \method only through back-projection and projection, never as a learned parameter, the model transfers across stereo cameras with different intrinsics (Appendix~\ref{sec:camera_generalization}).

\textbf{Flow transformer.} We split the grasp state into three distinct groups: translation ($3$-dim), wrist rotation ($6$-dim), and finger pose ($90$-dim). Each group is normalized and projected into a $D_m\!=\!512$-dim token; separate tokens keep geometrically distinct components from over-mixing and balance the gradient signal across groups. These tokens are passed through $L\!=\!6$ DiT~\cite{peebles2023dit} blocks that cross-attend to the scene tokens $\mathbf{s}$ and are conditioned on the timestep via AdaLN-Zero modulation. Three linear heads decode the output tokens back to $3$, $6$, and $90$ dims, and the velocity is integrated by the flow-matching ODE in normalized space, with the resulting clean state de-normalized to produce the predicted grasp state.

\subsection{Training}
\label{sec:method_training}

\textbf{Loss.} We combine a velocity-prediction MSE $\mathcal{L}_{\text{v}}$ with geometric supervision: we recover the predicted clean state $\hat{\mathbf{x}}_0=\mathbf{x}_t-t\,f_\phi(\mathbf{x}_t,t,\mathbf{s})$, pass it through MANO, and supervise the resulting 3D hand landmarks in the camera frame with an L1 loss $\mathcal{L}_{3\text{D}}$, using $\lambda_{\text{v}}\!=\!1$ and $\lambda_{3\text{D}}\!=\!20$:
\begin{equation}
\mathcal{L} = \lambda_{\text{v}}\,\mathcal{L}_{\text{v}}
            + \lambda_{3\text{D}}\,(1-t)\,\mathcal{L}_{3\text{D}},
\label{eq:loss}
\end{equation}
where the $(1-t)$ weight concentrates the geometric loss on near-clean steps where $\hat{\mathbf{x}}_0$ is meaningful. 

\textbf{Training details.} We train for 100K steps with AdamW at learning rate $10^{-4}$ and batch size $128$, using a 5K-step linear learning rate warmup. Only the PointNeXt encoder, RGB-PC fusion module, and flow transformer are optimized. Training timesteps are drawn uniformly from $[0,1]$, and at inference we generate samples with $50$-step Euler integration of the learned ODE. We keep an EMA from step 50K, validate in MuJoCo~\cite{todorov2012mujoco} (\S~\ref{sec:exp_sim}) every 5K steps. Training with DDP on two RTX 5090s (batch size $64$ per GPU) takes ${\sim}10$\,h including MuJoCo validation (Appendix~\ref{sec:hyperparameters}).

\section{Experiments}
\label{sec:experiments}

We introduce \benchmark (\S~\ref{sec:exp_benchmark}), then evaluate \method in simulation (\S~\ref{sec:exp_sim}) and real-world (\S~\ref{sec:exp_real}).

\begin{figure}[t]
    \centering
    \includegraphics[width=\linewidth]{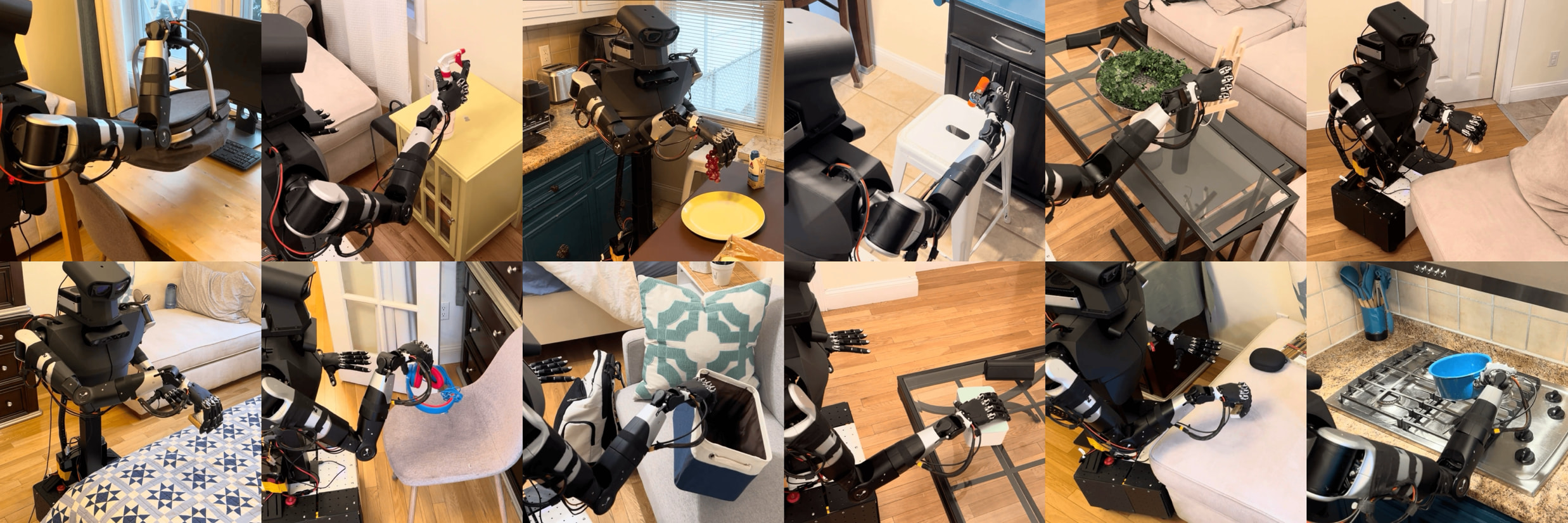}
    \caption{\textbf{Real world grasping with \method.} Grasp executions on unseen objects from \benchmark \texttt{test} split in an unseen home, performed by a YOR mobile manipulator equipped with WUJI hands.}
    \label{fig:real_world_grasps}
\end{figure}

\subsection{\benchmark}
\label{sec:exp_benchmark}

\begin{wrapfigure}{r}{0.39\linewidth}
    \centering
    \includegraphics[width=\linewidth]{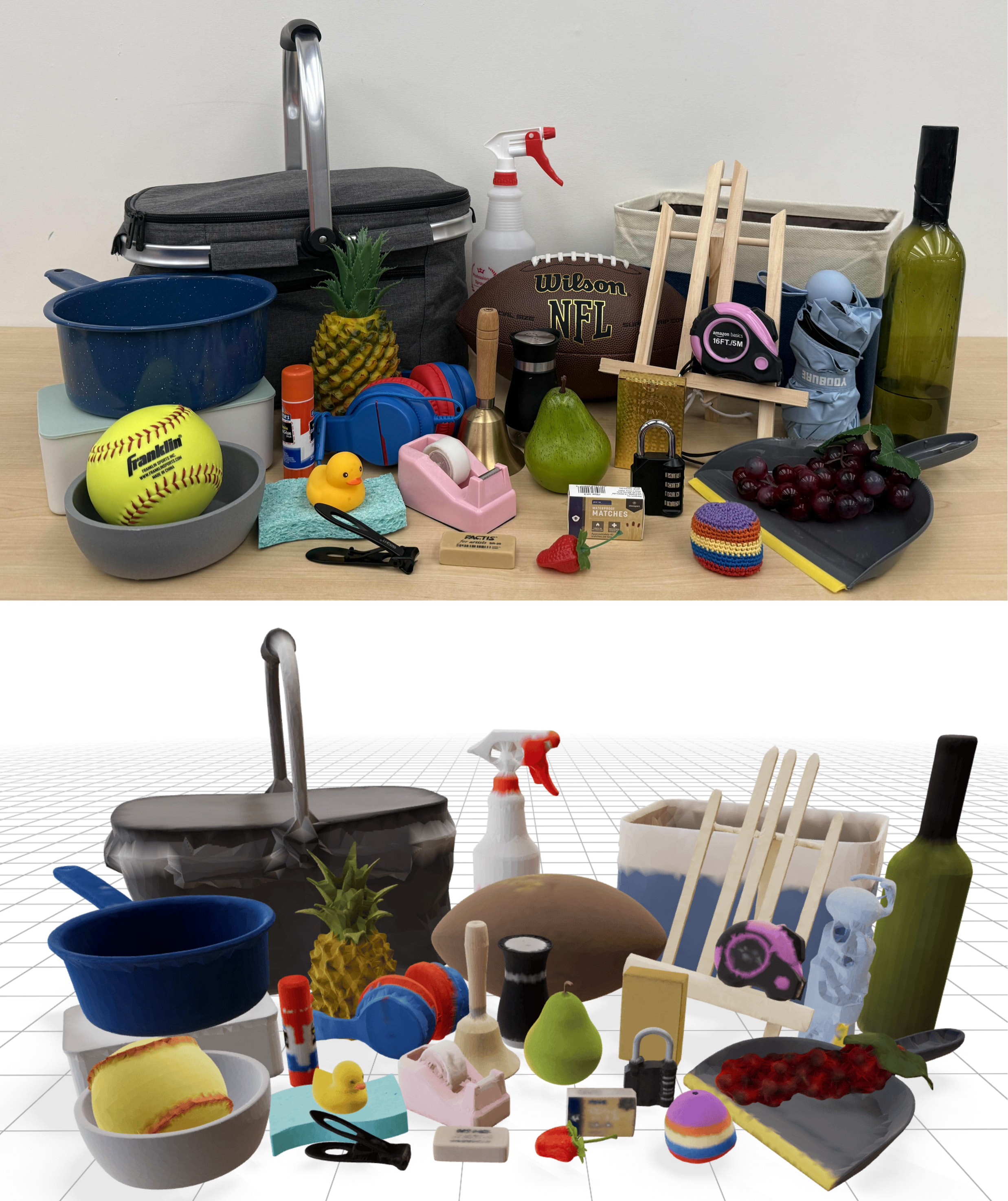}
    \caption{\textbf{\benchmark \texttt{test} split.} 30 unseen objects (top) from 5 geometric categories $\times$ 3 size bins (2 per class), with their simulation assets (bottom).}
    \label{fig:benchmark_objects}
\end{wrapfigure}

\textbf{Evaluation dataset.} We propose a difficult-to-grasp set of objects for evaluation. \benchmark comprises 90 objects spanning five geometric categories (cylindrical, spheroidal, prismatic, appendaged, amorphous) and three size bins (small, medium, large), with six objects per class combination (four \texttt{val}, two \texttt{test}). These objects are hard to grasp: many objects are articulated, very short ($\sim$1\,cm tall), or large and unwieldy, requiring diverse grasp poses and precise spatial accuracy for successful grasping. All 90 objects are unseen during training. The 30 \texttt{test} objects for real-world evaluation (Figure~\ref{fig:benchmark_objects}) can be purchased on Amazon for $\sim$250 USD, linked on our website. The \texttt{val} objects are used only in simulation for checkpoint selection, so only the unseen \texttt{test} objects are purchased for unbiased real-world evaluation.

\textbf{Metric-scale simulation assets.} To produce simulation-ready evaluation assets of the \benchmark objects at metric scale, we extend Multi-view SAM3D (MV-SAM3D)~\cite{li2026mvsam3dadaptivemultiviewfusion}, injecting Aria camera intrinsics, extrinsics, and stereo depth into their pipeline. We collect scans of 18 scenes, with five objects per scene using Aria Gen 2 glasses. We pass in five spread-out views per scene to the modified MV-SAM3D. We inspect the resulting meshes in Viser~\cite{yi2025viserimperativewebbased3d}, editing their scale and pose to align them with the Aria Gen 2 SLAM semidense point cloud, stereo depth points, and their 2D projections onto the input views. Lastly, we make the meshes watertight with Alpha Wrap~\cite{10.1145/3528223.3530152} and obtain their convex decomposition with CoACD~\cite{wei2022coacd} for simulation. For each object we also record 10 human grasps with Aria Gen 2, which we replay as a ground-truth \emph{human grasp oracle} (\S~\ref{sec:exp_sim}). We release this scan-to-asset pipeline, together with the MuJoCo environment, as \texttt{aria2mesh} (Appendix~\ref{sec:benchmark}).

\begin{table}[t]
    \centering
    \begin{minipage}[t]{0.68\linewidth}
        \vspace{0pt}
        \centering
        \setlength{\tabcolsep}{5pt}
        {\scriptsize
        \vspace{4pt}
        \begin{tabular}{lcccc}
            \toprule
            & \multicolumn{2}{c}{\textbf{\benchmark \texttt{val}}} & \multicolumn{2}{c}{\textbf{\benchmark \texttt{test}}} \\
            \cmidrule(lr){2-3} \cmidrule(lr){4-5}
            Method & SR (\%) $\uparrow$ & FC error (mm) $\downarrow$ & SR (\%) $\uparrow$ & FC error (mm) $\downarrow$ \\
            \midrule
            \rowcolor{rowhl}
            RGB + PC (full \method) & \textbf{71.5} {\scriptsize $\pm$ 1.8} & \textbf{19.0} {\scriptsize $\pm$ 0.8} & \textbf{73.0} {\scriptsize $\pm$ 2.6} & \textbf{14.6} {\scriptsize $\pm$ 0.9} \\
            \quad w/o crop & 61.2 {\scriptsize $\pm$ 2.0} & 21.6 {\scriptsize $\pm$ 0.9} & 58.0 {\scriptsize $\pm$ 2.8} & 25.7 {\scriptsize $\pm$ 1.5} \\
            \quad w/o point paint & 61.8 {\scriptsize $\pm$ 2.0} & 21.9 {\scriptsize $\pm$ 1.0} & 58.3 {\scriptsize $\pm$ 2.8} & 23.3 {\scriptsize $\pm$ 1.7} \\
            \quad w/o 3D loss & 39.2 {\scriptsize $\pm$ 2.0} & 33.0 {\scriptsize $\pm$ 1.2} & 32.7 {\scriptsize $\pm$ 2.7} & 35.7 {\scriptsize $\pm$ 2.2} \\
            PC only & 64.2 {\scriptsize $\pm$ 2.0} & 25.6 {\scriptsize $\pm$ 1.2} & 70.7 {\scriptsize $\pm$ 2.6} & 22.1 {\scriptsize $\pm$ 1.5} \\
            \quad w/o crop & 47.3 {\scriptsize $\pm$ 2.0} & 32.6 {\scriptsize $\pm$ 1.5} & 50.0 {\scriptsize $\pm$ 2.9} & 32.8 {\scriptsize $\pm$ 2.2} \\
            RGB only & 26.8 {\scriptsize $\pm$ 1.8} & 95.4 {\scriptsize $\pm$ 3.6} & 29.7 {\scriptsize $\pm$ 2.6} & 108.6 {\scriptsize $\pm$ 5.1} \\
            \midrule
            Human grasp (oracle) & 90.3 {\scriptsize $\pm$ 1.2} & 9.4 {\scriptsize $\pm$ 0.3} & 94.0 {\scriptsize $\pm$ 1.4} & 7.4 {\scriptsize $\pm$ 0.3} \\
            \bottomrule
            \vspace{4pt}
        \end{tabular}}%
        \vspace{6pt}
        \captionof{table}{\textbf{Simulation results and ablations.} Simulation success rate (SR) and fingertip contact (FC) error on \benchmark over 10 grasps per object (600 \texttt{val}, 300 \texttt{test}; mean $\pm$ SE). Each model is evaluated on the unseen \texttt{test} objects at its best-\texttt{val}-SR checkpoint. Human grasp represents an oracle upper bound.}
        \label{tab:sim_ablations}
    \end{minipage}\hfill
    \begin{minipage}[t]{0.29\linewidth}
        \vspace{0pt}
        \centering
        \includegraphics[width=\linewidth]{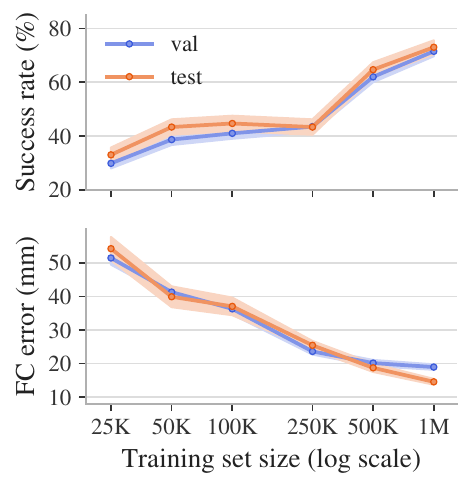}
        \captionof{figure}{\textbf{Dataset scaling.} Impact of dataset size on \benchmark SR and FC error (Eq.~\ref{eq:fc_error}); training sets are nested proper subsets.}
        \label{fig:data_scale}
    \end{minipage}
\end{table}

\begin{wrapfigure}{r}{0.39\linewidth}
    \centering
    \includegraphics[width=\linewidth]{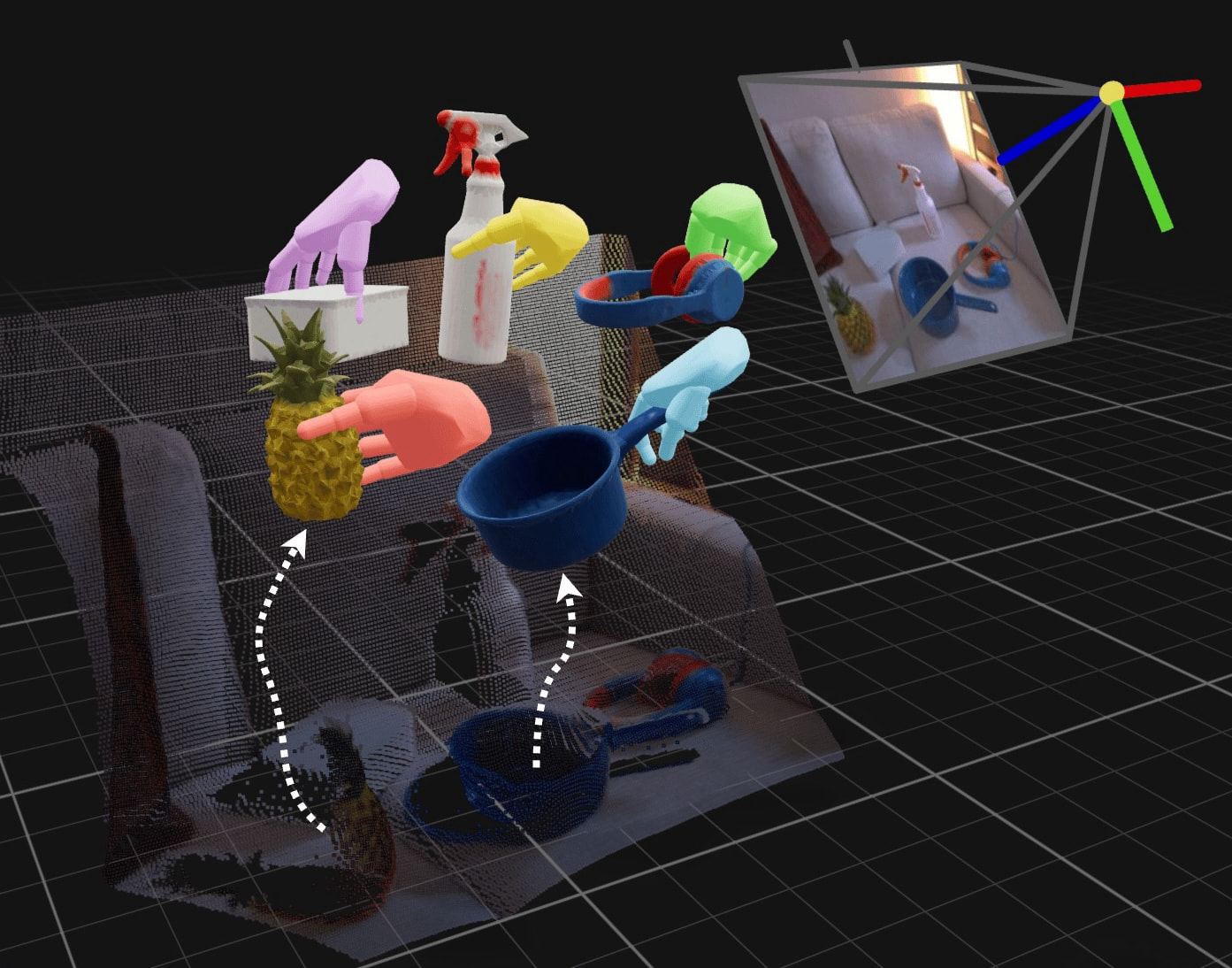}
    \caption{\textbf{Real-to-sim grasping.} Evaluating \method on \benchmark in simulation using real captured inputs.}
    \label{fig:sim_evaluation}
\end{wrapfigure}

\textbf{Simulation environment.} Each predicted grasp is evaluated in MuJoCo~\cite{todorov2012mujoco} (Figure~\ref{fig:sim_evaluation}) on a fresh single-object scene with the target resting on a table under gravity. A position-actuated MANO right hand (Appendix~\ref{sec:mano_mjcf}) executes an open-loop pre-grasp $\rightarrow$ grasp $\rightarrow$ lift rollout with gravity on. The pre-grasp pose offsets the predicted grasp by $3$\,cm along the two non-lateral wrist axes with fingers opened; the wrist then linearly interpolates to the predicted pose while the fingers close to their predicted MANO joint angles, with a small extra flexion to apply force. Then, the wrist lifts straight up by $0.5$\,m. A grasp succeeds if the object is no longer in contact with the surface after the lift.

\subsection{Simulation Grasping Evaluation}
\label{sec:exp_sim}

\textbf{Metrics.} \emph{Success rate} (SR, \%) is the fraction of trials in which the object remains grasped at the end of the lift. SR alone is coarse, for example, failures can result from open-loop execution, so it does not fully capture grasp quality. We therefore also report \emph{fingertip contact error} (FC error, mm), measuring how close the thumb and the closest supporting finger come to the object surface:
\begin{equation}
\mathrm{FC} \;=\; \tfrac{1}{2}\!\left(\,|d_{\text{thumb}}| \;+\; \min_{f \in \mathcal{F}}\, |d_{f}|\,\right),
\label{eq:fc_error}
\end{equation}
where $d_i$ is the signed distance from fingertip $i$ to the object surface and $\mathcal{F}$ is the set of non-thumb fingers (smaller is better). FC error assumes a good grasp brings the thumb and a supporting fingertip near the object surface. Both metrics are averaged over 10 grasps per object on 60 \texttt{val} and 30 \texttt{test} objects; we pick checkpoints by best \texttt{val} SR. Per-object simulation success rates are reported in Appendix Table~\ref{tab:object_stats}, and additional simulation results in Appendix~\ref{sec:simulation_results}.

\textbf{Ablation study.} Table~\ref{tab:sim_ablations} reports the model architecture ablation. Our full model takes a dual-modal \emph{RGB+PC} input, applies a $0.3$\,m \emph{crop} of the point cloud around the user-clicked point, augments each point with \emph{DINO point painting} features, and uses an auxiliary \emph{3D loss} on fingertip placement (Eq.~\ref{eq:loss}). To isolate the contribution of each input stream, we additionally compare against single-modality \emph{PC-only} and \emph{RGB-only} variants.

The \emph{human grasp oracle} replays the 10 recorded human grasps on each object, estimating an upper bound on what is achievable in our simulator. It falls short of $100\%$ due to hand-tracking error in our Aria Gen 2 data~\cite{projectaria_gen2_mps_benchmarks}, slight inaccuracies in object assets, and open-loop execution failures (Appendix~\ref{sec:sim_realism}). Our full model reaches $71.5\%$ \texttt{val} SR and $73.0\%$ \texttt{test} SR, within ${\sim}20$ points of the oracle ($90.3$ and $94.0\%$). Figure~\ref{fig:bench_grasps} shows that it produces stable, human-like grasps across object shapes, sizes, and viewpoints. The 3D loss is the most critical component: removing it cuts \texttt{test} SR by over $40$ points to $32.7\%$ and more than doubles FC error from $14.6$ to $35.7$\,mm, confirming that explicit 3D supervision is essential for accurate fingertip placement.
The $0.3$\,m crop and DINO point painting each cost ${\sim}10$ \texttt{val}-SR and ${\sim}15$ \texttt{test}-SR points when removed, showing that both a targeted spatial window for dense point cloud context and rich per-point features matter.

\begin{figure}[t]
    \centering
    \includegraphics[width=\linewidth]{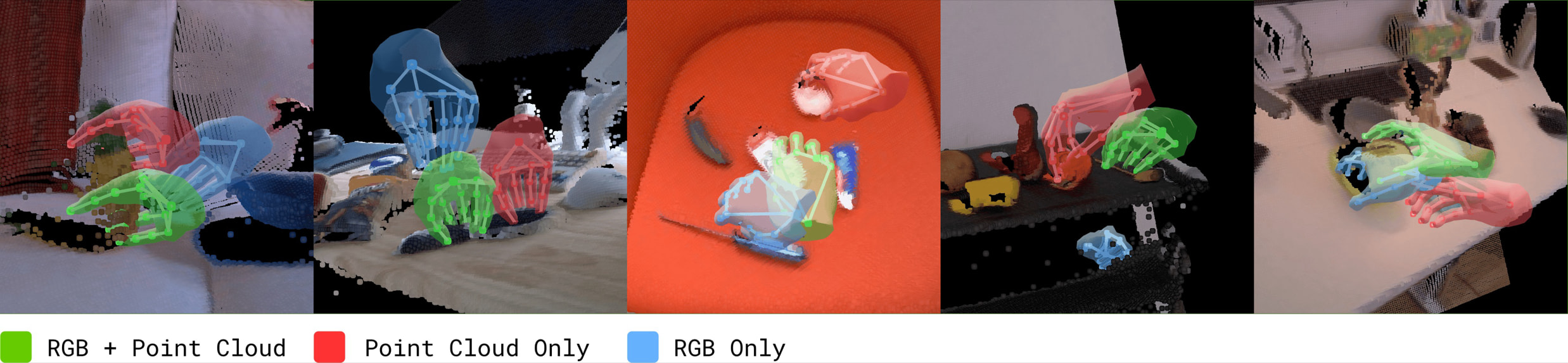}
    \caption{\textbf{Single-modality failures.} Cases where RGB-only or PC-only prediction fails but RGB+PC succeeds. Objects, left to right: pineapple, hair brush, anchovies, spoon, softball.}
    \label{fig:modality_failures}
\end{figure}

\textbf{RGB and depth are complementary.} On the modality axis, PC-only remains a strong standalone baseline at $64.2\%$ \texttt{val} SR and $70.7\%$ \texttt{test} SR, while RGB-only collapses to $26.8\%$ \texttt{val} SR and $29.7\%$ \texttt{test} SR. The contrast is sharper in FC error: RGB-only rises to $95$\,mm \texttt{val} and $109$\,mm \texttt{test}, and PC-only rises to $25.6$\,mm \texttt{val} and $22.1$\,mm \texttt{test}, up from $19.0$ and $14.6$\,mm for the full model, showing that RGB grounding sharpens fingertip placement. Figure~\ref{fig:modality_failures} shows that the two streams are complementary: RGB-only rarely reaches the vicinity of the object, while PC-only reaches it but lacks semantic grounding, grasping the leafy top of a pineapple or the bristles of a brush rather than the body or handle, and snapping to a larger nearby object when depth is unreliable on small targets. RGB+PC resolves these cases, motivating the dual-modal design.

\textbf{Data scaling.} We train the full RGB+PC model on different dataset sizes: 25K, 50K, 100K, 250K, 500K, and 1M (all) RGB frames. Figure~\ref{fig:data_scale} shows the data scaling study. From 25K to 1M frames, \texttt{test} SR climbs from 33\% to 73\% and FC error falls from 54.2\,mm to 14.6\,mm. Neither saturates at 1M, suggesting the model is still data-bound, not capacity-bound at this scale. The \texttt{val} and \texttt{test} curves track tightly across all sizes, so gains transfer to held-out objects without overfitting.

\subsection{Real-World Grasping Evaluation}
\label{sec:exp_real}

We evaluate \method on the 30 \benchmark \texttt{test} objects in the real world, running 10 trials per object, or 300 per method (Table~\ref{tab:real_world_results}). We retarget predicted MANO grasps to the robot hand (Appendix~\ref{sec:retargeting}) and execute it open-loop pre-grasp $\rightarrow$ grasp $\rightarrow$ lift. All rollout videos are on our website.

\begin{wrapfigure}{r}{0.5\linewidth}
    \centering
    \includegraphics[width=\linewidth]{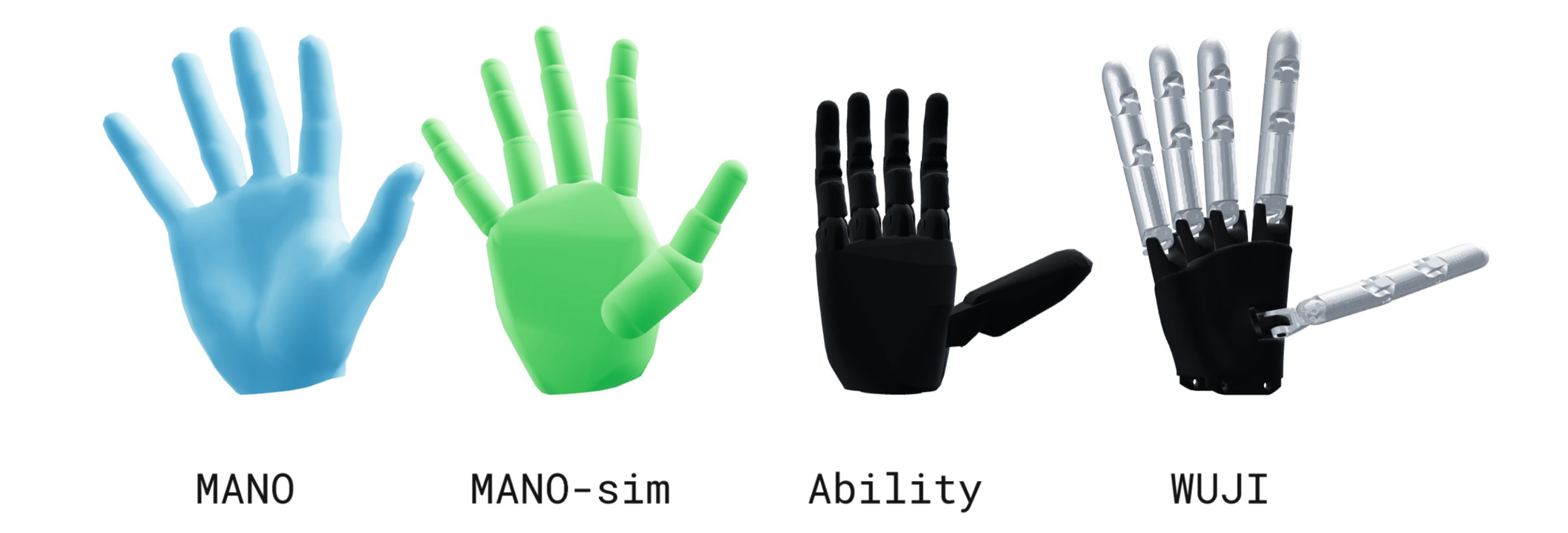}
    \caption{\textbf{Hand sizes.} The fixed-shape MANO hand alongside its simulation mesh and the Ability and WUJI robot hands. WUJI is a similar size to \method's fixed hand size, but Ability is much smaller.}
    \label{fig:hand_sizes}
\end{wrapfigure}

\textbf{Baselines.} We compare \method against two recent learning-based grasping methods. \emph{Dex1B}~\cite{ye2025dex1blearning1bdemonstrations} is a generative multi-fingered grasping model trained on 1B simulated demonstrations by combining grasp optimization with generative sampling, representing the sim-to-real paradigm for dexterous grasping. \emph{CAP} (Contact-Anchored Policies)~\cite{cui2026contactanchoredpoliciescontactconditioning} conditions parallel-jaw grasping on a specified contact anchor and reports strong real-world performance compared to other methods~\cite{fang2023anygrasprobustefficientgrasp, intelligence2025pi05visionlanguageactionmodelopenworld}.

\textbf{Tabletop experiments.} \method and Dex1B are deployed on a 6-DoF Ability hand~\cite{abilityhand} mounted on a 7-DoF xArm~\cite{xarm}, using a third-person ZED stereo camera for input observations. CAP follows its published configuration of an iPhone wrist camera and a parallel-jaw gripper for execution. Like \method, Dex1B is executed open-loop and occasionally predicts a grasp whose trajectory intersects the table; we count severe table-intersecting grasps as a failure to avoid damaging our hardware.

\method reaches $66.7\%$ overall success on the 30 \texttt{test} objects, exceeding Dex1B ($43.7\%$) and CAP ($32.7\%$) by $+23\%$ and $+34\%$ and grasping $28/30$ objects at least once. The evaluation involves no test-set bias (Appendix~\ref{sec:real_design}): the checkpoint with the best \texttt{val} SR (\S~\ref{sec:exp_sim}) is deployed directly on the \texttt{test} split with no per-object tuning, and each trial is a single grasp prediction followed by one open-loop execution without motion planning.

The absolute success rates are low as \benchmark is challenging, spanning sizes and geometries well beyond the medium, convex objects on which dexterous grasping methods typically report high success. Dex1B is accurate on this familiar regime (\eg pear $10/10$), but drops on large or unwieldy objects (storage bin $0/10$) and complex geometry (easel $2/10$). CAP is limited by its hardware: its parallel jaw cannot close around objects exceeding its span (storage bin $0/10$) nor precisely grasp thin objects (eraser $0/10$), and grasps mainly pinchable items (glue stick $10/10$).

\method is the most robust method across the geometry and size grid. It is alone in grasping large prismatic objects where both baselines fail (storage bin $10/10$ vs.\ $0/10$), and far outperforms both on objects with handles or irregular structure that demand precise, structure-aware contact (picnic basket $9/10$ vs.\ $1/10$ and $0/10$; spray bottle $9/10$ vs.\ $4/10$ and $1/10$; easel $8/10$ vs.\ $2/10$ and $4/10$), while remaining strong on small objects (eraser $6/10$ vs.\ $3/10$ and $0/10$; strawberry $7/10$ vs.\ $4/10$ and $0/10$). Common failures of \method are due to hardware limitations of the Ability hand: the football and wipe dispenser are grasped $0/10$ by all methods, as the small Ability hand cannot wrap them (Figure~\ref{fig:hand_sizes}).

\begin{table}[t]
  \centering
  \setlength{\tabcolsep}{2pt}
  \renewcommand{\arraystretch}{0.9}
  \newcolumntype{R}{>{\centering\arraybackslash}p{1.35cm}}
  {\scriptsize
  \begin{tabular}{lll|RRR|R|>{\raggedright\arraybackslash\hspace*{4pt}}p{2cm}R}
  \toprule
  \multicolumn{3}{c|}{\begin{tabular}[b]{@{}c@{}}\textbf{\benchmark} \\ \texttt{test} split\end{tabular}} & \multicolumn{3}{c|}{\begin{tabular}[b]{@{}c@{}}\textbf{Tabletop} \\ ZED + xArm + Ability\end{tabular}} & \multicolumn{1}{c|}{\begin{tabular}[b]{@{}c@{}}\textbf{Simulation} \\ MANO Hand\end{tabular}} & \multicolumn{2}{c}{\begin{tabular}[b]{@{}c@{}}\textbf{In-the-wild} \\ Aria + YOR + WUJI\end{tabular}} \\
  \cmidrule(lr){1-3} \cmidrule(lr){4-6} \cmidrule(lr){7-7} \cmidrule(lr){8-9}
  Geometry & Size & Object & Dex1B & CAP & \method & \method & Location & \method \\
  \cmidrule(lr){1-3} \cmidrule(lr){4-6} \cmidrule(lr){7-7} \cmidrule(lr){8-9}
  \multirow{6}{*}{Cylindrical}
    & \multirow{2}{*}{S}  & Glue stick      & 7/10 & \textbf{10/10} & 6/10 & 5/10 & Stool & 7/10 \\
    &                          & Pepper shaker   & 8/10 & 5/10 & \textbf{10/10} & 10/10 & Kitchen island & 6/10 \\
    & \multirow{2}{*}{M} & Umbrella        & 4/10 & \textbf{7/10} & 5/10 & 10/10 & Couch 1 & 6/10 \\
    &                          & Bowl            & 3/10 & 0/10 & \textbf{4/10} & 8/10 & Kitchen island & 1/10 \\
    & \multirow{2}{*}{L}  & Spray bottle    & 4/10 & 1/10 & \textbf{9/10} & 10/10 & Side table & 9/10 \\
    &                          & Wine bottle     & \textbf{7/10} & 3/10 & 3/10 & 10/10 & Kitchen counter & 10/10 \\
  \cmidrule(lr){1-3} \cmidrule(lr){4-6} \cmidrule(lr){7-7} \cmidrule(lr){8-9}
  \multirow{6}{*}{Spheroidal}
    & \multirow{2}{*}{S}  & Strawberry      & 4/10 & 0/10 & \textbf{7/10} & 4/10 & Bed 1 & 6/10 \\
    &                          & Hacky sack      & 5/10 & 8/10 & \textbf{9/10} & 10/10 & Bed 2 & 10/10 \\
    & \multirow{2}{*}{M} & Pear            & \textbf{10/10} & 4/10 & \textbf{10/10} & 10/10 & Dining table & 10/10 \\
    &                          & Softball        & \textbf{8/10} & 0/10 & 5/10 & 10/10 & Ottoman & 8/10 \\
    & \multirow{2}{*}{L}  & Pineapple       & 8/10 & 5/10 & \textbf{10/10} & 10/10 & Dining chair & 9/10 \\
    &                          & Football        & 0/10 & 0/10 & 0/10 & 8/10 & Nightstand & 0/10 \\
  \cmidrule(lr){1-3} \cmidrule(lr){4-6} \cmidrule(lr){7-7} \cmidrule(lr){8-9}
  \multirow{6}{*}{Prismatic}
    & \multirow{2}{*}{S}  & Eraser          & 3/10 & 0/10 & \textbf{6/10} & 6/10 & Bed 1 & 6/10 \\
    &                          & Match box       & 5/10 & 3/10 & \textbf{8/10} & 9/10 & Couch 2 & 8/10 \\
    & \multirow{2}{*}{M} & Card deck   & 3/10 & 0/10 & \textbf{8/10} & 4/10 & Bed 2 & 8/10 \\
    &                          & Sponge          & 3/10 & \textbf{6/10} & \textbf{6/10} & 8/10 & Kitchen counter & 7/10 \\
    & \multirow{2}{*}{L}  & Wipe dispenser  & 0/10 & 0/10 & 0/10 & 3/10 & Coffee table 1 & 4/10 \\
    &                          & Storage bin     & 0/10 & 0/10 & \textbf{10/10} & 7/10 & Couch 3 & 8/10 \\
  \cmidrule(lr){1-3} \cmidrule(lr){4-6} \cmidrule(lr){7-7} \cmidrule(lr){8-9}
  \multirow{6}{*}{Appendaged}
    & \multirow{2}{*}{S}  & Nail clipper   & 3/10 & 0/10 & \textbf{5/10} & 3/10 & Couch 2 & 4/10 \\
    &                          & Lock            & \textbf{6/10} & \textbf{6/10} & 5/10 & 8/10 & Dresser & 6/10 \\
    & \multirow{2}{*}{M} & Dustpan        & 5/10 & 2/10 & \textbf{6/10} & 6/10 & Bed 1 & 6/10 \\
    &                          & Handbell           & 8/10 & 5/10 & \textbf{10/10} & 7/10 & Couch 1 & 4/10 \\
    & \multirow{2}{*}{L}  & Saucepan        & \textbf{5/10} & 3/10 & 4/10 & 10/10 & Stove & 7/10 \\
    &                          & Picnic basket   & 1/10 & 0/10 & \textbf{9/10} & 5/10 & Desk & 6/10 \\
  \cmidrule(lr){1-3} \cmidrule(lr){4-6} \cmidrule(lr){7-7} \cmidrule(lr){8-9}
  \multirow{6}{*}{Amorphous}
    & \multirow{2}{*}{S}  & Rubber duck     & 3/10 & 4/10 & \textbf{10/10} & 5/10 & Couch 3 & 9/10 \\
    &                          & Tape measure    & 3/10 & \textbf{8/10} & \textbf{8/10} & 9/10 & Coffee table 1 & 5/10 \\
    & \multirow{2}{*}{M} & Tape dispenser  & \textbf{4/10} & 2/10 & 3/10 & 10/10 & Desk & 4/10 \\
    &                          & Grapes          & 2/10 & 6/10 & \textbf{10/10} & 5/10 & Dining table & 3/10 \\
    & \multirow{2}{*}{L}  & Headphones      & \textbf{7/10} & 6/10 & 6/10 & 4/10 & Desk chair & 2/10 \\
    &                          & Easel           & 2/10 & 4/10 & \textbf{8/10} & 5/10 & Coffee table 2 & 7/10 \\
  \cmidrule(lr){1-3} \cmidrule(lr){4-6} \cmidrule(lr){7-7} \cmidrule(lr){8-9}
  \multicolumn{3}{l|}{\textit{Overall success rate}} & 43.7\% & 32.7\% & \textbf{66.7\%} & 73.0\% & & 62.0\% \\
  \multicolumn{3}{l|}{\textit{Number of objects with $\geq 1$ success}} & 27/30 & 20/30 & \textbf{28/30} & 30/30 & & 29/30 \\
  \bottomrule
  \end{tabular}}%
  \vspace{1em}
  \caption{\textbf{Real-world grasp results on \benchmark.} Per-object success rates on 30 \texttt{test} split objects in two real-world settings. \method outperforms both baselines on the tabletop by $+23\%$ and $+34\%$ and achieves a comparable success rate in-the-wild, demonstrating both strong grasping capability and stable cross-embodiment, cross-environment generalization.}
  \label{tab:real_world_results}
  \end{table}

\textbf{In-the-wild experiments.} Because \method is trained on human grasps from diverse real-world locations, we expect it to remain robust outside the lab. Figure~\ref{fig:real_world_grasps} shows \method deployed on a modified YOR mobile manipulator~\cite{anjaria2026yormobilemanipulatorgeneralizable} with an AgileX NERO arm~\cite{neroarm}, the 20-DoF WUJI hand~\cite{wujihand}, and Aria Gen 2 for vision, testing transfer to a new embodiment, camera, and uncontrolled household. With no onsite tuning of the model or WUJI retargeting~\cite{wuji2026retargeting}, we run all 300 \benchmark \texttt{test} trials consecutively in one morning. Objects are placed across rooms (kitchen, living room, bedroom, office), and no two are evaluated from the same viewpoint. \method achieves a $62.0\%$ success rate in-the-wild, only $4.7$ points below the tabletop, indicating that \method transfers gracefully across embodiments and in-the-wild environments.

\begin{figure}[t]
    \centering
    \includegraphics[width=\linewidth]{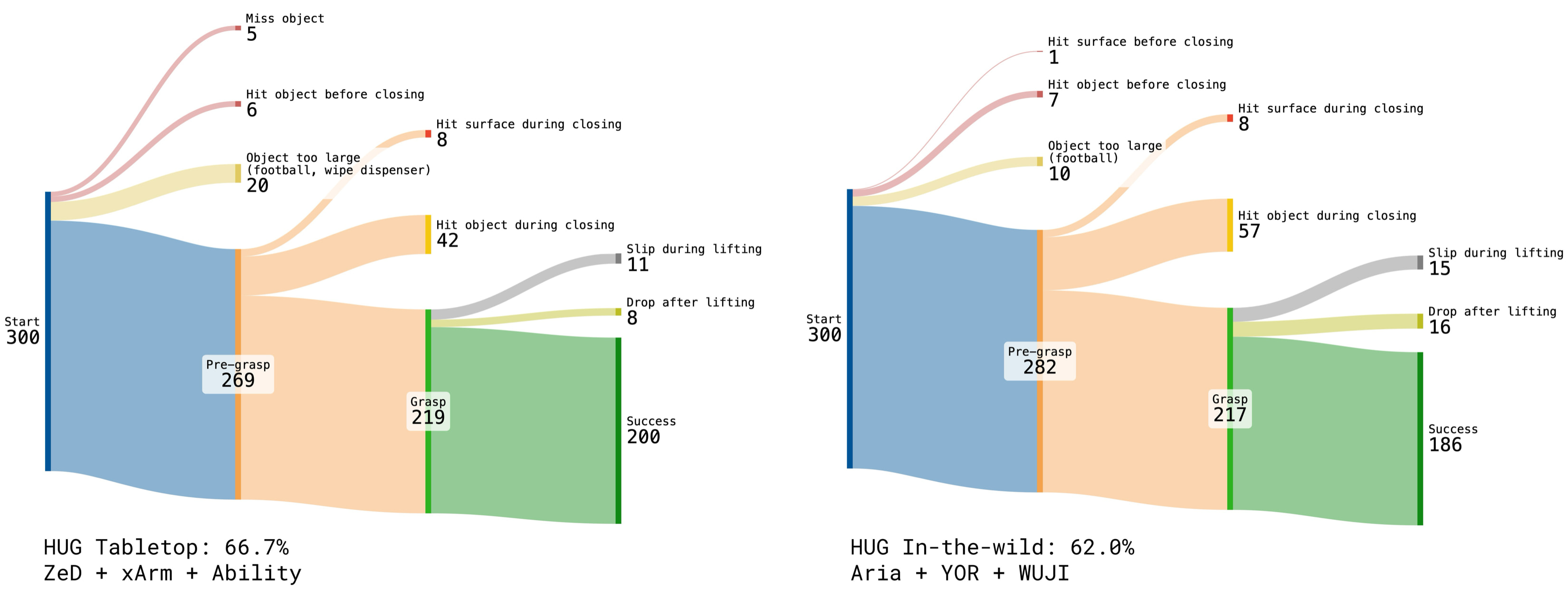}
    \caption{\textbf{Failure mode breakdown.} Grasp-outcome flow for the $300$ \benchmark \texttt{test} trials in each real-world setting, tracing every attempt through the pre-grasp, grasp, and lift stages into success or a specific failure mode.}
    \label{fig:failure_modes}
\end{figure}

\textbf{Failure modes.} Figure~\ref{fig:failure_modes} traces all $300$ \texttt{test} trials through the pre-grasp, grasp, and lift stages, with a consistent distribution across tabletop and in-the-wild. Most failures occur as the hand closes from pre-grasp to grasp, when it contacts the object or the table before the fingers settle (reporting tabletop then in-the-wild: hit object $42$ and $57$, hit surface $8$ and $8$); a smaller share misses or over-reaches before pre-grasp, often on objects too large to wrap (\eg the football); and the rest occur after a grasp is established, where the object slips during the lift ($11$ and $15$) or is dropped once raised ($8$ and $16$). Two directions should recover much of this gap. Motion planning beyond the open-loop trajectory would prevent the hand from striking the object or table as it closes. Force-aware closing would reduce post-grasp slips, since \method predicts a static pose with no notion of contact force.

\section{Conclusion}
\label{sec:conclusion}

We present \method, a framework that learns dexterous robot grasping entirely from \dataset, a large-scale egocentric dataset of 1M natural human grasps. \method predicts a MANO hand grasp from a single RGB-D image and retargets it to multiple dexterous hands at deployment, with no per-hand training. To evaluate \method, we build \benchmark, a benchmark of unseen everyday objects with metric-scale meshes that pairs simulation with real-robot trials. Across both, \method produces stable, human-like grasps that transfer zero-shot to new stereo cameras, robot embodiments, and uncontrolled households. We hope \dataset and \benchmark help move dexterous grasping toward the generality that humans achieve effortlessly.

\section{Limitations}
\label{sec:limitations}

\method has several limitations. \textbf{Hand modeling.} The model is trained on right-hand grasps only, with the MANO shape held fixed at a canonical value, so left-handed or bimanual grasping and hand-specific morphology are not modeled. Because all grasps are right-handed, some object orientations are easier to grasp than others; we vary object rotation across evaluation trials so that results are not biased toward favorable poses. \textbf{Retargeting.} The human-to-robot retargeting can fail when the target robot hand cannot reach a feasible analog of the predicted MANO pose. \textbf{Open-loop execution.} Real-world rollouts run open-loop, with no closed-loop visual feedback during contact or lift, which can cause failures on objects that shift or articulate during the trajectory. \textbf{Label noise under occlusion.} When the hand is occluded during the grasp, Aria Gen 2 hand tracking degrades, so the grasp label can be too loose or too tight. \textbf{Object scale.} Accuracy drops on very small objects, limited by the $224\!\times\!224$ input resolution, and on large or far objects, which are rare in egocentric data; normalizing the grasp translation prediction around the 3D query point may help the latter. \textbf{Single-grasp execution.} We predict and execute one grasp per trial, but generating many candidates and selecting the best is a natural extension. \textbf{Scope.} Our evaluation is indoor only.

%===============================================================================

\clearpage
% The acknowledgments are automatically included only in the final and preprint versions of the paper.
\acknowledgments{This work was supported by grants from LG, Qualcomm, Honda, Microsoft, NSF award 2339096, and ONR award N00014-22-1-2773. Lerrel Pinto is supported by the Sloan, Packard, and CIFAR Fellowships.

We thank James Fort and the Meta Project Aria team for their support using Aria Gen 2 glasses. We thank Kelly Lee, Zicheng Teng, Bowen Tan, and Blake Chang for help with data collection; and our many friends who let us collect data in their apartments. We thank Nikhil Chavan-Dafle and Jeff Cui for helpful discussions, Kanad Patel, Dhawal Kabra, and Neer Patel for robot hardware support, and Omar Rayyan, Nikhil Bhattasli, and Takuma Yoneda for MuJoCo advice.}

%===============================================================================

\bibliography{references}

@string{icra    = {IEEE International Conference on Robotics and Automation (ICRA)}}

@string{corl    = {Conference on Robot Learning (CoRL)}}

@string{cvpr    = {IEEE/CVF Conference on Computer Vision and Pattern Recognition (CVPR)}}

@string{eccv    = {European Conference on Computer Vision (ECCV)}}

@string{neurips = {Advances in Neural Information Processing Systems (NeurIPS)}}

@string{rss     = {Robotics: Science and Systems (RSS)}}

@string{iclr    = {International Conference on Learning Representations (ICLR)}}

@misc{gyenes2026fourierfeaturesletagents,
      title={Fourier Features Let Agents Learn High Precision Policies with Imitation Learning}, 
      author={Balázs Gyenes and Emiliyan Gospodinov and Jan Frieling and Enrico Krohmer and Nicolas Schreiber and Xiaogang Jia and Niklas Freymuth and Gerhard Neumann},
      year={2026},
      eprint={2606.12334},
      archivePrefix={arXiv},
      primaryClass={cs.LG},
      url={https://arxiv.org/abs/2606.12334}, 
}

@inproceedings{ye2023affordance,
  title={Affordance diffusion: Synthesizing hand-object interactions},
  author={Ye, Yufei and Li, Xueting and Gupta, Abhinav and De Mello, Shalini and Birchfield, Stan and Song, Jiaming and Tulsiani, Shubham and Liu, Sifei},
  booktitle={Proceedings of the IEEE/CVF Conference on Computer Vision and Pattern Recognition},
  pages={22479--22489},
  year={2023}
}

@inproceedings{ye2023diffusion,
  title={Diffusion-guided reconstruction of everyday hand-object interaction clips},
  author={Ye, Yufei and Hebbar, Poorvi and Gupta, Abhinav and Tulsiani, Shubham},
  booktitle={Proceedings of the IEEE/CVF international conference on computer vision},
  pages={19717--19728},
  year={2023}
}

@article{chen2025web2grasp,
  title={Web2grasp: Learning functional grasps from web images of hand-object interactions},
  author={Chen, Hongyi and Yao, Yunchao and Ye, Yufei and Xu, Zhixuan and Bharadhwaj, Homanga and Wang, Jiashun and Tulsiani, Shubham and Erickson, Zackory and Ichnowski, Jeffrey},
  journal={arXiv preprint arXiv:2505.05517},
  year={2025}
}

@inproceedings{li2025maniptrans,
  title={Maniptrans: Efficient dexterous bimanual manipulation transfer via residual learning},
  author={Li, Kailin and Li, Puhao and Liu, Tengyu and Li, Yuyang and Huang, Siyuan},
  booktitle={Proceedings of the IEEE/CVF Conference on Computer Vision and Pattern Recognition},
  pages={6991--7003},
  year={2025}
}

@article{oquab2023dinov2,
  title={Dinov2: Learning robust visual features without supervision},
  author={Oquab, Maxime and Darcet, Timoth{\'e}e and Moutakanni, Th{\'e}o and Vo, Huy and Szafraniec, Marc and Khalidov, Vasil and Fernandez, Pierre and Haziza, Daniel and Massa, Francisco and El-Nouby, Alaaeldin and others},
  journal={arXiv preprint arXiv:2304.07193},
  year={2023}
}

@inproceedings{zhou2019continuity,
  title={On the continuity of rotation representations in neural networks},
  author={Zhou, Yi and Barnes, Connelly and Lu, Jingwan and Yang, Jimei and Li, Hao},
  booktitle={Proceedings of the IEEE/CVF conference on computer vision and pattern recognition},
  pages={5745--5753},
  year={2019}
}

@misc{pinto2015supersizing,
      title={Supersizing Self-supervision: Learning to Grasp from 50K Tries and 700 Robot Hours}, 
      author={Lerrel Pinto and Abhinav Gupta},
      year={2015},
      eprint={1509.06825},
      archivePrefix={arXiv},
      primaryClass={cs.LG},
      url={https://arxiv.org/abs/1509.06825}, 
}

@misc{fang2023anygrasprobustefficientgrasp,
      title={AnyGrasp: Robust and Efficient Grasp Perception in Spatial and Temporal Domains}, 
      author={Hao-Shu Fang and Chenxi Wang and Hongjie Fang and Minghao Gou and Jirong Liu and Hengxu Yan and Wenhai Liu and Yichen Xie and Cewu Lu},
      year={2023},
      eprint={2212.08333},
      archivePrefix={arXiv},
      primaryClass={cs.RO},
      url={https://arxiv.org/abs/2212.08333}, 
}

@misc{fang2025anydexgraspgeneraldexterousgrasping,
      title={AnyDexGrasp: General Dexterous Grasping for Different Hands with Human-level Learning Efficiency}, 
      author={Hao-Shu Fang and Hengxu Yan and Zhenyu Tang and Hongjie Fang and Chenxi Wang and Cewu Lu},
      year={2025},
      eprint={2502.16420},
      archivePrefix={arXiv},
      primaryClass={cs.RO},
      url={https://arxiv.org/abs/2502.16420}, 
}

@misc{xu2023unidexgraspuniversalroboticdexterous,
      title={UniDexGrasp: Universal Robotic Dexterous Grasping via Learning Diverse Proposal Generation and Goal-Conditioned Policy}, 
      author={Yinzhen Xu and Weikang Wan and Jialiang Zhang and Haoran Liu and Zikang Shan and Hao Shen and Ruicheng Wang and Haoran Geng and Yijia Weng and Jiayi Chen and Tengyu Liu and Li Yi and He Wang},
      year={2023},
      eprint={2303.00938},
      archivePrefix={arXiv},
      primaryClass={cs.RO},
      url={https://arxiv.org/abs/2303.00938}, 
}

@misc{zhong2025dexgraspanythinguniversalrobotic,
      title={DexGrasp Anything: Towards Universal Robotic Dexterous Grasping with Physics Awareness}, 
      author={Yiming Zhong and Qi Jiang and Jingyi Yu and Yuexin Ma},
      year={2025},
      eprint={2503.08257},
      archivePrefix={arXiv},
      primaryClass={cs.CV},
      url={https://arxiv.org/abs/2503.08257}, 
}

@inbook{lu2024ugg,
   title={UGG: Unified Generative Grasping},
   ISBN={9783031728556},
   ISSN={1611-3349},
   url={http://dx.doi.org/10.1007/978-3-031-72855-6_24},
   DOI={10.1007/978-3-031-72855-6_24},
   booktitle={Computer Vision – ECCV 2024},
   publisher={Springer Nature Switzerland},
   author={Lu, Jiaxin and Kang, Hao and Li, Haoxiang and Liu, Bo and Yang, Yiding and Huang, Qixing and Hua, Gang},
   year={2024},
   month=nov, pages={414–433} 
}

@misc{wang2023dexgraspnetlargescaleroboticdexterous,
      title={DexGraspNet: A Large-Scale Robotic Dexterous Grasp Dataset for General Objects Based on Simulation}, 
      author={Ruicheng Wang and Jialiang Zhang and Jiayi Chen and Yinzhen Xu and Puhao Li and Tengyu Liu and He Wang},
      year={2023},
      eprint={2210.02697},
      archivePrefix={arXiv},
      primaryClass={cs.RO},
      url={https://arxiv.org/abs/2210.02697}, 
}

@misc{ye2025dex1blearning1bdemonstrations,
      title={Dex1B: Learning with 1B Demonstrations for Dexterous Manipulation}, 
      author={Jianglong Ye and Keyi Wang and Chengjing Yuan and Ruihan Yang and Yiquan Li and Jiyue Zhu and Yuzhe Qin and Xueyan Zou and Xiaolong Wang},
      year={2025},
      eprint={2506.17198},
      archivePrefix={arXiv},
      primaryClass={cs.RO},
      url={https://arxiv.org/abs/2506.17198}, 
}

@article{romero2017mano,
   title={Embodied hands: modeling and capturing hands and bodies together},
   volume={36},
   ISSN={1557-7368},
   url={http://dx.doi.org/10.1145/3130800.3130883},
   DOI={10.1145/3130800.3130883},
   number={6},
   journal={ACM Transactions on Graphics},
   publisher={Association for Computing Machinery (ACM)},
   author={Romero, Javier and Tzionas, Dimitrios and Black, Michael J.},
   year={2017},
   month=nov, pages={1–17} 
}

@article{wei2022coacd,
  title={Approximate convex decomposition for 3d meshes with collision-aware concavity and tree search},
  author={Wei, Xinyue and Liu, Minghua and Ling, Zhan and Su, Hao},
  journal={ACM Transactions on Graphics (TOG)},
  volume={41},
  number={4},
  pages={1--18},
  year={2022},
  publisher={ACM New York, NY, USA}
}

@misc{yi2025viserimperativewebbased3d,
      title={Viser: Imperative, Web-based 3D Visualization in Python}, 
      author={Brent Yi and Chung Min Kim and Justin Kerr and Gina Wu and Rebecca Feng and Anthony Zhang and Jonas Kulhanek and Hongsuk Choi and Yi Ma and Matthew Tancik and Angjoo Kanazawa},
      year={2025},
      eprint={2507.22885},
      archivePrefix={arXiv},
      primaryClass={cs.CV},
      url={https://arxiv.org/abs/2507.22885}, 
}

@misc{li2026mvsam3dadaptivemultiviewfusion,
      title={MV-SAM3D: Adaptive Multi-View Fusion for Layout-Aware 3D Generation}, 
      author={Baicheng Li and Dong Wu and Jun Li and Shunkai Zhou and Zecui Zeng and Lusong Li and Hongbin Zha},
      year={2026},
      eprint={2603.11633},
      archivePrefix={arXiv},
      primaryClass={cs.CV},
      url={https://arxiv.org/abs/2603.11633}, 
}

@misc{intelligence2025pi05visionlanguageactionmodelopenworld,
      title={$\pi_{0.5}$: a Vision-Language-Action Model with Open-World Generalization}, 
      author={Physical Intelligence and Kevin Black and Noah Brown and James Darpinian and Karan Dhabalia and Danny Driess and Adnan Esmail and Michael Equi and Chelsea Finn and Niccolo Fusai and Manuel Y. Galliker and Dibya Ghosh and Lachy Groom and Karol Hausman and Brian Ichter and Szymon Jakubczak and Tim Jones and Liyiming Ke and Devin LeBlanc and Sergey Levine and Adrian Li-Bell and Mohith Mothukuri and Suraj Nair and Karl Pertsch and Allen Z. Ren and Lucy Xiaoyang Shi and Laura Smith and Jost Tobias Springenberg and Kyle Stachowicz and James Tanner and Quan Vuong and Homer Walke and Anna Walling and Haohuan Wang and Lili Yu and Ury Zhilinsky},
      year={2025},
      eprint={2504.16054},
      archivePrefix={arXiv},
      primaryClass={cs.LG},
      url={https://arxiv.org/abs/2504.16054}, 
}

@inproceedings{arunachalam2023holo,
  title={Holo-dex: Teaching dexterity with immersive mixed reality},
  author={Arunachalam, Sridhar Pandian and G{\"u}zey, Irmak and Chintala, Soumith and Pinto, Lerrel},
  booktitle=icra,
  year={2023},
}

@inproceedings{iyer2024open,
  title={Open teach: A versatile teleoperation system for robotic manipulation},
  author={Iyer, Aadhithya and Peng, Zhuoran and Dai, Yinlong and Guzey, Irmak and Haldar, Siddhant and Chintala, Soumith and Pinto, Lerrel},
  booktitle=corl,
  year={2024},
}

@inproceedings{cotracker,
  title={Cotracker: It is better to track together},
  author={Karaev, Nikita and Rocco, Ignacio and Graham, Benjamin and Neverova, Natalia and Vedaldi, Andrea and Rupprecht, Christian},
  booktitle=eccv,
  year={2024}
}

@misc{doersch2023tapirtrackingpointperframe,
      title={TAPIR: Tracking Any Point with per-frame Initialization and temporal Refinement}, 
      author={Carl Doersch and Yi Yang and Mel Vecerik and Dilara Gokay and Ankush Gupta and Yusuf Aytar and Joao Carreira and Andrew Zisserman},
      year={2023},
      eprint={2306.08637},
      archivePrefix={arXiv},
      primaryClass={cs.CV},
      url={https://arxiv.org/abs/2306.08637}, 
}

@inproceedings{pavlakos2024reconstructing,
  title={Reconstructing hands in 3d with transformers},
  author={Pavlakos, Georgios and Shan, Dandan and Radosavovic, Ilija and Kanazawa, Angjoo and Fouhey, David and Malik, Jitendra},
  booktitle=cvpr,
  year={2024}
}

@misc{demo-diffusion,
      title={DemoDiffusion: One-Shot Human Imitation using pre-trained Diffusion Policy},
      author={Sungjae Park and Homanga Bharadhwaj and Shubham Tulsiani},
      year={2025},
      eprint={2506.20668},
      archivePrefix={arXiv},
      primaryClass={cs.RO},
      url={https://arxiv.org/abs/2506.20668},
}

@article{point-policy,
  title={Point Policy: Unifying Observations and Actions with Key Points for Robot Manipulation},
  author={Haldar, Siddhant and Pinto, Lerrel},
  journal={arXiv preprint arXiv:2502.20391},
  year={2025}
}

@inproceedings{hudor,
  title={Bridging the human to robot dexterity gap through object-oriented rewards},
  author={Guzey, Irmak and Dai, Yinlong and Savva, Georgy and Bhirangi, Raunaq and Pinto, Lerrel},
  booktitle={2025 IEEE International Conference on Robotics and Automation (ICRA)},
  pages={3344--3351},
  year={2025},
  organization={IEEE}
}

@article{wang2023mimicplay,
  title={MimicPlay: Long-Horizon Imitation Learning by Watching Human Play},
  author={Wang, Chen and Fan, Linxi and Sun, Jiankai and Zhang, Ruohan and Fei-Fei, Li and Xu, Danfei and Zhu, Yuke and Anandkumar, Anima},
  journal={arXiv preprint arXiv:2302.12422},
  year={2023}
}

@inproceedings{yang2021cpf,
    title = {{CPF}: Learning a Contact Potential Field to Model the Hand-Object Interaction},
    author = {Yang, Lixin and Zhan, Xinyu and Li, Kailin and Xu, Wenqiang and Li, Jiefeng and Lu, Cewu},
    booktitle = {ICCV},
    year = {2021}
}

@inproceedings{ego4d,
  title={Ego4d: Around the world in 3,000 hours of egocentric video},
  author={Grauman, Kristen and Westbury, Andrew and Byrne, Eugene and Chavis, Zachary and Furnari, Antonino and Girdhar, Rohit and Hamburger, Jackson and Jiang, Hao and Liu, Miao and Liu, Xingyu and others},
  booktitle=cvpr,
  year={2022}
}

@inproceedings{track2act,
  title={Track2act: Predicting point tracks from internet videos enables generalizable robot manipulation},
  author={Bharadhwaj, Homanga and Mottaghi, Roozbeh and Gupta, Abhinav and Tulsiani, Shubham},
  booktitle=eccv,
  year={2024},
}

@inproceedings{shaw2023videodex,
  title={Videodex: Learning dexterity from internet videos},
  author={Shaw, Kenneth and Bahl, Shikhar and Pathak, Deepak},
  booktitle=corl,
  year={2023},
}

@inproceedings{zeromimic,
  title={ZeroMimic: Distilling Robotic Manipulation Skills from Web Videos}, 
  author={Junyao Shi and Zhuolun Zhao and Tianyou Wang and Ian Pedroza and Amy Luo and Jie Wang and Jason Ma and Dinesh Jayaraman},
  year={2025},
  booktitle={International Conference on Robotics and Automation (ICRA)},
}

@inproceedings{srirama2024hrp,
  title={HRP: Human Affordances for Robotic Pre-Training},
  author={Srirama, Mohan Kumar and Dasari, Sudeep and Bahl, Shikhar and Gupta, Abhinav},
  booktitle=rss,
  year={2024}
}

@misc{aria-gen-2,
    title={{Project Aria Gen 2}},
    author={{Meta Reality Labs Research}},
    year={2026},
    howpublished={\url{https://facebookresearch.github.io/projectaria_tools/gen2/}},
    note={Accessed: 2026-06-15}
}

@article{aria-gen-1,
  title={Project aria: A new tool for egocentric multi-modal ai research},
  author={Engel, Jakob and Somasundaram, Kiran and Goesele, Michael and Sun, Albert and Gamino, Alexander and Turner, Andrew and Talattof, Arjang and Yuan, Arnie and Souti, Bilal and Meredith, Brighid and others},
  journal={arXiv preprint arXiv:2308.13561},
  year={2023}
}

@misc{egomimic,
  title={EgoMimic: Scaling Imitation Learning via Egocentric Video}, 
  author={Simar Kareer and Dhruv Patel and Ryan Punamiya and Pranay Mathur and Shuo Cheng and Chen Wang and Judy Hoffman and Danfei Xu},
  year={2024},
  eprint={2410.24221},
  archivePrefix={arXiv},
  primaryClass={cs.RO},
  url={https://arxiv.org/abs/2410.24221},
}

@misc{egozero,
      title={EgoZero: Robot Learning from Smart Glasses}, 
      author={Vincent Liu and Ademi Adeniji and Haotian Zhan and Raunaq Bhirangi and Pieter Abbeel and Lerrel Pinto},
      year={2025},
      eprint={2505.20290},
      archivePrefix={arXiv},
      primaryClass={cs.RO},
      url={https://arxiv.org/abs/2505.20290}, 
}

@misc{aina,
      title={Dexterity from Smart Lenses: Multi-Fingered Robot Manipulation with In-the-Wild Human Demonstrations}, 
      author={Irmak Guzey and Haozhi Qi and Julen Urain and Changhao Wang and Jessica Yin and Krishna Bodduluri and Mike Lambeta and Lerrel Pinto and Akshara Rai and Jitendra Malik and Tingfan Wu and Akash Sharma and Homanga Bharadhwaj},
      year={2025},
      eprint={2511.16661},
      archivePrefix={arXiv},
      primaryClass={cs.RO},
      url={https://arxiv.org/abs/2511.16661}, 
}

@misc{ruka,
      title={RUKA: Rethinking the Design of Humanoid Hands with Learning}, 
      author={Anya Zorin and Irmak Guzey and Billy Yan and Aadhithya Iyer and Lisa Kondrich and Nikhil X. Bhattasali and Lerrel Pinto},
      year={2025},
      eprint={2504.13165},
      archivePrefix={arXiv},
      primaryClass={cs.RO},
      url={https://arxiv.org/abs/2504.13165}, 
}

@INPROCEEDINGS{graspnet-1b,
  author={Fang, Hao-Shu and Wang, Chenxi and Gou, Minghao and Lu, Cewu},
  booktitle={2020 IEEE/CVF Conference on Computer Vision and Pattern Recognition (CVPR)}, 
  title={GraspNet-1Billion: A Large-Scale Benchmark for General Object Grasping}, 
  year={2020},
  volume={},
  number={},
  pages={11441-11450},
  doi={10.1109/CVPR42600.2020.01146}}

@inproceedings{okrobot, 
series={RSS2024},
  title={Demonstrating OK-Robot: What Really Matters in Integrating Open-Knowledge Models for Robotics},
  url={http://dx.doi.org/10.15607/RSS.2024.XX.091},
  DOI={10.15607/rss.2024.xx.091},
  booktitle={Robotics: Science and Systems XX},
  publisher={Robotics: Science and Systems Foundation},
  author={Liu, Peiqi and Orru, Yaswanth and Vakil, Jay and Paxton, Chris and Shafiullah, Nur and Pinto, Lerrel},
  year={2024},
  month=jul, collection={RSS2024}
}

@misc{pymeshlab,
  author       = {Alessandro Muntoni and Paolo Cignoni},
  title        = {{PyMeshLab}},
  year         = {2021},
  publisher    = {Zenodo},
  howpublished = {\url{https://doi.org/10.5281/zenodo.4438750}},
  note         = {DOI: 10.5281/zenodo.4438750}
}

@article{10.1145/3528223.3530152,
author = {Portaneri, C\'{e}dric and Rouxel-Labb\'{e}, Mael and Hemmer, Michael and Cohen-Steiner, David and Alliez, Pierre},
title = {Alpha wrapping with an offset},
year = {2022},
issue_date = {July 2022},
publisher = {Association for Computing Machinery},
address = {New York, NY, USA},
volume = {41},
number = {4},
issn = {0730-0301},
url = {https://doi.org/10.1145/3528223.3530152},
doi = {10.1145/3528223.3530152},
journal = {ACM Trans. Graph.},
month = jul,
articleno = {127},
numpages = {22}
}

@misc{carion2026sam3segmentconcepts,
      title={SAM 3: Segment Anything with Concepts}, 
      author={Nicolas Carion and Laura Gustafson and Yuan-Ting Hu and Shoubhik Debnath and Ronghang Hu and Didac Suris and Chaitanya Ryali and Kalyan Vasudev Alwala and Haitham Khedr and Andrew Huang and Jie Lei and Tengyu Ma and Baishan Guo and Arpit Kalla and Markus Marks and Joseph Greer and Meng Wang and Peize Sun and Roman Rädle and Triantafyllos Afouras and Effrosyni Mavroudi and Katherine Xu and Tsung-Han Wu and Yu Zhou and Liliane Momeni and Rishi Hazra and Shuangrui Ding and Sagar Vaze and Francois Porcher and Feng Li and Siyuan Li and Aishwarya Kamath and Ho Kei Cheng and Piotr Dollár and Nikhila Ravi and Kate Saenko and Pengchuan Zhang and Christoph Feichtenhofer},
      year={2026},
      eprint={2511.16719},
      archivePrefix={arXiv},
      primaryClass={cs.CV},
      url={https://arxiv.org/abs/2511.16719}, 
}

@misc{wuji2026retargeting,
  title={WujiHand Retargeting},
  author={Guanqi He and Wentao Zhang},
  year={2026},
  howpublished={\url{https://github.com/wuji-technology/wuji-retargeting}},
  note={Accessed: 2026-06-15}
}

@misc{wujihand,
  title={{WUJI Hand}},
  author={{WUJI Technology}},
  year={2026},
  howpublished={\url{https://docs.wuji.tech/docs/en/wuji-hand/v1/}},
  note={Accessed: 2026-06-15}
}

@misc{cui2026contactanchoredpoliciescontactconditioning,
      title={Contact-Anchored Policies: Contact Conditioning Creates Strong Robot Utility Models}, 
      author={Zichen Jeff Cui and Omar Rayyan and Haritheja Etukuru and Bowen Tan and Zavier Andrianarivo and Zicheng Teng and Yihang Zhou and Krish Mehta and Nicholas Wojno and Kevin Yuanbo Wu and Manan H Anjaria and Ziyuan Wu and Manrong Mao and Guangxun Zhang and Binit Shah and Yejin Kim and Soumith Chintala and Lerrel Pinto and Nur Muhammad Mahi Shafiullah},
      year={2026},
      eprint={2602.09017},
      archivePrefix={arXiv},
      primaryClass={cs.RO},
      url={https://arxiv.org/abs/2602.09017}, 
}

@misc{anjaria2026yormobilemanipulatorgeneralizable,
      title={YOR: Your Own Mobile Manipulator for Generalizable Robotics},
      author={Manan H Anjaria and Mehmet Enes Erciyes and Vedant Ghatnekar and Neha Navarkar and Haritheja Etukuru and Xiaole Jiang and Kanad Patel and Dhawal Kabra and Nicholas Wojno and Radhika Ajay Prayage and Soumith Chintala and Lerrel Pinto and Nur Muhammad Mahi Shafiullah and Zichen Jeff Cui},
      year={2026},
      eprint={2602.11150},
      archivePrefix={arXiv},
      primaryClass={cs.RO},
      url={https://arxiv.org/abs/2602.11150},
}

@inproceedings{peebles2023dit,
  title={Scalable Diffusion Models with Transformers},
  author={Peebles, William and Xie, Saining},
  booktitle={IEEE/CVF International Conference on Computer Vision (ICCV)},
  pages={4195--4205},
  year={2023}
}

@inproceedings{darcet2024registers,
  title={Vision Transformers Need Registers},
  author={Darcet, Timoth{\'e}e and Oquab, Maxime and Mairal, Julien and Bojanowski, Piotr},
  booktitle={International Conference on Learning Representations (ICLR)},
  year={2024}
}

@inproceedings{todorov2012mujoco,
  title={{MuJoCo}: A Physics Engine for Model-Based Control},
  author={Todorov, Emanuel and Erez, Tom and Tassa, Yuval},
  booktitle={IEEE/RSJ International Conference on Intelligent Robots and Systems (IROS)},
  pages={5026--5033},
  year={2012}
}

@inproceedings{chao2021dexycb,
  title={{DexYCB}: A Benchmark for Capturing Hand Grasping of Objects},
  author={Chao, Yu-Wei and Yang, Wei and Xiang, Yu and Molchanov, Pavlo and Handa, Ankur and Tremblay, Jonathan and Narang, Yashraj S. and Van Wyk, Karl and Iqbal, Umar and Birchfield, Stan and Kautz, Jan and Fox, Dieter},
  booktitle={IEEE/CVF Conference on Computer Vision and Pattern Recognition (CVPR)},
  pages={9044--9053},
  year={2021}
}

@inproceedings{qian2022pointnext,
  title={{PointNeXt}: Revisiting {PointNet++} with Improved Training and Scaling Strategies},
  author={Qian, Guocheng and Li, Yuchen and Peng, Houwen and Mai, Jinjie and Hammoud, Hasan and Elhoseiny, Mohamed and Ghanem, Bernard},
  booktitle={Advances in Neural Information Processing Systems (NeurIPS)},
  year={2022}
}

@inproceedings{tancik2020fourier,
  title={Fourier Features Let Networks Learn High Frequency Functions in Low Dimensional Domains},
  author={Tancik, Matthew and Srinivasan, Pratul P. and Mildenhall, Ben and Fridovich-Keil, Sara and Raghavan, Nithin and Singhal, Utkarsh and Ramamoorthi, Ravi and Barron, Jonathan T. and Ng, Ren},
  booktitle={Advances in Neural Information Processing Systems (NeurIPS)},
  year={2020}
}

@misc{sam3dteam2026sam3d3dfyimages,
      title={SAM 3D: 3Dfy Anything in Images}, 
      author={SAM 3D Team and Xingyu Chen and Fu-Jen Chu and Pierre Gleize and Kevin J Liang and Alexander Sax and Hao Tang and Weiyao Wang and Michelle Guo and Thibaut Hardin and Xiang Li and Aohan Lin and Jiawei Liu and Ziqi Ma and Anushka Sagar and Bowen Song and Xiaodong Wang and Jianing Yang and Bowen Zhang and Piotr Dollár and Georgia Gkioxari and Matt Feiszli and Jitendra Malik},
      year={2026},
      eprint={2511.16624},
      archivePrefix={arXiv},
      primaryClass={cs.CV},
      url={https://arxiv.org/abs/2511.16624}, 
}

@inproceedings{shaw2023leap,
  title={{LEAP Hand}: Low-Cost, Efficient, and Anthropomorphic Hand for Robot Learning},
  author={Shaw, Kenneth and Agarwal, Ananye and Pathak, Deepak},
  booktitle={Robotics: Science and Systems (RSS)},
  year={2023}
}

@inproceedings{qin2023anyteleop,
  title={{AnyTeleop}: A General Vision-Based Dexterous Robot Arm-Hand Teleoperation System},
  author={Qin, Yuzhe and Yang, Wei and Huang, Binghao and Van Wyk, Karl and Su, Hao and Wang, Xiaolong and Chao, Yu-Wei and Fox, Dieter},
  booktitle={Robotics: Science and Systems (RSS)},
  year={2023}
}

@misc{chavandafle2022simultaneousobjectreconstructiongrasp,
      title={Simultaneous Object Reconstruction and Grasp Prediction using a Camera-centric Object Shell Representation}, 
      author={Nikhil Chavan-Dafle and Sergiy Popovych and Shubham Agrawal and Daniel D. Lee and Volkan Isler},
      year={2022},
      eprint={2109.06837},
      archivePrefix={arXiv},
      primaryClass={cs.RO},
      url={https://arxiv.org/abs/2109.06837}, 
}

@misc{min2025stextsuperscript2mtextsuperscript2scalablestereomatching,
      title={{S\textsuperscript{2}M\textsuperscript{2}}: Scalable Stereo Matching Model for Reliable Depth Estimation}, 
      author={Junhong Min and Youngpil Jeon and Jimin Kim and Minyong Choi},
      year={2025},
      eprint={2507.13229},
      archivePrefix={arXiv},
      primaryClass={cs.CV},
      url={https://arxiv.org/abs/2507.13229}, 
}

@misc{yang2022oakinklargescaleknowledgerepository,
      title={OakInk: A Large-scale Knowledge Repository for Understanding Hand-Object Interaction}, 
      author={Lixin Yang and Kailin Li and Xinyu Zhan and Fei Wu and Anran Xu and Liu Liu and Cewu Lu},
      year={2022},
      eprint={2203.15709},
      archivePrefix={arXiv},
      primaryClass={cs.CV},
      url={https://arxiv.org/abs/2203.15709}, 
}

@misc{mandi2025dexmachina,
  title={{DexMachina}: Functional Retargeting for Bimanual Dexterous Manipulation},
  author={Mandi, Zhao and Hou, Yifan and Fox, Dieter and Narang, Yashraj and Mandlekar, Ajay and Song, Shuran},
  year={2025},
  eprint={2505.24853},
  archivePrefix={arXiv},
  primaryClass={cs.RO},
  url={https://arxiv.org/abs/2505.24853}
}

@misc{abilityhand,
  title={{Ability Hand}},
  author={{Psyonic}},
  howpublished={\url{https://www.psyonic.io/ability-hand}},
  year={2026},
  note={Accessed: 2026-06-15}
}

@misc{neroarm,
  title={{NERO}},
  author={{AgileX Robotics}},
  howpublished={\url{https://global.agilex.ai/products/nero}},
  year={2026},
  note={Accessed: 2026-06-15}
}

@misc{xarm,
  title={{xArm 7}},
  author={{UFACTORY}},
  howpublished={\url{https://www.ufactory.us/product/ufactory-xarm-7}},
  year={2026},
  note={Accessed: 2026-06-15}
}

@misc{ding2024bunnyvisionpro,
  title={{Bunny-VisionPro}: Real-Time Bimanual Dexterous Teleoperation for Imitation Learning},
  author={Ding, Runyu and Qin, Yuzhe and Zhu, Jiyue and Jia, Chengzhe and Yang, Shiqi and Yang, Ruihan and Qi, Xiaojuan and Wang, Xiaolong},
  year={2024},
  eprint={2407.03162},
  archivePrefix={arXiv},
  primaryClass={cs.RO},
  url={https://arxiv.org/abs/2407.03162}
}

@article{gupta2026grasp,
  title={Grasp to act: Dexterous grasping for tool use in dynamic settings},
  author={Gupta, Harsh and Mirzaee, Mohammad Amin and Yuan, Wenzhen},
  journal={IEEE Robotics and Automation Letters},
  volume={11},
  number={5},
  pages={6288--6295},
  year={2026},
  publisher={IEEE}
}

@inproceedings{sundermeyer2021contact,
  title={Contact-graspnet: Efficient 6-dof grasp generation in cluttered scenes},
  author={Sundermeyer, Martin and Mousavian, Arsalan and Triebel, Rudolph and Fox, Dieter},
  booktitle={2021 IEEE international conference on robotics and automation (ICRA)},
  pages={13438--13444},
  year={2021},
  organization={IEEE}
}

@inproceedings{mousavian20196,
  title={6-dof graspnet: Variational grasp generation for object manipulation},
  author={Mousavian, Arsalan and Eppner, Clemens and Fox, Dieter},
  booktitle={Proceedings of the IEEE/CVF international conference on computer vision},
  pages={2901--2910},
  year={2019}
}

@misc{singh2025dextrahrgbvisuomotorpoliciesgrasp,
      title={DextrAH-RGB: Visuomotor Policies to Grasp Anything with Dexterous Hands}, 
      author={Ritvik Singh and Arthur Allshire and Ankur Handa and Nathan Ratliff and Karl Van Wyk},
      year={2025},
      eprint={2412.01791},
      archivePrefix={arXiv},
      primaryClass={cs.RO},
      url={https://arxiv.org/abs/2412.01791}, 
}

@inproceedings{lum2024dextrahg,
title     = {Dextr{AH}-G: Pixels-to-Action Dexterous Arm-Hand Grasping with Geometric Fabrics},
author    = {Tyler Ga Wei Lum and Martin Matak and Viktor Makoviychuk and Ankur Handa and Arthur Allshire and Tucker Hermans and Nathan D. Ratliff and Karl Van Wyk},
booktitle = {8th Annual Conference on Robot Learning},
year      = {2024},
url       = {https://openreview.net/forum?id=S2Jwb0i7HN}

}

@inproceedings{christen2022dgrasp,
  title={D-grasp: Physically plausible dynamic grasp synthesis for hand-object interactions},
  author={Christen, Sammy and Kocabas, Muhammed and Aksan, Emre and Hwangbo, Jemin and Song, Jie and Hilliges, Otmar},
  booktitle={Proceedings of the IEEE/CVF Conference on Computer Vision and Pattern Recognition},
  pages={20577--20586},
  year={2022}
}

@inproceedings{wan2023unidexgrasp++,
  title={Unidexgrasp++: Improving dexterous grasping policy learning via geometry-aware curriculum and iterative generalist-specialist learning},
  author={Wan, Weikang and Geng, Haoran and Liu, Yun and Shan, Zikang and Yang, Yaodong and Yi, Li and Wang, He},
  booktitle={Proceedings of the IEEE/CVF International Conference on Computer Vision},
  pages={3891--3902},
  year={2023}
}

@inproceedings{etukuru2025robot,
  title={Robot utility models: General policies for zero-shot deployment in new environments},
  author={Etukuru, Haritheja and Naka, Norihito and Hu, Zijin and Lee, Seungjae and Mehu, Julian and Edsinger, Aaron and Paxton, Chris and Chintala, Soumith and Pinto, Lerrel and Shafiullah, Nur Muhammad Mahi},
  booktitle={2025 IEEE International Conference on Robotics and Automation (ICRA)},
  pages={8275--8283},
  year={2025},
  organization={IEEE}
}

@article{chi2024universal,
  title={Universal manipulation interface: In-the-wild robot teaching without in-the-wild robots},
  author={Chi, Cheng and Xu, Zhenjia and Pan, Chuer and Cousineau, Eric and Burchfiel, Benjamin and Feng, Siyuan and Tedrake, Russ and Song, Shuran},
  journal={arXiv preprint arXiv:2402.10329},
  year={2024}
}

@article{xu2025dexumi,
  title={Dexumi: Using human hand as the universal manipulation interface for dexterous manipulation},
  author={Xu, Mengda and Zhang, Han and Hou, Yifan and Xu, Zhenjia and Fan, Linxi and Veloso, Manuela and Song, Shuran},
  journal={arXiv preprint arXiv:2505.21864},
  year={2025}
}

@article{tao2025dexwild,
  title={Dexwild: Dexterous human interactions for in-the-wild robot policies},
  author={Tao, Tony and Srirama, Mohan Kumar and Liu, Jason Jingzhou and Shaw, Kenneth and Pathak, Deepak},
  journal={arXiv preprint arXiv:2505.07813},
  year={2025}
}

@article{gao2026dreamdojo,
  title={DreamDojo: A Generalist Robot World Model from Large-Scale Human Videos},
  author={Gao, Shenyuan and Liang, William and Zheng, Kaiyuan and Malik, Ayaan and Ye, Seonghyeon and Yu, Sihyun and Tseng, Wei-Cheng and Dong, Yuzhu and Mo, Kaichun and Lin, Chen-Hsuan and others},
  journal={arXiv preprint arXiv:2602.06949},
  year={2026}
}

@misc{projectaria_gen2_mps_benchmarks,
  title        = {Project Aria Gen 2 MPS Performance Benchmarks},
  author       = {{Meta Reality Labs Research}},
  howpublished = {\url{https://facebookresearch.github.io/projectaria_tools/gen2/technical-specs/mps/benchmarks/performance}},
  year         = {2025},
  note         = {Accessed: 2026-06-14},
}

%===============================================================================

\clearpage
\appendix
\begin{center}
  {\Large\bfseries Appendix of ``Human Universal Grasping''}
\end{center}

This appendix provides additional results, analyses, and implementation details that support the main paper. We include more details about the MANO parameter optimization in \S~\ref{sec:mano_optimization}, the \dataset curation with \texttt{aria2mano} in \S~\ref{sec:curation_pipeline}, the \method model implementation details and hyperparameters in \S~\ref{sec:model_details}, the \benchmark asset generation with \texttt{aria2mesh} in \S~\ref{sec:benchmark}, the simulation grasping evaluation in \S~\ref{sec:simulation_results}, and the real-world grasping evaluation in \S~\ref{sec:real_results}.

\startcontents[appendix]
\printcontents[appendix]{}{1}{\setcounter{tocdepth}{2}}
\newpage

\section{MANO Parameter Optimization}
\label{sec:mano_optimization}

\begin{wrapfigure}{r}{0.5\linewidth}
    \centering
    \includegraphics[width=\linewidth]{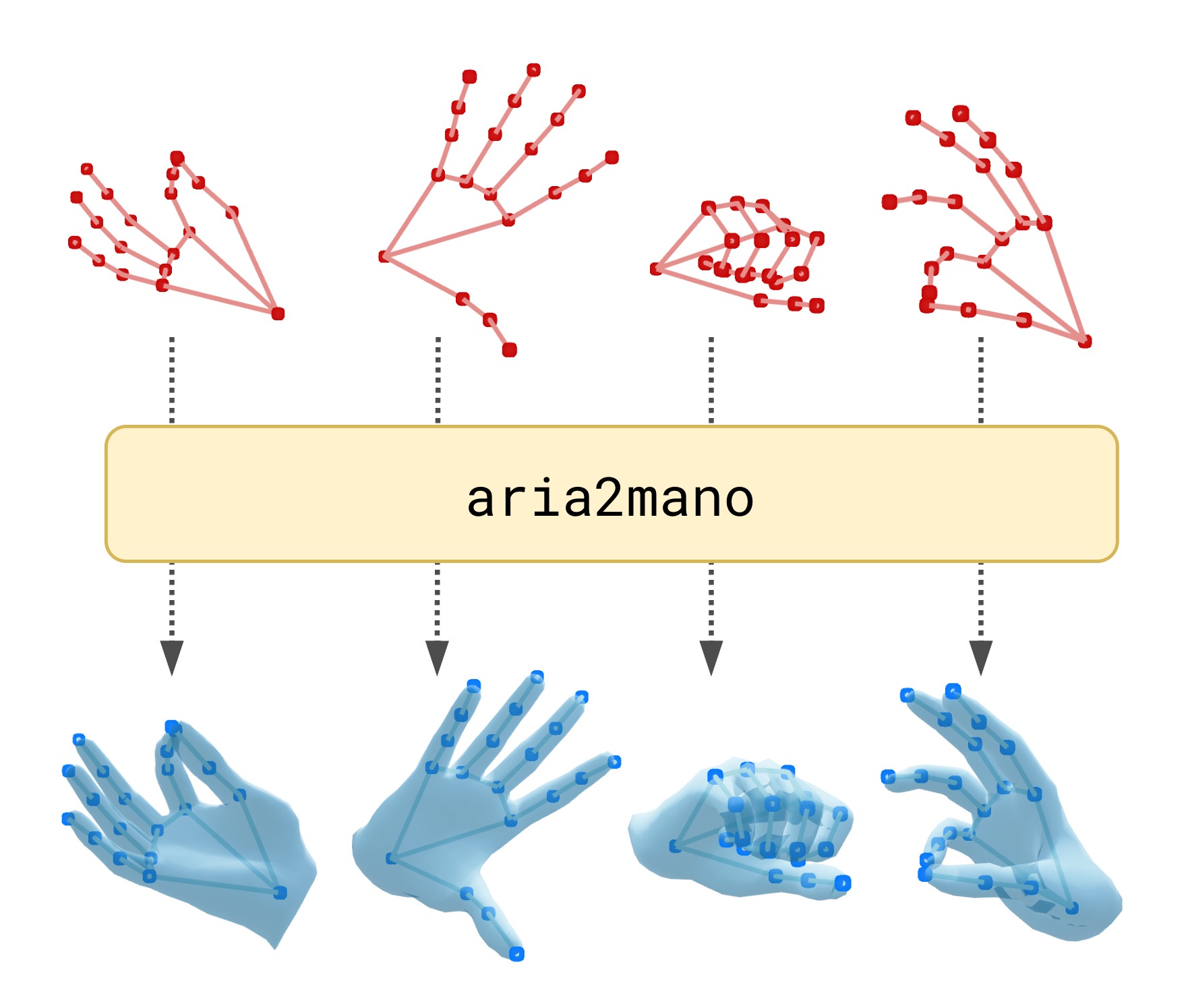}
    \caption{\textbf{\texttt{aria2mano} MANO fitting.} \texttt{aria2mano} fits a full articulated MANO hand (bottom, blue) to the sparse 21-landmark Aria skeleton (top, red) via a per-frame anatomically-constrained optimization, recovering pose, mesh, and dense joint angles from each grasp recording.}
    \label{fig:aria2mano}
\end{wrapfigure}

We elaborate on the MANO~\cite{romero2017mano} fitting procedure summarized in \S~\ref{sec:dataset}. Aria Gen 2 reports the wearer's hand only as a sparse 21-landmark skeleton in the device coordinate frame; to obtain the full articulated MANO representation (pose, shape, mesh), we run a per-frame optimization that aligns the MANO joint positions with the Aria landmarks subject to anatomical constraints. Figure~\ref{fig:aria2mano} illustrates the \texttt{aria2mano} fitting. This fitting procedure is released in our \texttt{aria2mano} package.

\paragraph{Optimized parameters.} We optimize the 15 finger-joint axis-angle parameters $\boldsymbol{\theta} \in \mathbb{R}^{15 \times 3}$ (45\,DoF in total). The wrist orientation and translation are fixed to the values reported by Aria, since Aria's wrist estimate is already metric and globally consistent across frames. We also optimize the 10-dimensional MANO shape $\boldsymbol{\beta} \in \mathbb{R}^{10}$ (later replaced by a fixed shape, \S~\ref{sec:fixed_shape}) for consistency with the \dataset and \benchmark.

\paragraph{Landmark alignment loss.} Aria landmarks are first reordered to match the MANO joint convention, and the wrist (fixed) and the carpometacarpal (CMC) joint of the thumb (which Aria does not report reliably) are masked out of the loss. We then minimize a weighted MSE between the MANO-reconstructed joint positions $\hat{\mathbf{x}}_i$ and the Aria landmarks $\mathbf{x}_i$:
\begin{equation}
\mathcal{L}_{\text{lm}} = \frac{\lambda_{\text{mse}}}{\sum_i w_i} \sum_{i=1}^{21} w_i \, \lVert \hat{\mathbf{x}}_i - \mathbf{x}_i \rVert_2^2,
\end{equation}
where the five fingertip landmarks receive a weight $w_i = 5$ and all other landmarks receive $w_i = 1$, and $\lambda_{\text{mse}} = 2{\times}10^4$ balances the landmark term against the anatomical prior. The fingertip up-weighting is essential because fingertips dominate downstream contact geometry, and small angular errors at proximal joints accumulate into large fingertip displacements.

\paragraph{Anatomical prior.} In addition to our landmark loss, we attach the anatomical validity loss $\mathcal{L}_{\text{anat}}$ from \texttt{manotorch}'s \texttt{AnatomyConstraintLossEE}~\cite{yang2021cpf}, which extracts per-joint Euler angles via forward kinematics on the MANO skeleton and penalizes deviations from human range-of-motion bounds. This prevents the optimizer from explaining landmarks with hyperextended or fully rotated finger poses.

\paragraph{Optimizer.} We minimize $\mathcal{L} = \mathcal{L}_{\text{lm}} + \mathcal{L}_{\text{anat}}$ with L-BFGS using strong-Wolfe line search, history size 10, gradient-norm tolerance $10^{-7}$, and gradient-change tolerance $10^{-3}$. To accelerate convergence we warm-start each frame from the optimized pose of the previous frame whenever temporal continuity is available, with a closure-call budget of 20; cold-started frames are capped at 50. Wrist and CMC entries are excluded from the loss by construction. We achieve an average fingertip error of less than 2 mm.

\section{\dataset Dataset Curation with \texttt{aria2mano}}
\label{sec:curation_pipeline}

\begin{figure}[t]
    \centering
    \includegraphics[width=\linewidth]{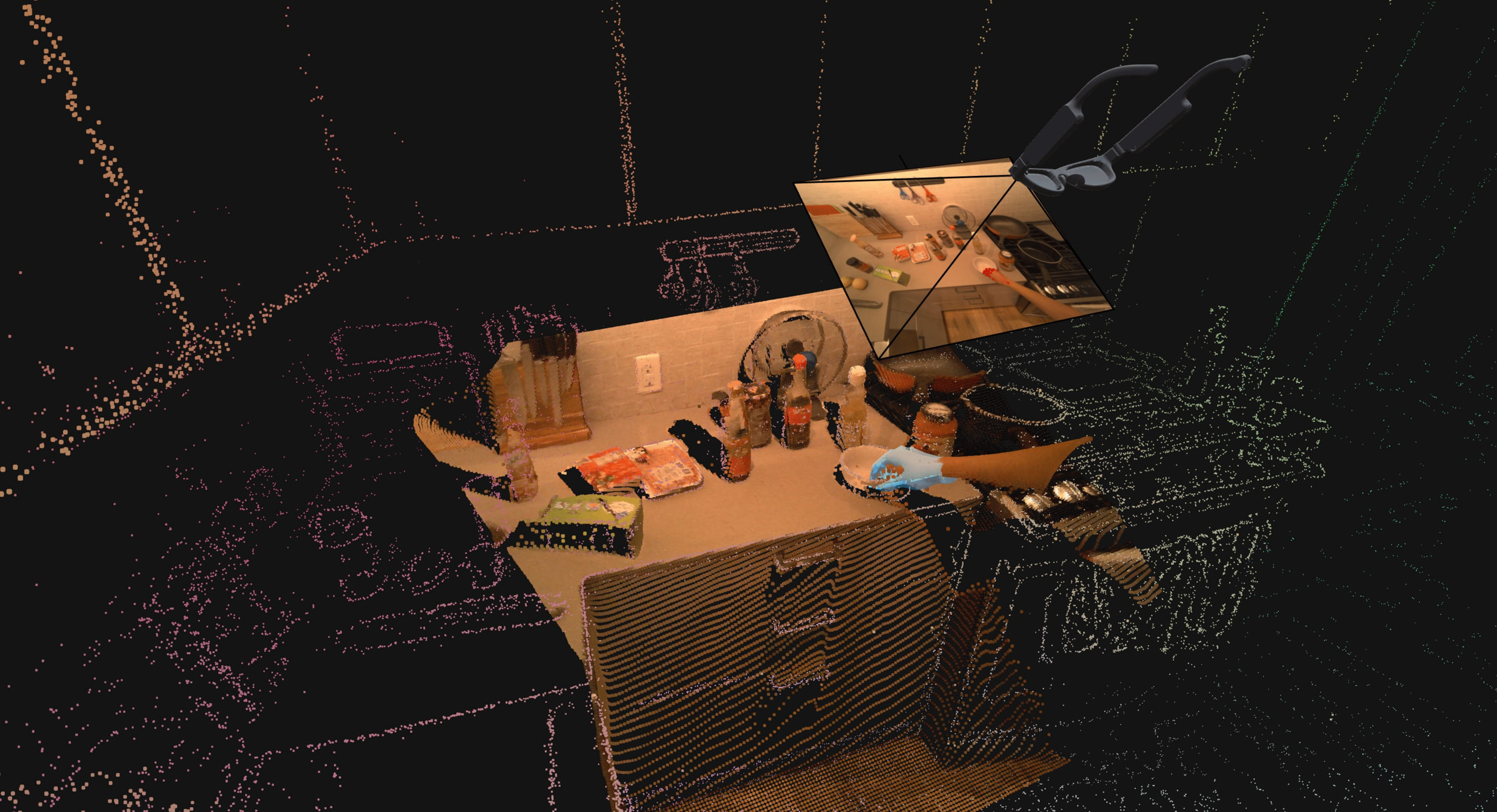}
    \caption{\textbf{Data collection.} A wearer captures a grasp with Aria Gen 2 glasses.}
    \label{fig:data_collection}
\end{figure}

This section details the curation pipeline that turns raw Aria Gen 2 recordings into the \dataset grasp dataset, summarized in \S~\ref{sec:dataset}. Each recording captures a single right-hand grasp of a single stationary object: the wearer moves their head around the object for $15$--$30$ seconds with both hands behind their back, then grasps the object without lifting it. From this footage we identify the grasped object, segment it across all frames, select the grasp frame, manually verify the result, and prepare the final training entries. The full pipeline is released as part of \texttt{aria2mano}. Figure~\ref{fig:data_collection} illustrates the capture setup.

\subsection{Automated Labeling}
\label{sec:auto_processing}

The first three stages run without human input: identifying the grasped object, segmenting it across the recording, and selecting the grasp frame.

\paragraph{Object Identification.}
\label{sec:vlm_object}
We localize the grasped object with a vision-language model (we used Gemini 3 Flash Preview). The VLM compares one ``before'' frame with no hand visible against four ``after'' frames in which the hand grasps the object, and returns $5$--$10$ points covering all parts of the object, including its body and any handles or appendages. A second VLM call verifies that each point lies on the grasped object and rejects the recording if at most one point does. The cost is roughly \$0.003 per recording.

\paragraph{Mask Propagation.}
\label{sec:sam3_mask}
The verified points prompt SAM3~\cite{carion2026sam3segmentconcepts} to segment the object and track it bidirectionally from the annotation frame across every frame of the recording, yielding a per-frame binary object mask.

\paragraph{Grasp-Frame Selection.}
\label{sec:grasp_frame}
We detect the grasp frame with a set of heuristics: the frame must have valid right-hand MANO tracking with all landmarks projecting in bounds, hand motion below $0.01$\,m per frame (stability), mean fingertip-to-mask distance below $100$\,px (proximity), and must fall within the last $10$\,seconds of the recording. Among the surviving candidates we select the frame of best MANO quality, measured as the geometric mean of fingertip error and anatomy loss, preferring frames with no visible left hand.

\subsection{Quality Control}
\label{sec:annotation_app}

\begin{figure}[t]
    \centering
    \includegraphics[width=\linewidth]{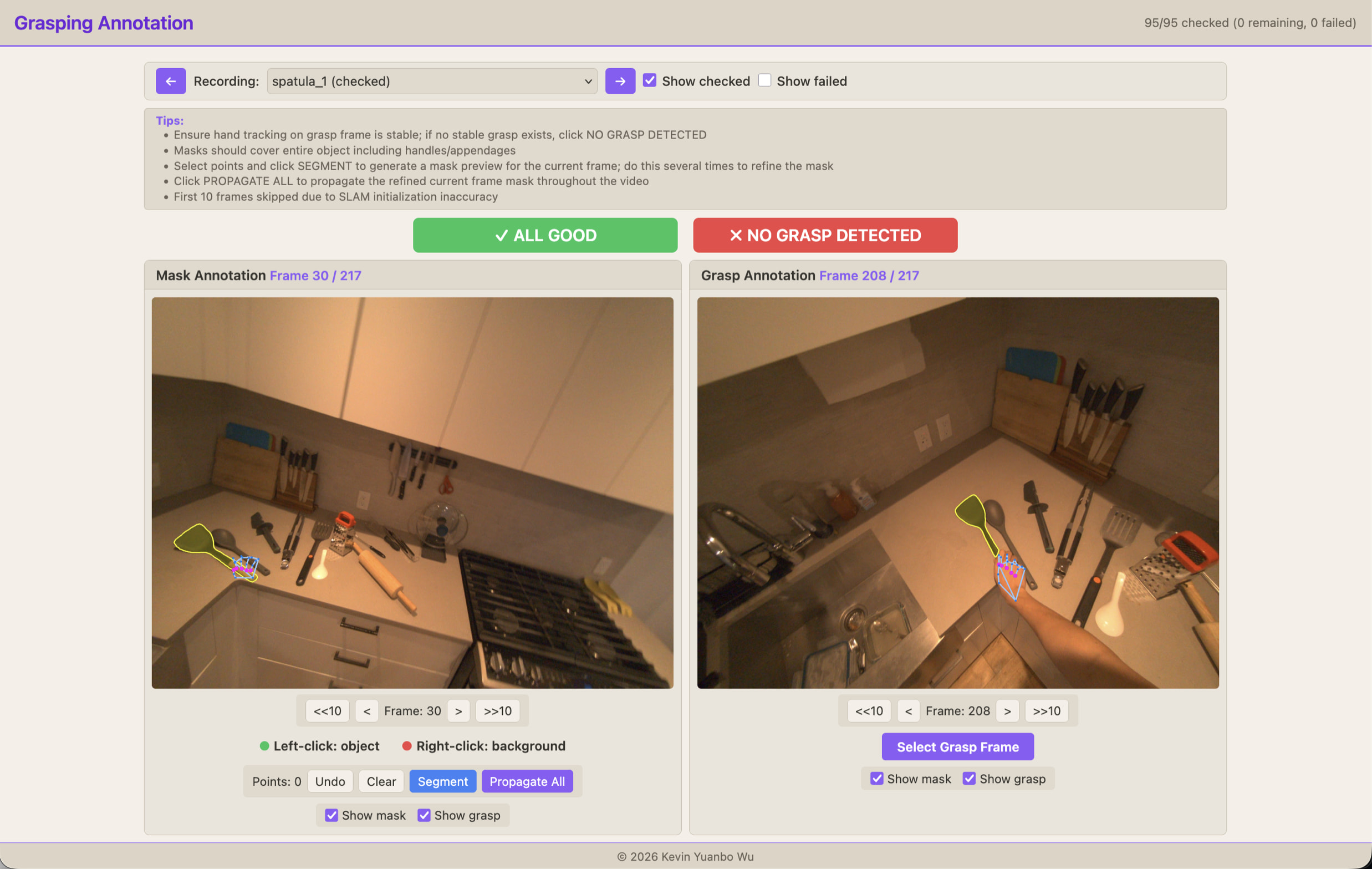}
    \caption{\textbf{Grasp annotation app.} The web interface used to verify and correct the automatic labels. The left panel refines the object mask with point prompts and SAM3 re-segmentation; the right panel steps through frames to select and verify the grasp frame. Each recording is marked checked once its mask and grasp pass review.}
    \label{fig:annotation_app}
\end{figure}

Automatic labeling is not 100\% accurate, so every recording is reviewed in a web annotation app (Figure~\ref{fig:annotation_app}). A reviewer can re-segment the mask with additional point prompts and re-propagate it, re-select the grasp frame, and either approve the recording or mark it as having no stable grasp. Approval writes a checked flag so the recording enters dataset preparation.

\subsection{Dataset Preparation}
\label{sec:dataset_prep}

\begin{wrapfigure}{r}{0.25\linewidth}
    \centering
    {\setlength{\fboxsep}{0pt}\setlength{\fboxrule}{0.6pt}%
    \fbox{\includegraphics[width=\dimexpr\linewidth-2\fboxrule-2\fboxsep\relax]{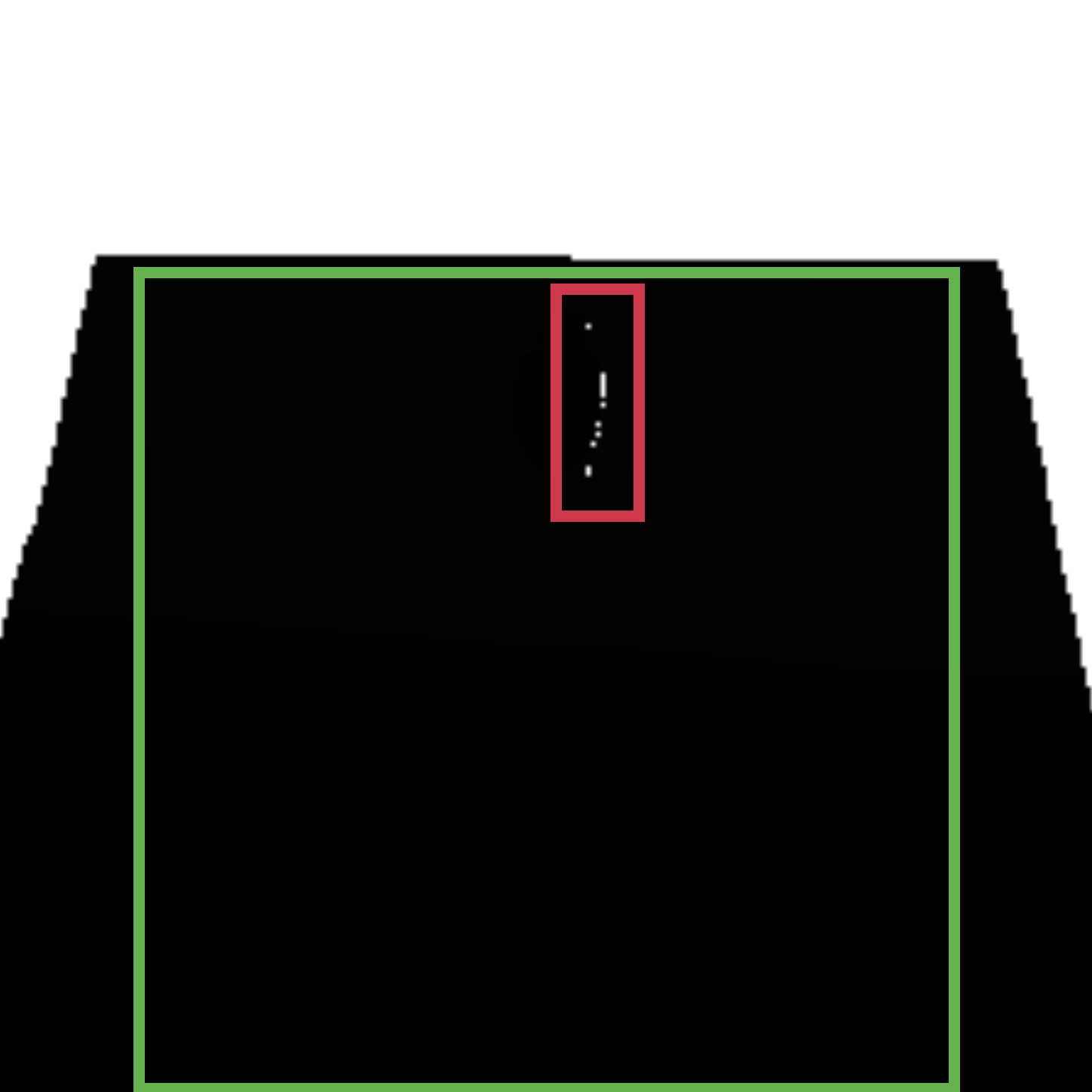}}}
    \caption{\textbf{RGB crop.} Stereo-left depth reprojected into the RGB frame: valid reprojected depth (black), the crop (green), and parallax holes filled by nearest neighbor (red).}
    \label{fig:rgb_crop}
\end{wrapfigure}

We export training entries from the checked recordings under a sequence of filters. \emph{Shared filters} keep frames whose index lies in $[20,\,\text{grasp}-10)$ (skipping the SLAM warm-up and the 10 frames before the grasp), drop the annotation frame, require at least $60\%$ valid depth-confidence pixels, and, on the center-cropped square, require a non-empty object mask with at least one grasp-hand landmark on it. \emph{Per-camera filters}, applied independently to the RGB and stereo-left grayscale streams, require at least five grasp-hand landmarks inside the image, no MPS-tracked hand projecting inside the image, and a non-empty final $224\!\times\!224$ object mask. The grasp pose is transformed from the world frame into each per-frame camera frame. The stereo-left grayscale images are included as a free augmentation, so the full dataset contains roughly $2$M entries ($1$M RGB plus $1$M grayscale). For the scaling study we build nested video-level subsets at $25$k, $50$k, $100$k, $250$k, $500$k, and $1$M RGB frames, each a strict prefix of the next.

\paragraph{RGB field-of-view crop.}
\label{sec:rgb_crop}
RGB images are cropped to fit the depth field of view, since the S2M2~\cite{min2025stextsuperscript2mtextsuperscript2scalablestereomatching} depth is reprojected from the stereo-left camera into the RGB frame and the two fields of view do not fully overlap (Figure~\ref{fig:rgb_crop}). After the center-square crop, RGB drops the top $25\%$ and equal side margins to a $1440\!\times\!1440$ window ($75\%$ of $1920$), then resizes to $224\!\times\!224$; the stored intrinsics reflect the crop. After cropping, roughly $99.9\%$ of depth is valid, and the remaining ${\sim}0.1\%$ parallax holes are nearest-neighbor filled with a distance transform, preserving sharp edges without smoothing across object boundaries.

\paragraph{Fixed MANO shape.}
\label{sec:fixed_shape}
\method predicts only hand pose, with the MANO shape $\boldsymbol{\beta}$ held fixed at a single canonical value. The data collectors have low diversity in hand size, so we set $\boldsymbol{\beta}$ to the first author's hand, which covers the data well on average and is at least as large as every hand in \benchmark (Figure~\ref{fig:hand_sizes}). We then recompute all grasp labels in \dataset and \benchmark with this fixed $\boldsymbol{\beta}$, keeping the recorded joint angles, so that a single hand size is used consistently throughout training and evaluation. For simulation we export an MJCF of this canonical hand and use it for every grasp rollout. The fixed shape can be changed and the dataset recomputed if a different hand size is desired. We detail the MJCF generation in Appendix~\ref{sec:mano_mjcf}.

\section{\method Model Implementation Details}
\label{sec:model_details}

This section provides implementation details of \method beyond the main-text method description: the point cloud crop that bounds the receptive field, how the model generalizes across cameras, the single-modality ablations and training dynamics, and the full hyperparameter list.

\subsection{Point Cloud Crop Radius}
\label{sec:crop_radius}

\begin{wrapfigure}{r}{0.35\linewidth}
    \centering
    \includegraphics[width=\linewidth]{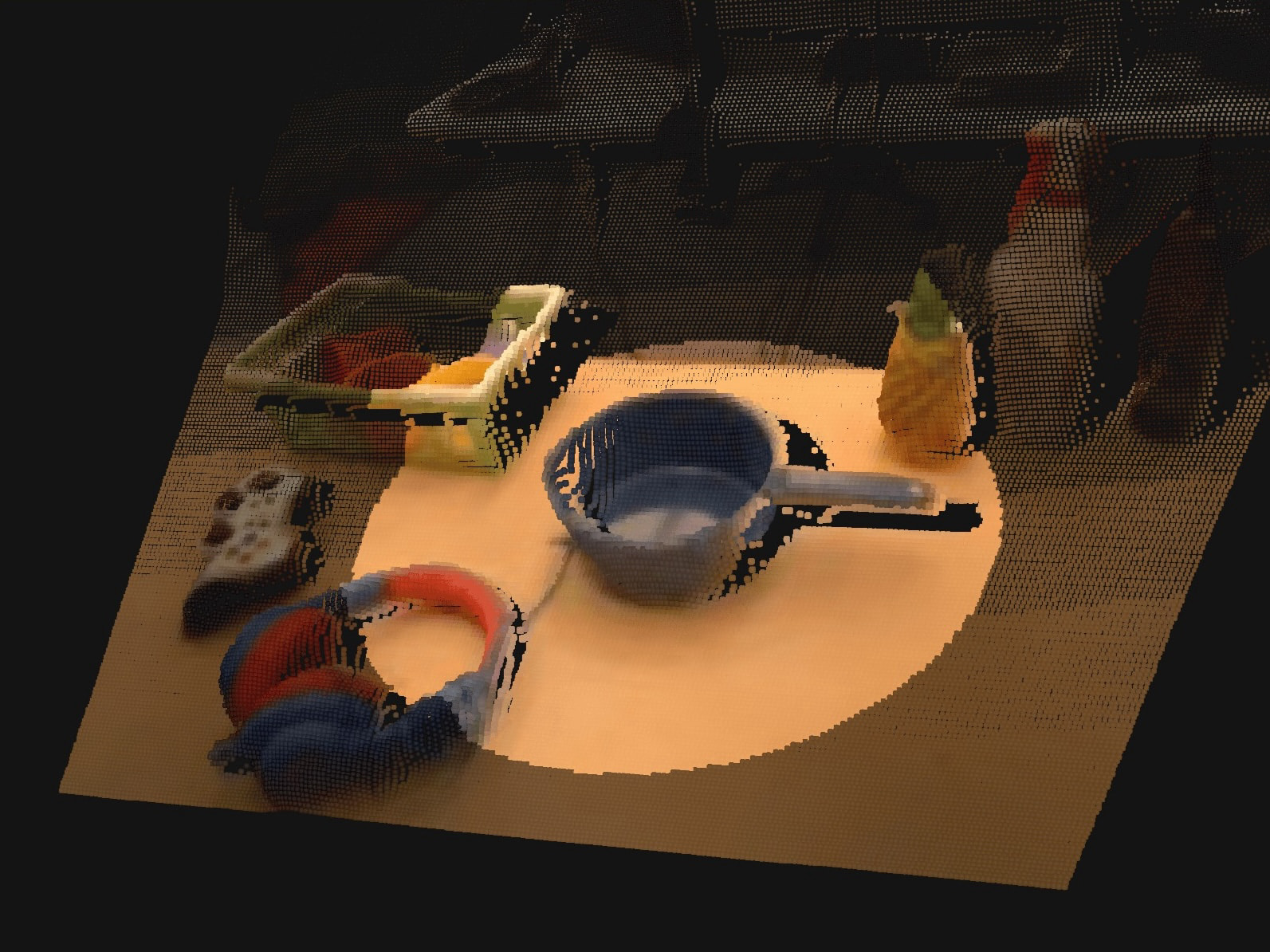}
    \caption{\textbf{Point cloud crop.} The $0.3$\,m radius crop around the 3D query point.}
    \label{fig:crop_radius}
\end{wrapfigure}

Before encoding, we crop the back-projected point cloud $\mathcal{P}$ to a ball of radius $r$ around the 3D query point $\mathbf{p}_q$,
\begin{equation}
\mathcal{P}_{\text{crop}} = \{\, \mathbf{p}\in\mathcal{P} : \lVert \mathbf{p}-\mathbf{p}_q\rVert_2 \le r \,\}, \qquad r = 0.3\,\text{m},
\end{equation}
and sample $N_p\!=\!4096$ points from $\mathcal{P}_{\text{crop}}$. The crop bounds the receptive field to $N_p$ points (hence $N\!=\!256$ tokens); over a full image these $256$ centroids would be too sparse to resolve the target, whereas the crop concentrates them on the object. Each metric centroid is encoded with Fourier features~\cite{tancik2020fourier}, which has been shown to help point cloud encoders leverage geometric details more effectively~\cite{gyenes2026fourierfeaturesletagents} than Cartesian features. The radius sets the spatial scope of the model, \ie the largest object it can grasp, so it is fixed by the intended application rather than tuned for accuracy, and we do not ablate it. We use $0.3$\,m as roughly the largest object graspable with one hand. A smaller radius would likely raise metrics on our \texttt{val} and \texttt{test} sets, which are biased toward smaller objects, but at the cost of generality; a model specializing in small objects can be retrained with a smaller radius. Figure~\ref{fig:crop_radius} visualizes the crop.

\subsection{Camera Generalization}
\label{sec:camera_generalization}

The camera intrinsics $\mathbf{K}$ enter \method only through geometric operations: back-projecting a pixel $(u,v)$ with depth $d$ to a metric point
\begin{equation}
\mathbf{p} = d\,\mathbf{K}^{-1}\,[\,u,\,v,\,1\,]^\top,
\end{equation}

which lifts the 2D click to the 3D query point $\mathbf{p}_q$, and the inverse projection $[\,u',v',1\,]^\top \propto \mathbf{K}\,\mathbf{c}_i$ that maps each centroid $\mathbf{c}_i$ onto the image plane for point painting. No layer takes $\mathbf{K}$ as a learned input. The model therefore transfers across stereo cameras with different intrinsics without retraining, which is why training on Aria recordings deploys directly to a ZED stereo camera in our tabletop experiments. We have also tested \method on Realsense D415/D435 with success despite their grayscale images.

\subsection{Training Details and Hyperparameters}
\label{sec:hyperparameters}

Table~\ref{tab:hyperparameters} lists the architecture, training, and inference hyperparameters of \method, and Table~\ref{tab:model_params} reports the per-module parameter counts. MANO-fitting hyperparameters (L-BFGS settings, landmark weights, anatomical prior) are given in Appendix~\ref{sec:mano_optimization}. Figure~\ref{fig:training_curves} shows training curves for all ablation configurations.

\begin{table}[t]
\centering
\caption{\textbf{\method full model hyperparameters.}}
\label{tab:hyperparameters}
\setlength{\tabcolsep}{6pt}
\renewcommand{\arraystretch}{1.1}
{\scriptsize
\begin{tabular}{lll}
\toprule
Group & Hyperparameter & Value \\
\midrule
\multirow{14}{*}{Architecture}
 & RGB encoder & DINOv2-Base + registers (frozen) \\
 & RGB tokens & $256$ ($16\times16$ patch grid) \\
 & Point encoder & PointNeXt U-Net (trainable) \\
 & PointNeXt width / blocks & $c=64$ ($512$D out), blocks $[1,2,1,1]$ (B) \\
 & PointNeXt SA radii & $[0.025, 0.05, 0.10, 0.20]$\,m \\
 & Per-point RGB at stem & yes (stem in-dim $6$) \\
 & Input points $N_p$ & $4096$ \\
 & Point tokens $N$ & $256$ \\
 & Fusion transformer & $4$ layers, $D_f = 1024$, $8$ heads \\
 & Flow transformer & $6$ DiT blocks, $D_m = 512$, $8$ heads \\
 & MANO target dim & $99$ ($3$ trans $+$ $6$D wrist $+$ $90$ fingers) \\
 & Fourier scale & $1.0$ \\
 & Dropout & $0.1$ \\
 & MANO shape $\boldsymbol{\beta}$ & [-2.37, -1.25, -2.05, -0.85, 1.66, -1.35, -1.85, -0.67, -1.69, -1.21] \\
\midrule
\multirow{3}{*}{Conditioning}
 & Query point & 3D, Fourier-feature encoded \\
 & Point cloud crop radius & $0.3$\,m \\
 & Fusion & Point painting (bilinear DINOv2 sampling) \\
\midrule
\multirow{14}{*}{Training}
 & Optimizer & AdamW \\
 & Base learning rate & $1\times 10^{-4}$ \\
 & PointNeXt learning rate & $4\times 10^{-4}$ ($4\times$) \\
 & Weight decay & $1\times 10^{-3}$ \\
 & Gradient clipping & $1.0$ (max norm) \\
 & Batch size & $64$ / GPU (128 effective)\\
 & Steps & $100\text{K}$ \\
 & Warmup & $5\text{K}$ steps, linear \\
 & Precision & bf16 \\
 & EMA & decay $0.9999$, from step $50\text{K}$ \\
 & Joint loss & L1 \\
 & Loss weights & $\lambda_{\text{v}} = 1$, $\lambda_{3\text{D}} = 20$ \\
 & Hardware & $2\times$ RTX 5090, DDP, ${\sim}10$\,h \\
 & Validation cadence & every $5\text{K}$ steps (MuJoCo) \\
 & ODE solver & $50$-step Euler \\
\bottomrule
\end{tabular}}
\end{table}

\begin{table}[t]
\centering
\caption{\textbf{\method full model parameter counts.}}
\label{tab:model_params}
\setlength{\tabcolsep}{8pt}
\renewcommand{\arraystretch}{1.1}
{\scriptsize
\begin{tabular}{lrr}
\toprule
Module & Total & Trainable \\
\midrule
RGB encoder (DINOv2, frozen)    & $86{,}583{,}552$  & $0$ \\
Point encoder (PointNeXt)       & $15{,}587{,}776$  & $15{,}587{,}776$ \\
Fusion transformer              & $67{,}392{,}256$  & $67{,}390{,}720$ \\
MANO layer (frozen)             & $370{,}329$       & $0$ \\
Flow transformer                & $37{,}682{,}374$  & $37{,}682{,}176$ \\
\midrule
Total                           & $207{,}244{,}224$ & $120{,}660{,}672$ \\
\bottomrule
\end{tabular}}
\end{table}

\begin{figure}[t]
    \centering
    \includegraphics[width=\linewidth]{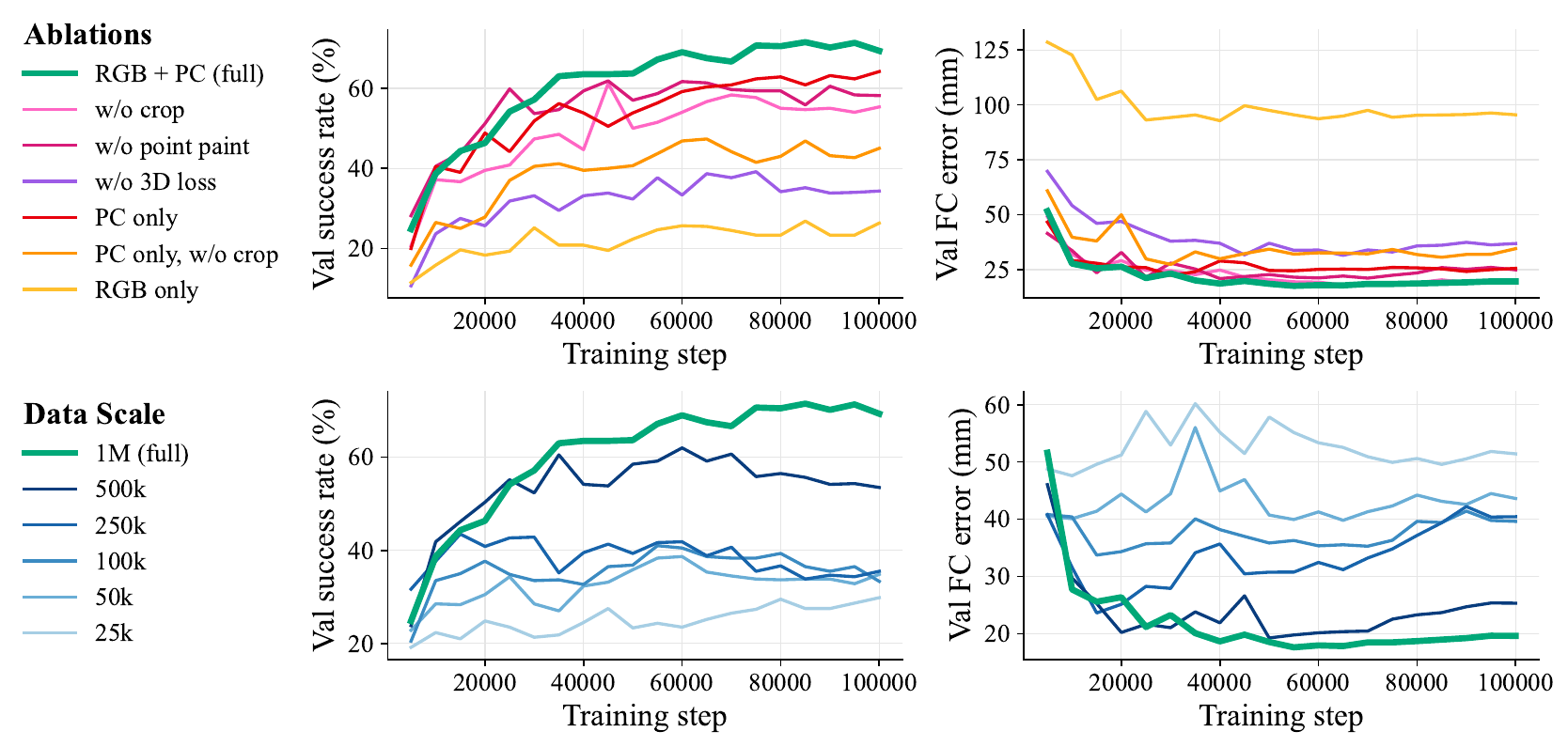}
    \caption{\textbf{Training curves.} Success rate and fingertip contact error on the \benchmark \texttt{val} split for model ablations (top) and data scaling experiments (bottom).}
    \label{fig:training_curves}
\end{figure}

\section{\benchmark Asset Generation with \texttt{aria2mesh}}
\label{sec:benchmark}

\benchmark is a simulation benchmark of $90$ everyday objects with metric-scale meshes and physics-ready simulation assets, reconstructed from Aria Gen 2 recordings with our released \texttt{aria2mesh} pipeline. This section details the asset-generation pipeline and reports per-object statistics for the full object set.

\subsection{Metric-scale 3D Object Reconstruction}
\label{sec:asset_pipeline}

\begin{figure}[t]
    \centering
    \includegraphics[width=\linewidth]{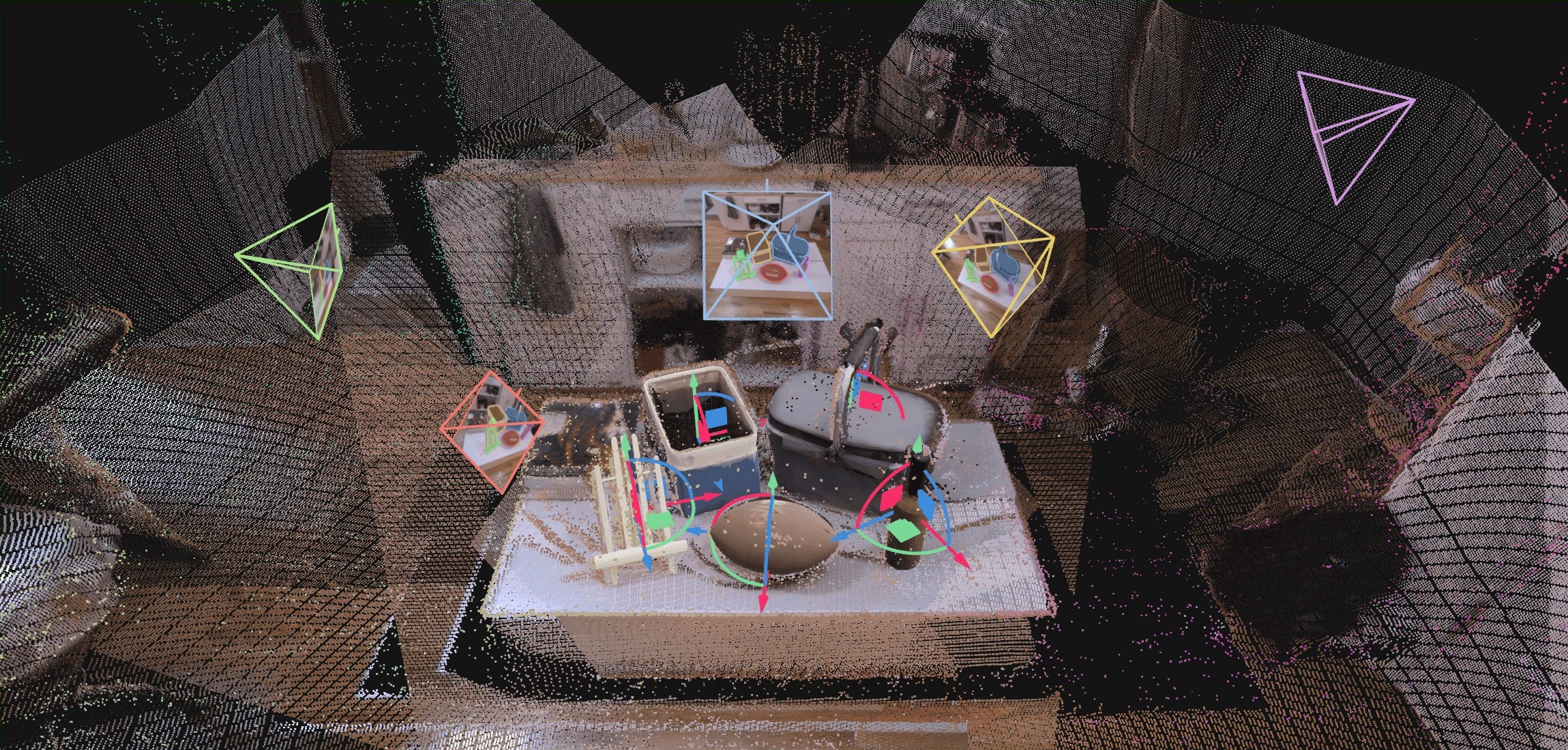}
    \caption{\textbf{\texttt{aria2mesh} asset reconstruction.} Five egocentric Aria views (camera frustums) are fused with Multi-view SAM3D, then pose-optimized and gravity-aligned, and finally manually edited against the semidense point cloud and dense stereo depth points to recover per-object metric-scale meshes and poses (axes).}
    \label{fig:aria2mesh}
\end{figure}

Each \benchmark object is reconstructed from five egocentric Aria Gen 2~\cite{aria-gen-2} views with Multi-view SAM3D~\cite{li2026mvsam3dadaptivemultiviewfusion}, a training-free extension of SAM3D~\cite{sam3dteam2026sam3d3dfyimages}, into which we inject Aria camera intrinsics, extrinsics, and stereo depth. We then pose-optimize and gravity-align each mesh, manually verify and edit its scale and pose in Viser~\cite{yi2025viserimperativewebbased3d} against the SLAM semidense point cloud and the dense stereo depth, make it watertight with Alpha Wrap~\cite{10.1145/3528223.3530152} implemented in PyMeshLab~\cite{pymeshlab}, and produce a convex decomposition with CoACD~\cite{wei2022coacd}. Figure~\ref{fig:aria2mesh} shows the multi-view reconstruction recovering metric-scale meshes and poses from the five Aria views. The pipeline exports a visual mesh, convex collision parts, and both a URDF and an MJCF per object, loadable directly in PyBullet and MuJoCo. Because reconstruction runs from a short egocentric recording, a real-world object is turned into a metric-scale, simulation-ready asset in minutes, making it practical to grow the benchmark or build task-specific object sets.

\subsection{Object Statistics}
\label{sec:object_stats}

Table~\ref{tab:object_stats} lists all $90$ objects with their size class, geometric category, mass, and volume, a thumbnail of both the reconstructed simulation mesh and the real object, and the per-object success rates of the best checkpoint: the ground-truth (human grasp oracle) simulation SR, the \method simulation SR, and the real-world tabletop and in-the-wild SR. The 30 \texttt{test} objects are listed first, followed by the \texttt{val} objects, which we only use for simulation evaluation.

\clearpage% caption + table start together on a fresh page; heading+intro stay on the previous page
\captionof{table}{\textbf{\benchmark object statistics.} All $90$ objects with category, mass, volume, a real-object and simulation-mesh thumbnail, and best-checkpoint success rates: ground-truth (human grasp oracle) sim SR, \method sim SR, \method tabletop SR, and \method in-the-wild SR. The latter two SRs are results from Table~\ref{tab:real_world_results} (Tabletop: ZED + xArm + Ability; Wild: Aria + YOR + WUJI). The \texttt{test} objects (top) carry real-world numbers; \texttt{val} objects (bottom) are simulation-only.}
\label{tab:object_stats}
{\scriptsize
\setlength{\tabcolsep}{2.5pt}% 12 columns; 3pt overflows the text width by ~4pt
\renewcommand{\arraystretch}{1.4}% strut taller than the thumbnail so every row has uniform pitch (\multirow centers exactly)
\newcommand{\othumb}[1]{\makebox[11pt][c]{%
  $\vcenter{\hbox{\includegraphics[width=11pt,height=11pt,keepaspectratio]{tables/object_thumbnails/#1}}}$}\rule[-3.3pt]{0pt}{11pt}}% fixed 11x11pt box, aspect preserved, invisible strut keeps row metrics uniform
\begin{longtable}{lllccrrlcccc}
\toprule
Geometry & Size & Object & Real & Sim & Mass (g) & Vol.\ (cm$^3$) & Scene & GT Sim SR & Sim SR & Tabletop SR & Wild SR \\
\midrule
\endfirsthead
\multicolumn{12}{l}{\itshape continued from previous page}\\
\toprule
Geometry & Size & Object & Real & Sim & Mass (g) & Vol.\ (cm$^3$) & Scene & GT Sim SR & Sim SR & Tabletop SR & Wild SR \\
\midrule
\endhead
\multicolumn{12}{r}{\itshape continued on next page}\\
\endfoot
\bottomrule
\endlastfoot
\multicolumn{12}{l}{\textbf{\benchmark \texttt{test}}} \\*
\midrule
\multirow{6}{*}{Cylindrical}
  & \multirow{2}{*}{S} & Glue stick & \othumb{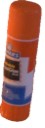} & \othumb{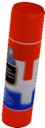} & 42.4 & 64.4 & \texttt{test/small\_1} & 10/10 & 5/10 & 6/10 & 7/10 \\*
  &                    & Pepper shaker & \othumb{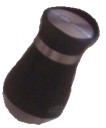} & \othumb{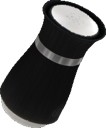} & 112.9 & 222.7 & \texttt{test/small\_2} & 10/10 & 10/10 & 10/10 & 6/10 \\*
  \cdashline{2-12}[.4pt/1.4pt]
  & \multirow{2}{*}{M} & Umbrella & \othumb{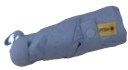} & \othumb{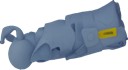} & 208.1 & 136.9 & \texttt{test/medium\_1} & 10/10 & 10/10 & 5/10 & 6/10 \\*
  &                    & Bowl & \othumb{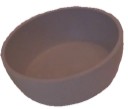} & \othumb{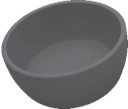} & 158.6 & 226.9 & \texttt{test/medium\_2} & 9/10 & 8/10 & 4/10 & 1/10 \\*
  \cdashline{2-12}[.4pt/1.4pt]
  & \multirow{2}{*}{L} & Spray bottle & \othumb{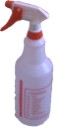} & \othumb{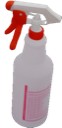} & 450.8 & 1216.1 & \texttt{test/large\_1} & 9/10 & 10/10 & 9/10 & 9/10 \\*
  &                    & Wine bottle & \othumb{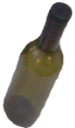} & \othumb{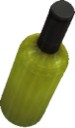} & 317.6 & 1026.1 & \texttt{test/large\_2} & 10/10 & 10/10 & 3/10 & 10/10 \\
\midrule
\multirow{6}{*}{Spheroidal}
  & \multirow{2}{*}{S} & Strawberry & \othumb{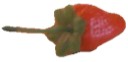} & \othumb{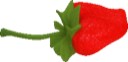} & 3.1 & 13.4 & \texttt{test/small\_1} & 10/10 & 4/10 & 7/10 & 6/10 \\*
  &                    & Hacky sack & \othumb{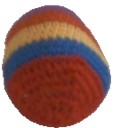} & \othumb{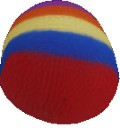} & 34.6 & 9.7 & \texttt{test/small\_2} & 10/10 & 10/10 & 9/10 & 10/10 \\*
  \cdashline{2-12}[.4pt/1.4pt]
  & \multirow{2}{*}{M} & Pear & \othumb{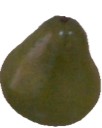} & \othumb{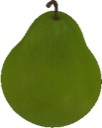} & 16.9 & 246.3 & \texttt{test/medium\_1} & 10/10 & 10/10 & 10/10 & 10/10 \\*
  &                    & Softball & \othumb{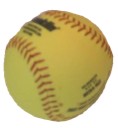} & \othumb{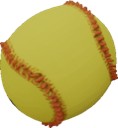} & 202.2 & 539.1 & \texttt{test/medium\_2} & 10/10 & 10/10 & 5/10 & 8/10 \\*
  \cdashline{2-12}[.4pt/1.4pt]
  & \multirow{2}{*}{L} & Pineapple & \othumb{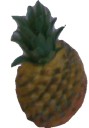} & \othumb{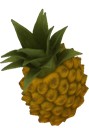} & 153.6 & 181.6 & \texttt{test/large\_1} & 10/10 & 10/10 & 10/10 & 9/10 \\*
  &                    & Football & \othumb{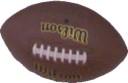} & \othumb{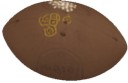} & 421.4 & 3371.4 & \texttt{test/large\_2} & 10/10 & 8/10 & 0/10 & 0/10 \\
\midrule
\multirow{6}{*}{Prismatic}
  & \multirow{2}{*}{S} & Eraser & \othumb{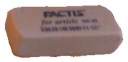} & \othumb{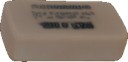} & 24.7 & 30.9 & \texttt{test/small\_1} & 7/10 & 6/10 & 6/10 & 6/10 \\*
  &                    & Match box & \othumb{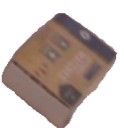} & \othumb{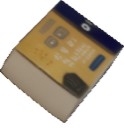} & 10.0 & 40.6 & \texttt{test/small\_2} & 9/10 & 9/10 & 8/10 & 8/10 \\*
  \cdashline{2-12}[.4pt/1.4pt]
  & \multirow{2}{*}{M} & Card deck & \othumb{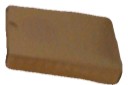} & \othumb{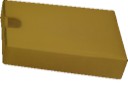} & 121.8 & 29.5 & \texttt{test/medium\_1} & 10/10 & 4/10 & 8/10 & 8/10 \\*
  &                    & Sponge & \othumb{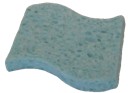} & \othumb{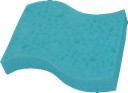} & 10.0 & 160.1 & \texttt{test/medium\_2} & 8/10 & 8/10 & 6/10 & 7/10 \\*
  \cdashline{2-12}[.4pt/1.4pt]
  & \multirow{2}{*}{L} & Wipe dispenser & \othumb{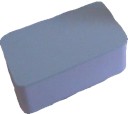} & \othumb{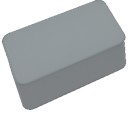} & 237.3 & 1317.3 & \texttt{test/large\_1} & 10/10 & 3/10 & 0/10 & 4/10 \\*
  &                    & Storage bin & \othumb{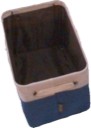} & \othumb{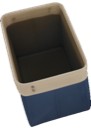} & 237.2 & 2632.3 & \texttt{test/large\_2} & 10/10 & 7/10 & 10/10 & 8/10 \\
\midrule
\multirow{6}{*}{Appendaged}
  & \multirow{2}{*}{S} & Nail clipper & \othumb{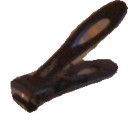} & \othumb{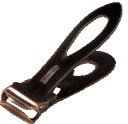} & 43.9 & 5.9 & \texttt{test/small\_1} & 9/10 & 3/10 & 5/10 & 4/10 \\*
  &                    & Lock & \othumb{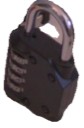} & \othumb{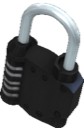} & 111.8 & 30.1 & \texttt{test/small\_2} & 10/10 & 8/10 & 5/10 & 6/10 \\*
  \cdashline{2-12}[.4pt/1.4pt]
  & \multirow{2}{*}{M} & Dustpan & \othumb{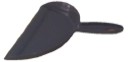} & \othumb{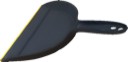} & 41.1 & 243.1 & \texttt{test/medium\_1} & 9/10 & 6/10 & 6/10 & 6/10 \\*
  &                    & Handbell & \othumb{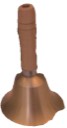} & \othumb{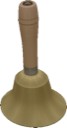} & 142.7 & 55.8 & \texttt{test/medium\_2} & 10/10 & 7/10 & 10/10 & 4/10 \\*
  \cdashline{2-12}[.4pt/1.4pt]
  & \multirow{2}{*}{L} & Saucepan & \othumb{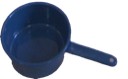} & \othumb{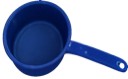} & 360.9 & 398.0 & \texttt{test/large\_1} & 10/10 & 10/10 & 4/10 & 7/10 \\*
  &                    & Picnic basket & \othumb{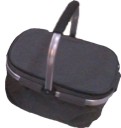} & \othumb{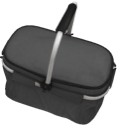} & 625.8 & 18021.7 & \texttt{test/large\_2} & 10/10 & 5/10 & 9/10 & 6/10 \\
\midrule
\multirow{6}{*}{Amorphous}
  & \multirow{2}{*}{S} & Rubber duck & \othumb{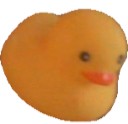} & \othumb{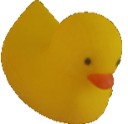} & 10.8 & 41.6 & \texttt{test/small\_1} & 10/10 & 5/10 & 10/10 & 9/10 \\*
  &                    & Tape measure & \othumb{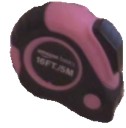} & \othumb{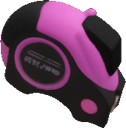} & 187.2 & 31.2 & \texttt{test/small\_2} & 10/10 & 9/10 & 8/10 & 5/10 \\*
  \cdashline{2-12}[.4pt/1.4pt]
  & \multirow{2}{*}{M} & Tape dispenser & \othumb{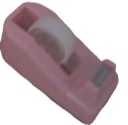} & \othumb{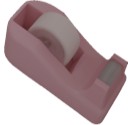} & 214.9 & 56.2 & \texttt{test/medium\_1} & 10/10 & 10/10 & 3/10 & 4/10 \\*
  &                    & Grapes & \othumb{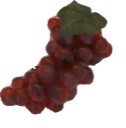} & \othumb{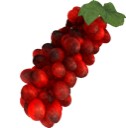} & 55.8 & 191.6 & \texttt{test/medium\_2} & 9/10 & 5/10 & 10/10 & 3/10 \\*
  \cdashline{2-12}[.4pt/1.4pt]
  & \multirow{2}{*}{L} & Headphones & \othumb{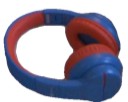} & \othumb{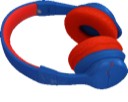} & 193.9 & 495.2 & \texttt{test/large\_1} & 6/10 & 4/10 & 6/10 & 2/10 \\*
  &                    & Easel & \othumb{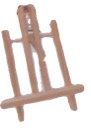} & \othumb{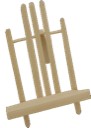} & 119.7 & 172.8 & \texttt{test/large\_2} & 7/10 & 5/10 & 8/10 & 7/10 \\
\midrule
\multicolumn{12}{l}{\textbf{\benchmark \texttt{val}}} \\*
\midrule
\multirow{12}{*}{Cylindrical}
  & \multirow{4}{*}{S} & Battery & \othumb{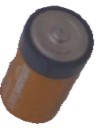} & \othumb{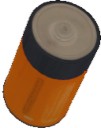} & 132.9 & 56.5 & \texttt{val/small\_3} & 10/10 & 9/10 & -- & -- \\*
  &                    & Glue tube & \othumb{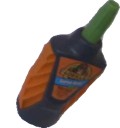} & \othumb{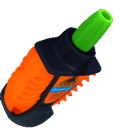} & 31.2 & 82.8 & \texttt{val/small\_4} & 10/10 & 7/10 & -- & -- \\*
  &                    & Perfume bottle & \othumb{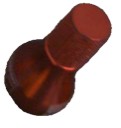} & \othumb{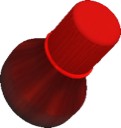} & 142.6 & 142.0 & \texttt{val/small\_5} & 10/10 & 9/10 & -- & -- \\*
  &                    & Toothpaste & \othumb{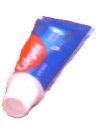} & \othumb{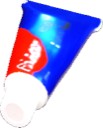} & 33.2 & 52.1 & \texttt{val/small\_6} & 8/10 & 2/10 & -- & -- \\*
  \cdashline{2-12}[.4pt/1.4pt]
  & \multirow{4}{*}{M} & Almonds & \othumb{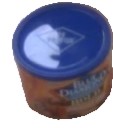} & \othumb{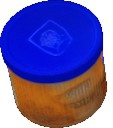} & 209.5 & 413.8 & \texttt{val/medium\_3} & 10/10 & 6/10 & -- & -- \\*
  &                    & Thermos & \othumb{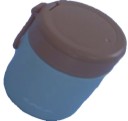} & \othumb{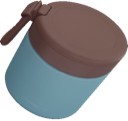} & 297.3 & 777.9 & \texttt{val/medium\_4} & 9/10 & 9/10 & -- & -- \\*
  &                    & Coffee cup & \othumb{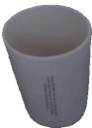} & \othumb{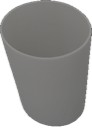} & 15.7 & 50.1 & \texttt{val/medium\_5} & 9/10 & 8/10 & -- & -- \\*
  &                    & Flowerpot & \othumb{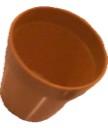} & \othumb{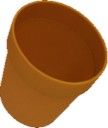} & 37.8 & 34.1 & \texttt{val/medium\_6} & 9/10 & 10/10 & -- & -- \\*
  \cdashline{2-12}[.4pt/1.4pt]
  & \multirow{4}{*}{L} & French vanilla & \othumb{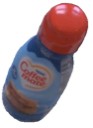} & \othumb{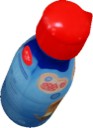} & 1109.3 & 1216.1 & \texttt{val/large\_3} & 10/10 & 10/10 & -- & -- \\*
  &                    & Mixing bowl & \othumb{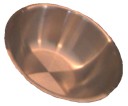} & \othumb{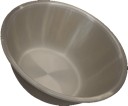} & 245.2 & 332.6 & \texttt{val/large\_4} & 8/10 & 9/10 & -- & -- \\*
  &                    & Vase & \othumb{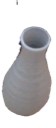} & \othumb{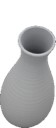} & 77.6 & 286.5 & \texttt{val/large\_5} & 9/10 & 10/10 & -- & -- \\*
  &                    & Oats & \othumb{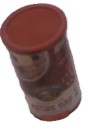} & \othumb{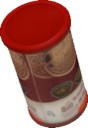} & 562.3 & 1695.2 & \texttt{val/large\_6} & 10/10 & 9/10 & -- & -- \\
\midrule
\multirow{12}{*}{Spheroidal}
  & \multirow{4}{*}{S} & Squash ball & \othumb{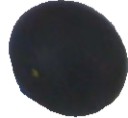} & \othumb{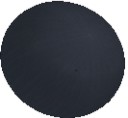} & 23.1 & 39.4 & \texttt{val/small\_3} & 10/10 & 8/10 & -- & -- \\*
  &                    & Apricot & \othumb{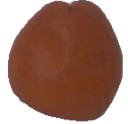} & \othumb{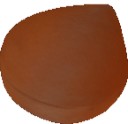} & 72.8 & 90.8 & \texttt{val/small\_4} & 10/10 & 8/10 & -- & -- \\*
  &                    & Small onion & \othumb{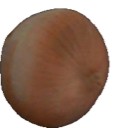} & \othumb{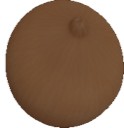} & 85.7 & 121.5 & \texttt{val/small\_5} & 10/10 & 6/10 & -- & -- \\*
  &                    & Garlic bulb & \othumb{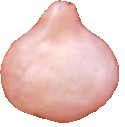} & \othumb{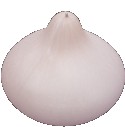} & 50.5 & 75.0 & \texttt{val/small\_6} & 10/10 & 10/10 & -- & -- \\*
  \cdashline{2-12}[.4pt/1.4pt]
  & \multirow{4}{*}{M} & Apple & \othumb{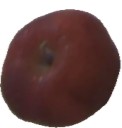} & \othumb{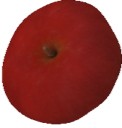} & 230.4 & 337.1 & \texttt{val/medium\_3} & 10/10 & 8/10 & -- & -- \\*
  &                    & Bundled socks & \othumb{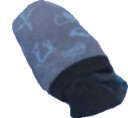} & \othumb{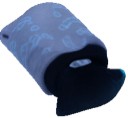} & 49.6 & 314.4 & \texttt{val/medium\_4} & 10/10 & 9/10 & -- & -- \\*
  &                    & Bell pepper & \othumb{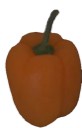} & \othumb{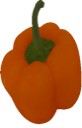} & 180.6 & 429.0 & \texttt{val/medium\_5} & 9/10 & 9/10 & -- & -- \\*
  &                    & Massager & \othumb{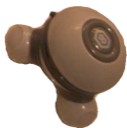} & \othumb{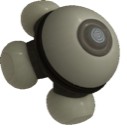} & 174.1 & 372.2 & \texttt{val/medium\_6} & 10/10 & 10/10 & -- & -- \\*
  \cdashline{2-12}[.4pt/1.4pt]
  & \multirow{4}{*}{L} & Eggplant & \othumb{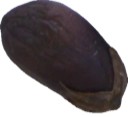} & \othumb{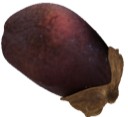} & 530.9 & 866.4 & \texttt{val/large\_3} & 10/10 & 8/10 & -- & -- \\*
  &                    & Cabbage & \othumb{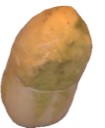} & \othumb{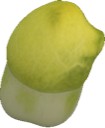} & 1027.1 & 2049.2 & \texttt{val/large\_4} & 10/10 & 7/10 & -- & -- \\*
  &                    & Small globe & \othumb{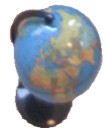} & \othumb{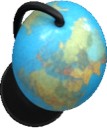} & 67.6 & 922.8 & \texttt{val/large\_5} & 10/10 & 9/10 & -- & -- \\*
  &                    & Squash & \othumb{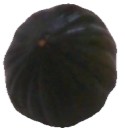} & \othumb{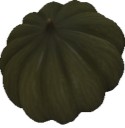} & 1005.9 & 1184.4 & \texttt{val/large\_6} & 9/10 & 8/10 & -- & -- \\
\midrule
\multirow{12}{*}{Prismatic}
  & \multirow{4}{*}{S} & Wooden cube & \othumb{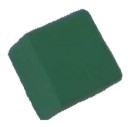} & \othumb{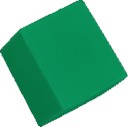} & 13.2 & 35.0 & \texttt{val/small\_3} & 10/10 & 8/10 & -- & -- \\*
  &                    & Salt shaker & \othumb{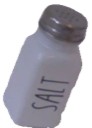} & \othumb{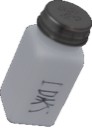} & 87.8 & 136.2 & \texttt{val/small\_4} & 10/10 & 9/10 & -- & -- \\*
  &                    & Wooden bridge & \othumb{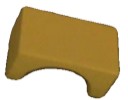} & \othumb{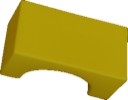} & 12.9 & 10.7 & \texttt{val/small\_5} & 10/10 & 9/10 & -- & -- \\*
  &                    & Anchovies & \othumb{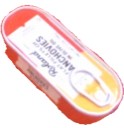} & \othumb{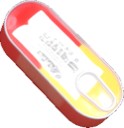} & 67.3 & 76.5 & \texttt{val/small\_6} & 2/10 & 8/10 & -- & -- \\*
  \cdashline{2-12}[.4pt/1.4pt]
  & \multirow{4}{*}{M} & Outlets & \othumb{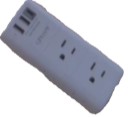} & \othumb{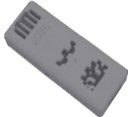} & 168.7 & 272.4 & \texttt{val/medium\_3} & 9/10 & 8/10 & -- & -- \\*
  &                    & Pocky & \othumb{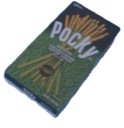} & \othumb{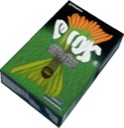} & 92.2 & 436.1 & \texttt{val/medium\_4} & 7/10 & 0/10 & -- & -- \\*
  &                    & Milk box & \othumb{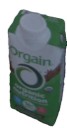} & \othumb{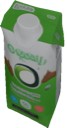} & 370.5 & 423.6 & \texttt{val/medium\_5} & 10/10 & 10/10 & -- & -- \\*
  &                    & Remote & \othumb{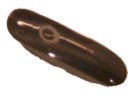} & \othumb{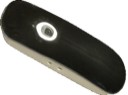} & 54.1 & 24.4 & \texttt{val/medium\_6} & 10/10 & 7/10 & -- & -- \\*
  \cdashline{2-12}[.4pt/1.4pt]
  & \multirow{4}{*}{L} & Keyboard & \othumb{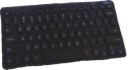} & \othumb{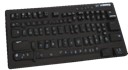} & 176.4 & 827.2 & \texttt{val/large\_3} & 10/10 & 8/10 & -- & -- \\*
  &                    & Milk carton & \othumb{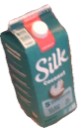} & \othumb{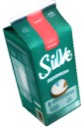} & 1970.0 & 2052.5 & \texttt{val/large\_4} & 8/10 & 4/10 & -- & -- \\*
  &                    & Cereal box & \othumb{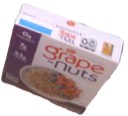} & \othumb{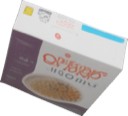} & 638.7 & 2251.7 & \texttt{val/large\_5} & 10/10 & 9/10 & -- & -- \\*
  &                    & Picture frame & \othumb{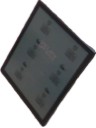} & \othumb{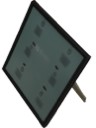} & 223.4 & 595.3 & \texttt{val/large\_6} & 4/10 & 2/10 & -- & -- \\
\midrule
\multirow{12}{*}{Appendaged}
  & \multirow{4}{*}{S} & Spring clamp & \othumb{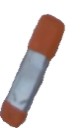} & \othumb{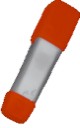} & 7.9 & 5.7 & \texttt{val/small\_3} & 9/10 & 6/10 & -- & -- \\*
  &                    & Wrench & \othumb{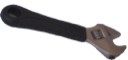} & \othumb{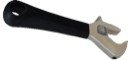} & 118.1 & 42.3 & \texttt{val/small\_4} & 8/10 & 9/10 & -- & -- \\*
  &                    & Spoon & \othumb{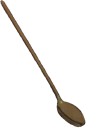} & \othumb{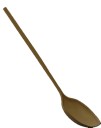} & 18.5 & 5.0 & \texttt{val/small\_5} & 5/10 & 3/10 & -- & -- \\*
  &                    & Screwdriver & \othumb{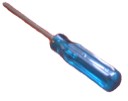} & \othumb{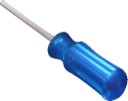} & 94.2 & 115.0 & \texttt{val/small\_6} & 10/10 & 9/10 & -- & -- \\*
  \cdashline{2-12}[.4pt/1.4pt]
  & \multirow{4}{*}{M} & Hair brush & \othumb{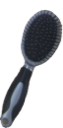} & \othumb{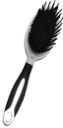} & 90.7 & 250.3 & \texttt{val/medium\_3} & 7/10 & 5/10 & -- & -- \\*
  &                    & Hammer & \othumb{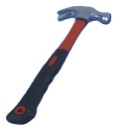} & \othumb{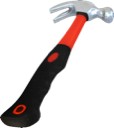} & 390.0 & 237.3 & \texttt{val/medium\_4} & 9/10 & 4/10 & -- & -- \\*
  &                    & Mug & \othumb{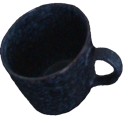} & \othumb{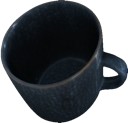} & 425.3 & 156.9 & \texttt{val/medium\_5} & 9/10 & 5/10 & -- & -- \\*
  &                    & Cleaning brush & \othumb{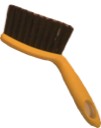} & \othumb{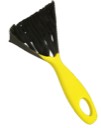} & 56.7 & 210.7 & \texttt{val/medium\_6} & 7/10 & 4/10 & -- & -- \\*
  \cdashline{2-12}[.4pt/1.4pt]
  & \multirow{4}{*}{L} & Dumbbell & \othumb{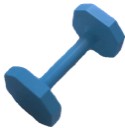} & \othumb{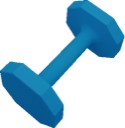} & 126.3 & 578.1 & \texttt{val/large\_3} & 8/10 & 7/10 & -- & -- \\*
  &                    & Frying pan & \othumb{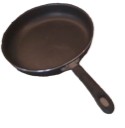} & \othumb{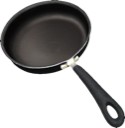} & 1691.9 & 800.9 & \texttt{val/large\_4} & 10/10 & 10/10 & -- & -- \\*
  &                    & Watering can & \othumb{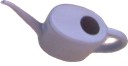} & \othumb{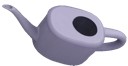} & 95.2 & 430.4 & \texttt{val/large\_5} & 10/10 & 9/10 & -- & -- \\*
  &                    & Ab roller & \othumb{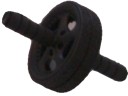} & \othumb{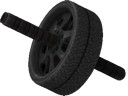} & 506.7 & 668.4 & \texttt{val/large\_6} & 8/10 & 8/10 & -- & -- \\
\midrule
\multirow{12}{*}{Amorphous}
  & \multirow{4}{*}{S} & Windup toy & \othumb{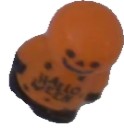} & \othumb{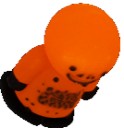} & 12.7 & 37.0 & \texttt{val/small\_3} & 10/10 & 8/10 & -- & -- \\*
  &                    & Broccoli & \othumb{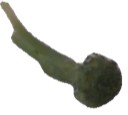} & \othumb{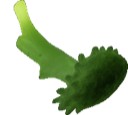} & 12.8 & 21.0 & \texttt{val/small\_4} & 10/10 & 8/10 & -- & -- \\*
  &                    & Mini stapler & \othumb{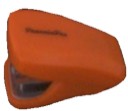} & \othumb{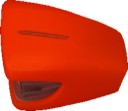} & 40.0 & 49.0 & \texttt{val/small\_5} & 10/10 & 5/10 & -- & -- \\*
  &                    & Jalapeno & \othumb{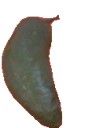} & \othumb{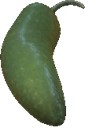} & 34.1 & 83.0 & \texttt{val/small\_6} & 9/10 & 3/10 & -- & -- \\*
  \cdashline{2-12}[.4pt/1.4pt]
  & \multirow{4}{*}{M} & Chips & \othumb{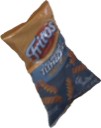} & \othumb{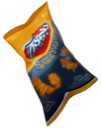} & 98.3 & 823.5 & \texttt{val/medium\_3} & 10/10 & 6/10 & -- & -- \\*
  &                    & Bird toy & \othumb{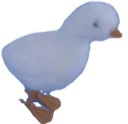} & \othumb{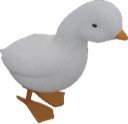} & 11.4 & 212.0 & \texttt{val/medium\_4} & 10/10 & 3/10 & -- & -- \\*
  &                    & Bear figure & \othumb{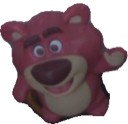} & \othumb{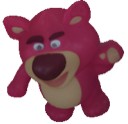} & 116.4 & 49.7 & \texttt{val/medium\_5} & 10/10 & 8/10 & -- & -- \\*
  &                    & Turtle toy & \othumb{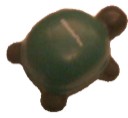} & \othumb{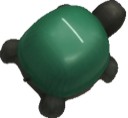} & 24.2 & 31.2 & \texttt{val/medium\_6} & 10/10 & 8/10 & -- & -- \\*
  \cdashline{2-12}[.4pt/1.4pt]
  & \multirow{4}{*}{L} & Sneaker & \othumb{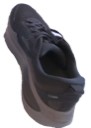} & \othumb{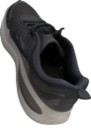} & 327.0 & 857.3 & \texttt{val/large\_3} & 8/10 & 6/10 & -- & -- \\*
  &                    & Robot hand & \othumb{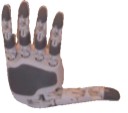} & \othumb{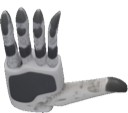} & 393.9 & 288.2 & \texttt{val/large\_4} & 10/10 & 4/10 & -- & -- \\*
  &                    & Backpack & \othumb{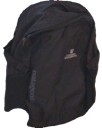} & \othumb{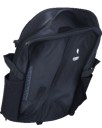} & 735.0 & 12340.3 & \texttt{val/large\_5} & 6/10 & 2/10 & -- & -- \\*
  &                    & Earmuffs & \othumb{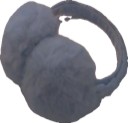} & \othumb{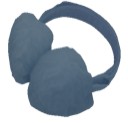} & 95.1 & 1181.0 & \texttt{val/large\_6} & 9/10 & 9/10 & -- & -- \\
\end{longtable}
\addtocounter{table}{-1}% longtable steps the table counter; caption already supplied via \captionof
}
\clearpage% keep the following subsection off the table's two pages

\section{Simulation Grasping Evaluation}
\label{sec:simulation_results}

This section details the simulated MANO hand used for grasp rollouts, the realism caveats behind the human grasp oracle, and the dense metrics that complement success rate. Per-object simulation success rates for all $90$ objects, the GT Sim SR and Sim SR columns, are reported in Table~\ref{tab:object_stats}. Qualitative predicted grasps on \benchmark scenes are shown in the main text (Figure~\ref{fig:bench_grasps}).

\subsection{Simulated MANO Hand}
\label{sec:mano_mjcf}

To evaluate grasps in MuJoCo~\cite{todorov2012mujoco} we bake the fixed-shape MANO hand (Appendix~\ref{sec:fixed_shape}) into a physics-ready MJCF directly from the MANO model~\cite{romero2017mano}. We forward the differentiable MANO layer~\cite{yang2021cpf} at the flat zero pose with the fixed shape $\boldsymbol{\beta}$, assign each of the $778$ mesh vertices to the MANO bone of maximum linear-blend-skinning weight, and model each phalanx as a capsule spanning its proximal-to-child joints, keeping the palm as a convex mesh. We use this capsule hand for all rollouts as its smooth normals give more stable contacts. A right and a left hand are produced from the same shape, the left by mirroring the shape basis about its $x$-axis.

A total mass of $0.4$\,kg is split across bones by hull volume, each with a uniform-density inertia. The wrist is a $6$-DoF free joint (three slides plus a ball joint) and every finger joint is a ball joint, all driven by critically damped position actuators stable at the $2$\,ms simulation step. A softer, higher-priority contact compliance is applied to the palm, since recorded human grasps often rest the palm $1$--$2$\,cm inside the rigid object mesh and a stiff palm would eject it on lift.

\subsection{The Human Grasp Oracle}
\label{sec:sim_realism}

The human grasp oracle replays the recorded human grasps and does not reach $100\%$ success. The gap is informative rather than a flaw of the model, and stems from several sources: (i) Aria Gen 2 hand-tracking error~\cite{projectaria_gen2_mps_benchmarks}, where occluded fingers tend to be tracked as too open or too closed, so the recorded grasp is slightly loose or tight; (ii) the objects are rigid, neither articulated nor deformable; (iii) the pre-grasp $\rightarrow$ grasp $\rightarrow$ lift trajectory is open-loop and programmed to resemble a real-robot execution, unlike the closed-loop human grasp it is derived from. The oracle therefore measures the quality of the data and assets, and provides a realistic ceiling for the learned model.

\paragraph{How precise must grasps be?}
\label{sec:gt_grasp_offset}
\begin{wrapfigure}[15]{r}{0.4\linewidth}
    \centering
    \includegraphics[width=\linewidth]{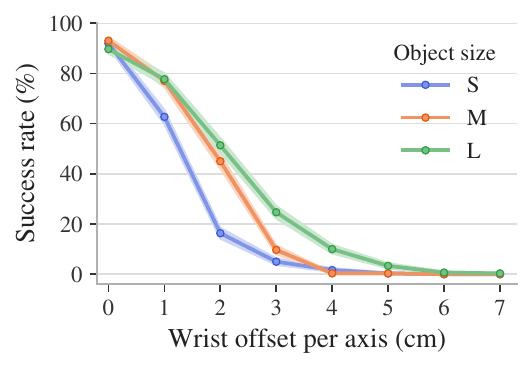}
    \caption{\textbf{GT grasp offset.} \benchmark SR (\%) vs.\ ground-truth grasp displacement along all three wrist axes.}
    \label{fig:gt_grasp_offset}
\end{wrapfigure}

To quantify how much spatial precision a successful grasp demands, we displace each ground-truth grasp simultaneously along all three wrist axes before execution,
\begin{equation}
\mathbf{t}' = \mathbf{t} + \delta\,(\mathbf{e}_x + \mathbf{e}_y + \mathbf{e}_z), \qquad \delta \in \{1, 2, \ldots\}\,\text{cm},
\end{equation}
and measure the resulting SR over all $90$ \benchmark objects (\texttt{val} and \texttt{test}), broken down by object size. Figure~\ref{fig:gt_grasp_offset} reports the falloff. At zero offset the grasp is the recorded human grasp, so SR equals the oracle ($\sim$90\%); it then drops steeply, as only $2$\,cm per axis (a Euclidean shift of $2\sqrt{3}\approx3.5$\,cm) already roughly halves SR for medium and large objects and cuts it below $20\%$ for small ones. The tolerance scales with object size, as expected. Small objects collapse to near-zero SR by $3$--$4$\,cm per axis, whereas large objects, which afford larger stable contact regions, retain about $25\%$ SR at $3$\,cm and $10\%$ at $4$\,cm before vanishing by $6$\,cm. Even the most forgiving objects therefore tolerate only a few centimeters of error, showing that the benchmark rewards precise placement rather than coarse proximity to the object.

\subsection{Dense Grasp Quality Metrics}
\label{sec:dense_metrics}

\paragraph{Penetration metrics.}
\label{sec:penetration}
Beyond the fingertip contact error (\S~\ref{sec:exp_sim}, Eq.~\ref{eq:fc_error}), we report two penetration metrics at grasp closure, both standard in the hand-object grasping literature \cite{yang2022oakinklargescaleknowledgerepository}: the maximum penetration depth $\max_i \max(0, -d_i)$ over hand surface samples $i$ with signed surface distance $d_i$ (negative inside the object), and the hand-object intersection volume $\lvert \mathcal{H} \cap \mathcal{O}\rvert$. On their own, however, these metrics are flawed: a grasp can have low penetration depth and volume while being entirely unstable, for instance with the object merely tangent to the back of the palm. Two grasps, one 10 cm inside the object and one 10 cm outside, will average to have 0 penetration depth. Penetration volume is similarly flawed. It is normally reported as ``lower is better'', but is 0 for any grasp that is entirely outside the object, regardless of distance. 

To rectify this, we report the percentage of grasps in which the object is penetrated by the hand at grasp closure. We report penetration depth and volume averages over only grasps that are intersecting the object. For grasps not intersecting the object, we report an average miss distance, or the nearest distance from the hand to the object surface. Lower is better for all three of these metrics. Table~\ref{tab:sim_ablations_extended} reports these metrics on our ablations, extending the main ablation (Table~\ref{tab:sim_ablations}) across the same models on the \texttt{val} and \texttt{test} splits. The full \method model performs best on these new metrics, futher evidence of the grasp quality.

Still, these static penetration metrics should be read with care. Because our grasps come from real human grasps on real objects, and the simulated objects are rigid, a grasp of a soft object reports large penetration even though it is correct; a grasp of a pillow, for example, is almost entirely in penetration against its rigid mesh. As guidance, read the table by row against its SR in Table~\ref{tab:sim_ablations_extended} rather than minimizing penetration alone: a variant with both low penetration and high SR is cleanly accurate, whereas low penetration with low SR indicates fingertips stopping short of contact, and the oracle's nonzero penetration sets the floor attributable to soft objects and tracking noise. We therefore treat penetration as diagnostic rather than a standalone quality score.

\begin{table}[t]
    \centering
    \setlength{\tabcolsep}{5pt}
    {\scriptsize
    \textbf{\benchmark \texttt{val}}\\[2pt]
    \begin{tabular}{lccccccc}
      \toprule
      Method & SR (\%) $\uparrow$ & Obj. & Int.\ (\%) & Pen.\ depth (mm) & Miss dist.\ (mm) & Pen.\ vol.\ (\%) & FC error (mm) $\downarrow$ \\
      \midrule
      \rowcolor{rowhl}
      RGB + PC (full \method) & \textbf{71.5} {\scriptsize $\pm$ 1.8} & 59/60 & \textbf{90.5} {\scriptsize $\pm$ 1.2} & \textbf{13.1} {\scriptsize $\pm$ 0.4} & \textbf{16.4} {\scriptsize $\pm$ 3.4} & \textbf{1.9} {\scriptsize $\pm$ 0.1} & \textbf{19.0} {\scriptsize $\pm$ 0.8} \\
      \quad w/o crop & 61.2 {\scriptsize $\pm$ 2.0} & \textbf{60}/60 & 88.0 {\scriptsize $\pm$ 1.3} & 15.4 {\scriptsize $\pm$ 0.5} & 21.3 {\scriptsize $\pm$ 3.4} & 3.1 {\scriptsize $\pm$ 0.2} & 21.6 {\scriptsize $\pm$ 0.9} \\
      \quad w/o point paint & 61.8 {\scriptsize $\pm$ 2.0} & 57/60 & 88.2 {\scriptsize $\pm$ 1.3} & 14.3 {\scriptsize $\pm$ 0.4} & 29.8 {\scriptsize $\pm$ 4.6} & 2.6 {\scriptsize $\pm$ 0.2} & 21.9 {\scriptsize $\pm$ 1.0} \\
      \quad w/o 3D loss & 39.2 {\scriptsize $\pm$ 2.0} & 57/60 & 68.3 {\scriptsize $\pm$ 1.9} & 14.4 {\scriptsize $\pm$ 0.6} & 20.9 {\scriptsize $\pm$ 1.8} & 3.5 {\scriptsize $\pm$ 0.3} & 33.0 {\scriptsize $\pm$ 1.2} \\
      PC only & 64.2 {\scriptsize $\pm$ 2.0} & 58/60 & 81.3 {\scriptsize $\pm$ 1.6} & 13.6 {\scriptsize $\pm$ 0.4} & 32.0 {\scriptsize $\pm$ 3.1} & 2.2 {\scriptsize $\pm$ 0.1} & 25.6 {\scriptsize $\pm$ 1.2} \\
      \quad w/o crop & 47.3 {\scriptsize $\pm$ 2.0} & 57/60 & 72.2 {\scriptsize $\pm$ 1.8} & 15.6 {\scriptsize $\pm$ 0.6} & 30.6 {\scriptsize $\pm$ 3.3} & 2.8 {\scriptsize $\pm$ 0.2} & 32.6 {\scriptsize $\pm$ 1.5} \\
      RGB only & 26.8 {\scriptsize $\pm$ 1.8} & 46/60 & 42.5 {\scriptsize $\pm$ 2.0} & 15.5 {\scriptsize $\pm$ 0.7} & 74.9 {\scriptsize $\pm$ 3.4} & 5.5 {\scriptsize $\pm$ 0.6} & 95.4 {\scriptsize $\pm$ 3.6} \\
      \midrule
      Human grasp (oracle) & 90.3 {\scriptsize $\pm$ 1.2} & 60/60 & 98.2 {\scriptsize $\pm$ 0.5} & 13.1 {\scriptsize $\pm$ 0.4} & 2.3 {\scriptsize $\pm$ 0.7} & 2.5 {\scriptsize $\pm$ 0.1} & 9.4 {\scriptsize $\pm$ 0.3} \\
      \bottomrule
    \end{tabular}

    \vspace{8pt}
    \textbf{\benchmark \texttt{test}}\\[2pt]
    \begin{tabular}{lccccccc}
      \toprule
      Method & SR (\%) $\uparrow$ & Obj. & Int.\ (\%) & Pen.\ depth (mm) & Miss dist.\ (mm) & Pen.\ vol.\ (\%) & FC error (mm) $\downarrow$ \\
      \midrule
      \rowcolor{rowhl}
      RGB + PC (full \method) & \textbf{73.0} {\scriptsize $\pm$ 2.6} & \textbf{30}/30 & \textbf{93.0} {\scriptsize $\pm$ 1.5} & \textbf{11.2} {\scriptsize $\pm$ 0.5} & \textbf{12.2} {\scriptsize $\pm$ 3.7} & \textbf{1.6} {\scriptsize $\pm$ 0.1} & \textbf{14.6} {\scriptsize $\pm$ 0.9} \\
      \quad w/o crop & 58.0 {\scriptsize $\pm$ 2.8} & 29/30 & 82.0 {\scriptsize $\pm$ 2.2} & 13.8 {\scriptsize $\pm$ 0.8} & 20.9 {\scriptsize $\pm$ 5.0} & 2.4 {\scriptsize $\pm$ 0.2} & 25.7 {\scriptsize $\pm$ 1.5} \\
      \quad w/o point paint & 58.3 {\scriptsize $\pm$ 2.8} & \textbf{30}/30 & 83.3 {\scriptsize $\pm$ 2.2} & 12.2 {\scriptsize $\pm$ 0.7} & 28.7 {\scriptsize $\pm$ 6.7} & 2.3 {\scriptsize $\pm$ 0.2} & 23.3 {\scriptsize $\pm$ 1.7} \\
      \quad w/o 3D loss & 32.7 {\scriptsize $\pm$ 2.7} & 27/30 & 65.0 {\scriptsize $\pm$ 2.8} & 12.1 {\scriptsize $\pm$ 0.9} & 29.2 {\scriptsize $\pm$ 4.5} & 2.2 {\scriptsize $\pm$ 0.2} & 35.7 {\scriptsize $\pm$ 2.2} \\
      PC only & 70.7 {\scriptsize $\pm$ 2.6} & \textbf{30}/30 & 84.3 {\scriptsize $\pm$ 2.1} & 11.3 {\scriptsize $\pm$ 0.6} & 22.5 {\scriptsize $\pm$ 3.9} & \textbf{1.6} {\scriptsize $\pm$ 0.1} & 22.1 {\scriptsize $\pm$ 1.5} \\
      \quad w/o crop & 50.0 {\scriptsize $\pm$ 2.9} & 29/30 & 71.7 {\scriptsize $\pm$ 2.6} & 12.6 {\scriptsize $\pm$ 0.8} & 28.8 {\scriptsize $\pm$ 3.9} & 2.4 {\scriptsize $\pm$ 0.3} & 32.8 {\scriptsize $\pm$ 2.2} \\
      RGB only & 29.7 {\scriptsize $\pm$ 2.6} & 19/30 & 46.7 {\scriptsize $\pm$ 2.9} & 12.7 {\scriptsize $\pm$ 0.9} & 90.5 {\scriptsize $\pm$ 7.3} & 5.2 {\scriptsize $\pm$ 0.6} & 108.6 {\scriptsize $\pm$ 5.1} \\
      \midrule
      Human grasp (oracle) & 94.0 {\scriptsize $\pm$ 1.4} & 30/30 & 99.3 {\scriptsize $\pm$ 0.5} & 10.2 {\scriptsize $\pm$ 0.4} & 1.5 {\scriptsize $\pm$ 0.1} & 1.3 {\scriptsize $\pm$ 0.1} & 7.4 {\scriptsize $\pm$ 0.3} \\
      \bottomrule
    \end{tabular}}
    \vspace{6pt}
    \caption{\textbf{Extended ablation metrics.} Additional penetration and contact metrics on \benchmark for the ablation models of Table~\ref{tab:sim_ablations}, at each model's best-\texttt{val}-SR checkpoint, on the \texttt{val} (top) and \texttt{test} (bottom) splits. Per-grasp mean $\pm$ SE over 600 \texttt{val} / 300 \texttt{test} grasps. Obj.\ counts objects with $\geq$1 successful grasp (out of 60 \texttt{val} / 30 \texttt{test}). Human grasp is an oracle upper bound.}
    \label{tab:sim_ablations_extended}
\end{table}

\section{Real-World Grasping Evaluation}
\label{sec:real_results}

This section details the real-world evaluation design for the tabletop and in-the-wild settings, the retargeting of predicted MANO grasps to the Ability and WUJI robot hands, qualitative trends, and the value of dexterity over a gripper. Per-object real-world success rates are reported in the Tabletop SR and Wild SR columns of Table~\ref{tab:object_stats}.

\subsection{Real-World Evaluation Design}
\label{sec:real_design}

\begin{wrapfigure}{r}{0.23\linewidth}
    \centering
    \includegraphics[width=\linewidth]{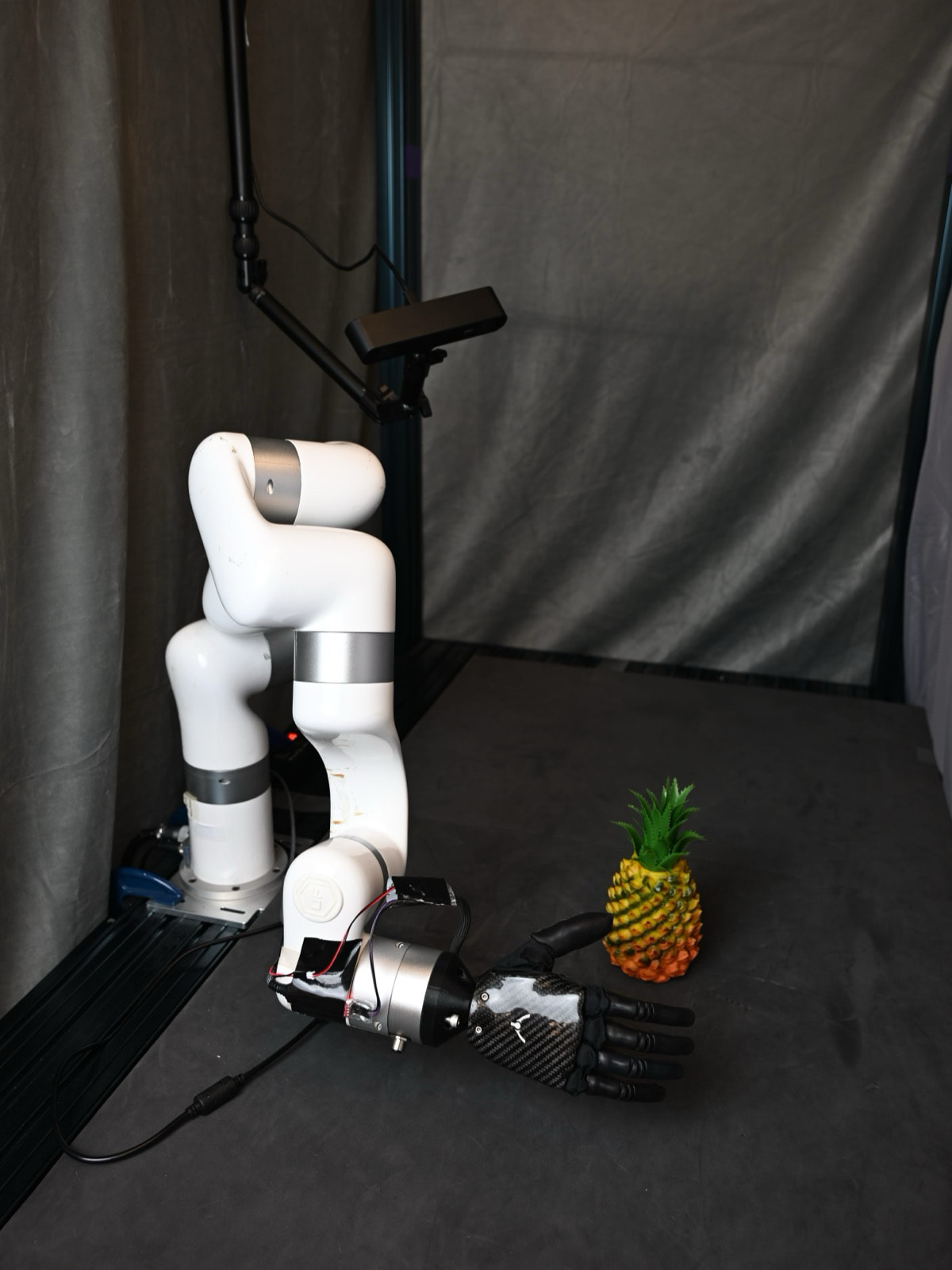}
    \caption{\textbf{Real-robot tabletop setup.} Ability hand + ZED camera + xArm.}
    \label{fig:tabletop_setup}
\end{wrapfigure}

We select the checkpoint with the best \benchmark \texttt{val} SR in simulation, and without ever testing the model in the real world, run all $300$ \benchmark \texttt{test} trials consecutively, each a single grasp prediction followed by one open-loop execution, with no cuts and no retries. We do not tune our model nor the open-loop execution strategy, discussed in \S~\ref{sec:retargeting}, to our \benchmark \texttt{test} split objects either on the tabletop or in-the-wild. Therefore, we believe our results are a fair assessment of \method's generalization ability. Uncut videos of all trials are on our website.

We place each object at a single location for tabletop trials (Figure~\ref{fig:tabletop_setup}), varying only its position and rotation slightly across the $10$ trials. We avoid clutter in the tabletop experiments to ensure repeatability, and to ensure fair comparison to Dex1B, which is constrained to a single-object tabletop setting. As \method is not constrained to the tabletop, we perform in-the-wild experiments in diverse cluttered settings to demonstrate its robustness to real-world clutter.

\subsection{Retargeting to Robot Hands}
\label{sec:retargeting}

\method predicts a MANO grasp, which we map to a target robot hand at deployment. We retarget to the Ability hand with AnyTeleop~\cite{qin2023anyteleop} and to the WUJI hand with WUJI retargeting~\cite{wuji2026retargeting}. Figure~\ref{fig:hand_sizes} compares the MANO hand with the robot hands.

Because robot hands differ in size from MANO, we align each robot hand's fingertips to MANO's using a single fixed offset, estimated from a simulation visualization. All offsets are expressed in the robot's hand frame (different from the MANO hand frame): the origin is at the palm, the $z$-axis points toward the fingertips, the $x$-axis points outward from the palmar surface, and the $y$-axis points toward the thumb. We apply no correction for the WUJI hand, whereas the Ability hand uses a translation of $[0.020,\,0,\,0.025]$~m together with a $10^\circ$ rotation about the $y$-axis. At deployment, the hand first moves to a fixed-shape pre-grasp whose wrist is offset from the predicted grasp by $[-0.05,\,0,\,-0.02]$~m for WUJI and $[-0.04,\,0,\,-0.01]$~m for Ability. It then linearly interpolates to the grasp pose, applies a per-joint force-close, and lifts the object while holding this pose.

\subsection{Qualitative Observations}
\label{sec:qualitative}

\textbf{Easiest and hardest objects.} The easiest objects are rounded and convex, sized to fit the hand (\eg pear, pineapple, hacky sack), which afford many stable enveloping grasps. The hardest are objects too large for the hand to wrap (\eg football, wipe dispenser) or irregular and articulated objects that are difficult to grasp open-loop (\eg nail clipper, headphones).

\textbf{Robustness.} Because \dataset is collected across many environments and lighting conditions, \method is robust to changes in lighting, object rotation, and viewpoint, and operates on both color and grayscale stereo cameras since the dataset includes grayscale frames. We also observe successful grasps on reflective surfaces, where stereo depth is noisier.

\textbf{Hands versus gripper.}
\label{sec:hands_vs_gripper}
Our comparison against CAP, a strong gripper policy, on identical objects shows that the benefit of dexterity is object-dependent. On large or irregular objects, an antipodal gripper finds no stable pair of opposing faces and tends to slip, whereas an enveloping multi-finger grasp cages the object. On heavy objects whose graspable feature is offset from the center of mass, a two-finger pinch approximates a single contact line that gravitational torque pivots the object out of, while a multi-finger grasp distributes contacts and resists that torque. On small, thin, or pinchable objects the gripper remains competitive. Dexterity thus pays off precisely on the large, irregular, heavy, and off-center objects that grippers handle worst.

\end{document}